\RequirePackage{fix-cm}
\documentclass[twocolumn]{svjour3}          
\smartqed  
\usepackage{amssymb}
\usepackage{graphicx}
\usepackage{url}
\usepackage{color}

\usepackage{amsmath}
\usepackage{enumitem}
\usepackage{multirow}
\usepackage{subfig}
\usepackage{tikz,forest}
\usepackage{booktabs}
\usepackage{pdflscape}
\usepackage{array}
\usepackage{makecell}
\usepackage{caption}
\usepackage{comment}
\usepackage[makeroom]{cancel}

\newcommand{\PreserveBackslash}[1]{\let\temp=\\#1\let\\=\temp}
\newcolumntype{C}[1]{>{\PreserveBackslash\centering}p{#1}}
\newcolumntype{R}[1]{>{\PreserveBackslash\raggedleft}p{#1}}
\newcolumntype{L}[1]{>{\PreserveBackslash\raggedright}p{#1}}

\usepackage{tikz}
\newcommand*\circled[4]{\tikz[baseline=(char.base)]{
    \node[shape=circle, fill=#2, draw=#3, text=#4, inner sep=2pt] (char) {#1};}}

\definecolor{darkblue}{rgb}{0.0, 0.0, 0.55}
\definecolor{darkcyan}{rgb}{0.0, 0.55, 0.55}

\journalname{Neural Computing and Applications}
\begin{document}


\title{Collaborative Training of Heterogeneous Reinforcement Learning Agents in Environments with Sparse Rewards:\\What and When to Share?}

\titlerunning{Collaborative Training of Heterogeneous Reinforcement Learning Agents under Sparse Rewards}
\authorrunning{Andres A., Villar-Rodriguez E., Del Ser J.}
\author{Alain Andres$^\ast$ 
\thanks{$^\ast$: Corresponding author (alain.andres@tecnalia.com)} \and Esther Villar-Rodriguez \and Javier Del Ser}


\institute{A. Andres, E. Villar-Rodriguez, J. Del Ser\at
           TECNALIA, Basque Research and Technology Alliance (BRTA), 48160 Derio, Spain \\
           \and
           A. Andres, J. Del Ser \at
           University of the Basque Country (UPV/EHU), 48013 Bilbao, Spain
}

\date{Received: date / Accepted: date}

\maketitle

\begin{abstract}
In the early stages of human life, babies develop their skills by exploring different scenarios motivated by their inherent satisfaction rather than by extrinsic rewards from the environment. This behavior, referred to as intrinsic motivation, has emerged as one solution to address the exploration challenge derived from reinforcement learning environments with sparse rewards. Diverse exploration approaches have been proposed to accelerate the learning process over single- and multi-agent problems with homogeneous agents. However, scarce studies have elaborated on collaborative learning frameworks between heterogeneous agents deployed into the same environment, but interacting with different instances of the latter without any prior knowledge. Beyond the heterogeneity, each agent's characteristics grant access only to a subset of the full state space, which may hide different exploration strategies and optimal solutions. In this work we combine ideas from intrinsic motivation and transfer learning. Specifically, we focus on sharing parameters in actor-critic model architectures and on combining information obtained through intrinsic motivation with the aim of having a more efficient exploration and faster learning. We test our strategies through experiments performed over a modified ViZDooM's \textit{My Way Home} scenario, which is more challenging than its original version and allows evaluating the heterogeneity between agents. Our results reveal different ways in which a collaborative framework with little additional computational cost can outperform an independent learning process without knowledge sharing. Additionally, we depict the need for modulating correctly the importance between the extrinsic and intrinsic rewards to avoid undesired agent behaviors.

\end{abstract}


\tableofcontents

\section{Introduction}\label{sec:introduction}

Reinforcement Learning (RL) has captured the attention of the research community in the last few years when its capabilities were shown not to be limited only to tabular settings, but also be used in complex state spaces by virtue of neural computation \cite{mnih2015human,silver2016alphago}. Broadly speaking, the learning process of these algorithms depends on the definition of a reward function, which establishes the objective an agent must accomplish by means of its behavioral policy. As a result, RL has spurred many advances and applications in robo\-tics, manufacturing, healthcare and transportation, among others \cite{li2019reinforcement}.

A natural problem in RL emerges from the fact that the design of such a reward function is not trivial. In this context, it is becoming increasingly popular to use \emph{sparse rewards} to determine whether the task has been solved by giving just a final feedback signal. By counterpart, these problems are harder to address by a RL agent, as there is little correlation between the success and different successive events that may occur during the interaction of the agent with the environment \cite{hare2019dealing}. As opposed to humans who have prior understanding about visual information and different events, agents may lack such a priori knowledge, which induces an increasing level of difficulty to decide the proper decision to be taken at each step \cite{dubey2018investigating}. 

In order to face the above challenge, several techniques have been proposed to date, including Intrinsic Motivation (IM) \cite{aubret2019survey}, Imitation Learning \cite{ross2011reduction,rusu2015policy} and Inverse Reinforcement Learning \cite{finn2016guided}, among others. Moreover, competitions have been held around these topics in order to encourage research efforts and evaluate the potential of new approaches \cite{juliani2019obstacle,kuttler2020nethack}. The reason for this notable interest arises from the psychological quest towards understanding how humans learn different behaviors in the absence of reinforcement signals, emphasizing on the importance of being motivated to explore new alternatives \cite{ryan2000intrinsic,grigorescu2020curiosity} and thereby relating the evolution in humans to its adoption in robotics \cite{oudeyer2016evolution,cangelosi2018babies}. Since then, significant research has been invested into RL algorithms designed to overcome exploration issues \cite{aubret2019survey}, where the agent is encouraged (\textit{motivated}) to follow a certain behavior (\textit{explore}) for its inherent satisfaction rather that for other exogenous stimuli (\textit{environment reward}) \cite{barto2013intrinsic}. Consequently, the agent generates its own intrinsic reward signals \cite{bellemare2016unifying,pathak2017curiosity} that help in the exploration stage to find out the best possible solutions. This idea can be further understood from the intuition gained from the following real-world example, which will later help formulate the problem tackled in this manuscript:

\vspace{2mm}\textit{A bike rider wants to downhill a given mountain across the shortest path and as fast as possible. However, he/she does not know the mountain, and the unique signal that determines when it has finishes is when arriving to the destination. Thus, he/she does not know whether the decision in a bifurcation was right, if he/she got stacked close to the final line, or even if he/she has spent too much time when compared to other bikers. Due to so much uncertainty without feedback signals, the agent (bike rider) should drive his/her decisions based on his/her own motivation and curiosity}.
\vspace{2mm}

On the other hand, when aiming to enhance the performance of a single agent and accelerate its learning process over dense or sparse scenarios, a broadly used strategy is to use algorithms that allow for the implementation of multiple distributed parallel environments \cite{mnih2016asynchronous,kapturowski2018recurrent,espeholt2018impala}. However, when multiple different agents/learners are involved, their policies may not be the same, so the aforementioned parallel algorithm implementations are not straightforward. It is in this niche where collaborative strategies arise \cite{johnson1994cooperative,gokhale1995collaborative}, most of which are proposed under the umbrella of multi-agent problems. For the sake of clarity, we refer as \emph{multi-agent} to the task where different agents interact between each other over the same environment in a cooperative or competitive manner, so that state transitions are the result of the joint action of all agents \cite{bucsoniu2010multi}. Nevertheless, there is little work related to how agents can share knowledge when they interact with different independent instances of the same environment. Following the above example:
 
\vspace{2mm}\textit{Instead of one bike rider, let us have two riders going downhill in the same mountain. Both of them learn independently from each other over the same environment and with the same opportunities, but they do not run during the same hours (namely, actions taken by one do not interfere into the decisions of the other). Knowledge gathered by any of them can positively impact the other rider if shared (when to slow down, how to take a curve, which path to follow, etc)}.
\vspace{2mm}

This type of problem can be exacerbated even further when those agents are heterogeneous in either their observation or action space, which probably lead them to have different optimal solutions and, in consequence, different action distributions: 

\vspace{2mm}\textit{Now both riders are assumed to have different bikes. One of such bikes allows making jumps over obstacles such as tree branches or rocks, whereas the other bike can not perform any jump. Consequently, some tracks of the mountain are only accessible for one agent. Indeed, the best global solution may be hindered behind those obstacle-blocked tracks.}
\vspace{2mm}

As can be inferred from the above example, the problem cannot be classified as \textit{single-agent}, nor does it belong to {multi-agent} RL. In the targeted setup, multiple agents with different skills interact with an own copy of the environment. As a consequence of having different action spaces, agents access to specific subsets of the state space due to the existence of restricted areas requiring special skills, which eventually yields individual optimal solutions (diversification of the optimal space). 

Ultimately, our problem resides in between single-  multi-agent settings, being uncertain how the heterogeneously skilled agents could explore efficiently. Is it advantageous to carry out the learning processes in a completely independent fashion? (as in a single-agent problem). Or should agents learn under a full collaborative learning framework? (as in multi-agent). Is there any hybrid learning strategy capable of deciding when to collaborate and when to learn in isolation?. When dealing with homogeneous agents (i.e., \textit{riders use the same bike model}), distributed learning strategies may be suitable; however, dealing with heterogeneous skills (corr. \textit{different bike models}) adds the major complication of having state-action pairs that cannot be executable by all agents.

When it comes to knowledge transfer among RL agents, the related literature has employed different terminologies to address similar problems, where similar terms have been used with equal meanings and in others are inconsistent \cite{da2019survey}. Here we refer as Collaborative Learning to the challenge of being able to reuse, combine or even adapt knowledge transferred from one source to another. In this sense, a wide area focused on these problems is Transfer Learning \cite{zhu2020transfer}. One way to tackle information exchange between agents is by assuming some kind of prior knowledge learned by any of the agents, so that it can be \emph{distilled} to other agents under a teacher-student framework \cite{rusu2015policy}. Unfortunately, this distillation is not straightforward to realize when agents are heterogeneous. Another approach is to transfer knowledge in an online fashion with no priors \cite{zhan2015online,lai2020dual}. However, works relying on this strategy have not considered the case of having agents with different characteristics. 

To the best of our knowledge, no previous work has tackled the problem of heterogeneous RL agents with sparse rewards when transfer knowledge is done in an online fashion for a faster learning of the agents, while potentially having different optimal solutions for the same environment. In this study we build upon preliminary findings reported in \cite{9534146} and extend them in terms of the following specific contributions:
\begin{itemize}
    \item We analyze the current state-of-the-art on intrinsic motivation techniques in both single and multi-agent RL problems, putting into perspective the need for designing ad-hoc learning strategies to efficiently cope with the heterogeneity between agents.
    \item We perform an experimental ablation study between different knowledge sharing strategies over a complex RL environment. The goal of this ablation study is to shed light on the best solution to address the heterogeneity between agents, which can be easily extended to other types of approaches. To this end, we focus on solving the environment with an on-policy algorithm that requires minimal memory footprint, as well as the least possible parallel environments so as to achieve good results with little computational effort.
    \item We provide insights about future research lines in the scope of heterogeneous agents, identifying what remains to be studied in problems where intrinsic motivation is used to face the sparseness issue.
\end{itemize}

The rest of the paper is structured as follows: Section \ref{sec:related_work} first reviews research works aligned with the scope of this article, focusing on exploration strategies in sparse rewards and knowledge reuse for collaborative strategies. Next, Section \ref{sec:problem_statement} formulates the problem tackled in our work, followed by Section \ref{sec:preliminary} where we introduce deep RL concepts and intrinsic reward generation techniques that surround our work. Section \ref{sec:methodology} proposes different mechanisms to tackle efficiently the problem under analysis. Section \ref{sec:experiments} describes the different experimental setups and the ablations performed in our experiments. Thereafter, Section \ref{sec:results} discusses on the obtained results. In Section \ref{sec:discussion} we explain the main problem we have found out, as well as the future directions that depart from this work and others related to intrinsic motivation and sparse rewards. Finally, Section \ref{sec:conclusions} summarizes the main conclusions of the study.


\section{Related Work}\label{sec:related_work}

Before proceeding with the statement of the problem under target, we first proceed by revising contributions reported in the past where intrinsic motivation mechanisms have been used in both single and multi-agent problems (Subsection \ref{ssec:intrinsic_motivation}). The section also overviews how knowledge is shared from the perspective of Transfer Learning in the scope of Deep Learning-based RL, which we hereafter refer to as DRL (Subsection \ref{ssec:drl_transfer}). Finally, rationale for the contribution of this work beyond the state of the art is given in Subsection \ref{ssec:contribution}.

\subsection{Intrinsic Motivation} \label{ssec:intrinsic_motivation}

Intrinsic motivation allows the agent to gain new knowledge autonomously in a certain environment without the necessity of having the supervision of an expert. This endows the agent with the ability of learning behaviors that are independent of their main task \cite{aubret2019survey}. 

Focusing on knowledge acquisition techniques, one way to generate intrinsic rewards are the so-called \emph{count-based} methods which, as its name suggests, calculate an exploration bonus based on how many times a given state has been visited by the agent. The use of pseudo-counts \cite{bellemare2016unifying,ostrovski2017count} is followed in non-tabular approaches, which have been improved by additional techniques such as hash functions \cite{tang2017exploration} and successor representations \cite{machado2020count}.

Another possibility is to compute intrinsic rewards by taking into account the prediction error between state transitions \cite{pathak2017curiosity,burda2018large,yang2019flow}. The problem of this family of solutions is that they tend to get attracted by stochastic inputs depending on the environment that are related by the dynamics of the scenarios rather than to actions selected by the agents. In order to minimize this problem, the work in \cite{savinov2018episodic} calculates the reachability from one state to others stored with an episodic memory. In \cite{burda2018exploration} the source of error is fixed to a target network to yield a deterministic error that decreases the more similar the outputs of a predictor network are to another learner source network. Further along this line, the authors in \cite{pathak2019self} propose to compute the intrinsic reward as the variance between the predictions issued by an ensemble of models.

One of the issues that may arise in settings utilizing intrinsic motivation strategies is a bad balance between intrinsic and extrinsic (i.e. environment-based) motivation, which could lead to undesired behaviors \cite{rosser2021curiosity,badia2020never,taiga2019benchmarking}. To overcome this issue, in \cite{badia2020agent57} a meta-controller is proposed to select a proper exploration/exploitation balance between intrinsic motivation and extrinsic reward. Alternatively, using concepts derived from meta-learning, several works have approximated the problem by defining a bi-level optimization framework where the optimization of the intrinsic network parameters are influenced by the direction of the extrinsic gradient, ensuring that the main extrinsic objective is increased \cite{zheng2018learning,du2019liir,dai2022diversity}.

Recently, another point of view outlined in \cite{ecoffet2021first} focuses on how to effectively take advantage of exploration techniques by re-initializing the agent in selective start points, which is also the core idea behind \cite{ugadiarov2021long}. Those approaches are motivated due to the \textit{detachment-derailment} dilemma\footnote{\emph{Detachment} refers to losing track of interesting areas to explore, whereas \emph{derailment} occurs when the exploratory mechanism of the algorithm prevents it from returning to previously visited states \cite{ecoffet2021first}.} that prevails in most intrinsic motivation techniques. The core idea is to cluster states and restart the environment in the most promising cluster, so that exploration begins therefrom. Derailment was also faced differently in \cite{yan2020exploring} by using an action-balance system that modifies the way how the action-sampling process is executed. The work in \cite{badia2020agent57} avoids these problems by setting two degrees of novelty (intra and inter episodic).

The success of intrinsic motivation techniques has catalyzed their adoption in a wide range of RL problems \cite{taiga2019benchmarking,successexplorationenvs2019}, surpassing beyond the single-agent domain. Indeed, these techniques have also been adopted in scenarios with multiple RL agents, which is even more challenging as the actions executed by an agent could influence the observation/state space of other agents. This term (\emph{influence}) was actually introduced in \cite{jaques2019social} by proposing the generation of a causal influence reward through KL-divergence between the policies influenced and not by that agent's action selection. This causal influence is used to increase the reward signal obtained from the environment. Likewise, in \cite{wang2019influence} a follow-up study is presented where, instead of computing the influence with policy distributions, the agent's transition probability function is used to measure the actual impact of one agent's policy over another. More recently, \cite{chitnis2020intrinsic} proposed rewards that encourage having synergistic effects into changing the environment.

Interestingly, little research has been done in multi-agent curiosity assuming that the scenario does not change due to the agents' decisions (without measuring the influence). This scarce activity may be also due to the lack of a consistent hypothesis to have good results. In this context, \cite{iqbal2019coordinated} analyzed the effect of using decentralized curiosity modules and then combining them with different methods. An ablation study was done for different tasks and scenarios and proposing its own non-linear method. In \cite{bohmer2019exploration}, a intrinsically rewarded centralized agent was proposed, which is linked to the decentralized agents as they share the replay buffer from which they collect the training experiences. This work showed that, although decentralized agents are not rewarded with intrinsic bonuses, they improve the exploration process. Moreover, if decentralized agents were encouraged with intrinsic motivation directly, their results got worse. The impact between individual and joint curiosity was examined in \cite{schafercuriosity}, arriving at the conclusion that the latter has more stability with similar results, although the sparse reward frequency is much higher than in our work, and also exhibits different types of observations.

\subsection{Transfer Learning} \label{ssec:drl_transfer}

Transfer Learning aims to accelerate the learning process through knowledge reusability. To this end, it is mandatory to have a connection between the two learning activities \cite{zhuang2020comprehensive}. The success will strongly depend on selecting \texttt{what, when and how} to manage the knowledge from a source to a target, as the transferred knowledge does not always have the desired positive impact. Even more important is to minimize the possible negative transfer \cite{wang2019characterizing}.

Depending on the application niche, the taxonomy and terms used for transfer learning can vary. On one hand, homogeneous or heterogeneous transfer learning depend on whether the source and target domains are equivalent \cite{day2017survey}. Similarly, the study in \cite{da2019survey} distinguishes between intra- and inter-agent knowledge transfer when the agents have different sensors and possible internal representations. 

As has been advanced in the introduction, this work assumes different action spaces between agents, hence, we deal with heterogeneous agents and inter-agent transfer, as their solution spaces can be different from each other. This is the case of \cite{calvo2018heterogeneous}, where the junctions of a city are not equal and they train them separately as the information is not directly reusable; and also that in \cite{zolna2019reinforced}, where they develop a strategy based on how to reuse experiences collected from an expert when there are differences respect to the student's action domain, potentially leading to different optimal solutions.
In this line of research most studies gravitate on the \textit{teacher-student} framework, wherein the teacher is an agent that has previously gained knowledge and who gives instructions to the student \cite{da2020agents}. However, this procedure requires a two-step learning process: first we need to train an agent -- or equivalently, obtain an expert that can be a human supervisor -- and afterwards, transfer the acquired knowledge to the other agent  \cite{rusu2015policy,parisotto2015actor}. This two-step process can be circumvented via online knowledge transfer in several ways. One possibility is to utilize feature representations \cite{zhu2020transfer}, so that it is possible to reuse representations even if afterwards the reward function changes. Another method is to share parameters of the same network for a better generalization and robustness against noise \cite{song2018collaborative}. Attention mechanisms have been also combined with parameter sharing so as to have more diverse targets and enhance the learning convergence \cite{chen2020online}. Ensembles of students that learn collaboratively and teach each other during the train process have been also proven to effectively model training experience in the form of posterior distributions \cite{zhang2018deep}. Furthermore, dual policy distillation between students has been proposed in \cite{lai2020dual} to prioritize distilling disadvantage states from the peer policy to explore different aspects of the environment. Nevertheless, all these studies were carried out by taking into account homogeneous students. Consequently, the application of these advances to heterogeneous agents still remains uncharted.

\subsection{Contribution beyond the State of the Art} \label{ssec:contribution}

One of the most distinctive characteristic of our work with respect to the current literature is that we assume two agents deployed over identical environments with different action domains, where the dynamics of the environment change according to action effects. In this sense, the most related works to handle such action heterogeneity are \cite{calvo2018heterogeneous,zolna2019reinforced}, but they do not use intrinsic motivation to deal with exploration issues. Moreover, the first addresses the heterogeneity dilemma by training all agents independently from each other. In the second work, however, the state space is not influenced by the action-domain differences, and the advantage of using some actions or others hinges on the fact that, in nature, some actions available in one of the agents help achieve the goal faster, but imprinting no changes in the environment. In addition, this approach needs for an expert.

Another point of novelty regards how to reuse the aforementioned low-level information declared to be good for its transfer across closely related tasks. As depicted in \cite{da2019survey,taylor2009transfer}, \textit{what to share} is not trivial (i.e., instances, value functions or action advice, among others), and its selection strongly depends on different constraints imposed by the environment and the algorithm to be used. As will be explained later, we propose to share the value function $V(s)$ in order to improve the speed in the learning process and obtain better value estimates for the involved agents. We study the usefulness of using recurrent neural layers, where the weights are shared but not the hidden states, which are subject to the skills of each agent. 
\begin{figure*}[ht]
    \centering
    \subfloat[]{\includegraphics[width=0.42\columnwidth]{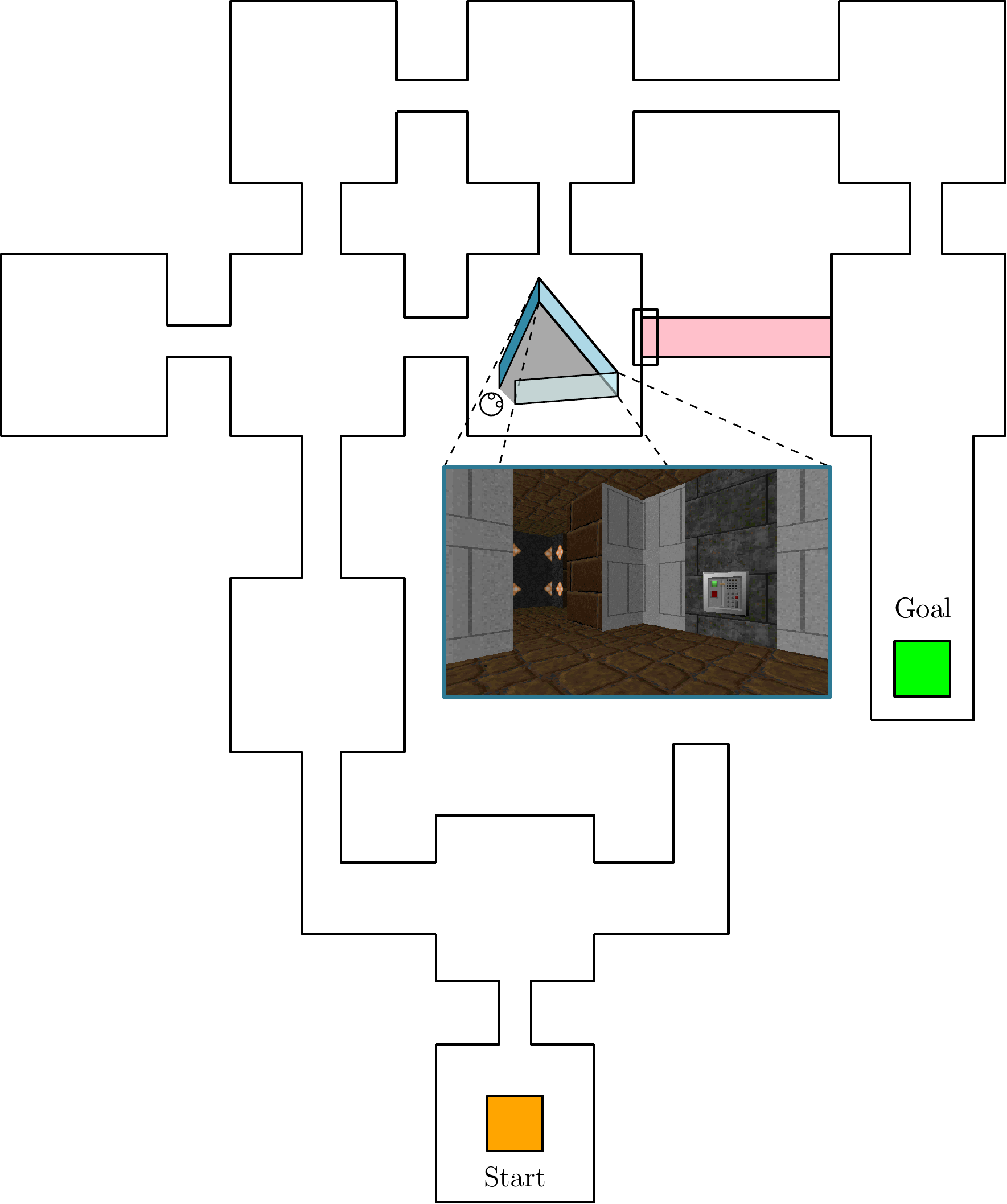}} \hfill
    \subfloat[]{\includegraphics[width=0.42\columnwidth]{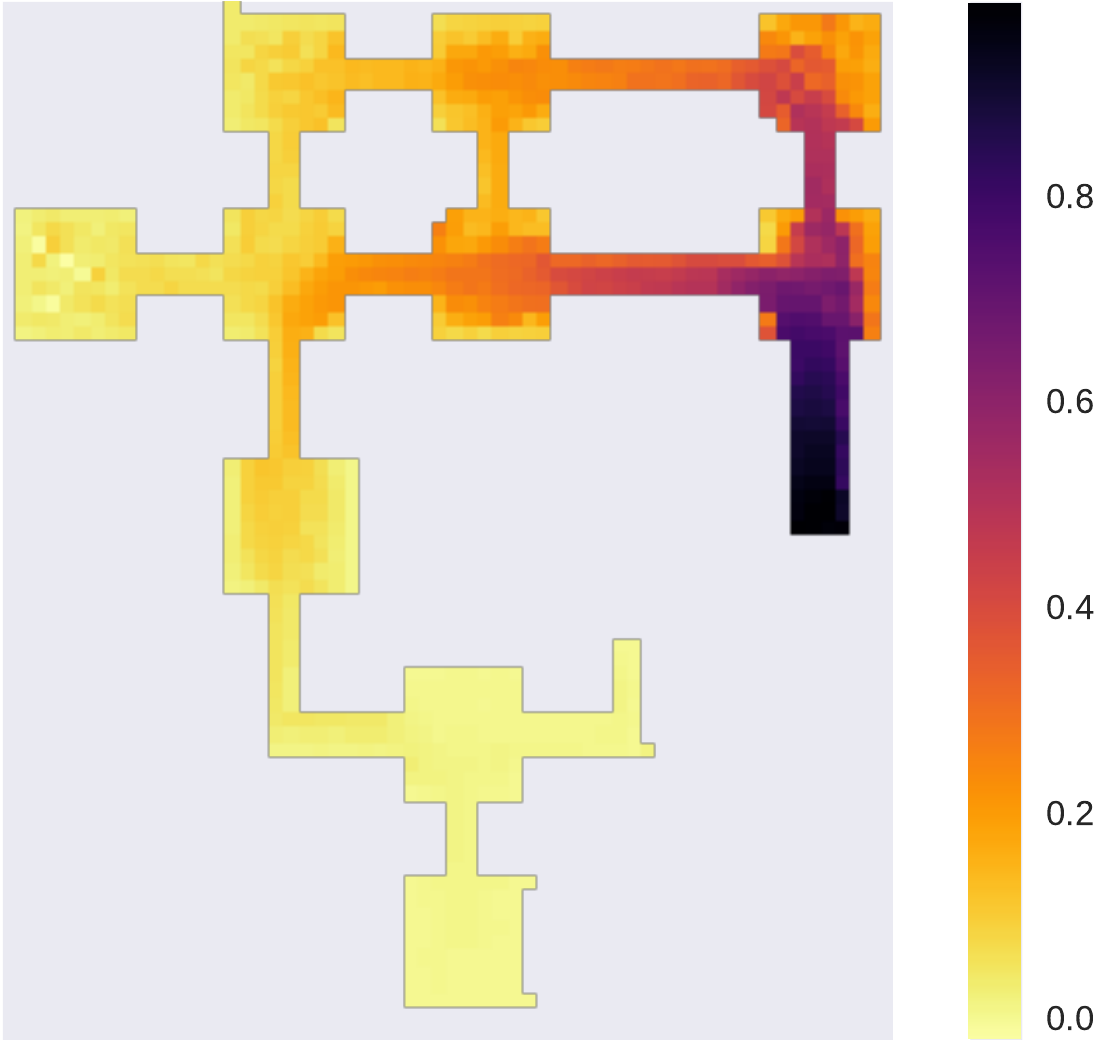}}\hfill
    \subfloat[]{\includegraphics[width=0.42\columnwidth]{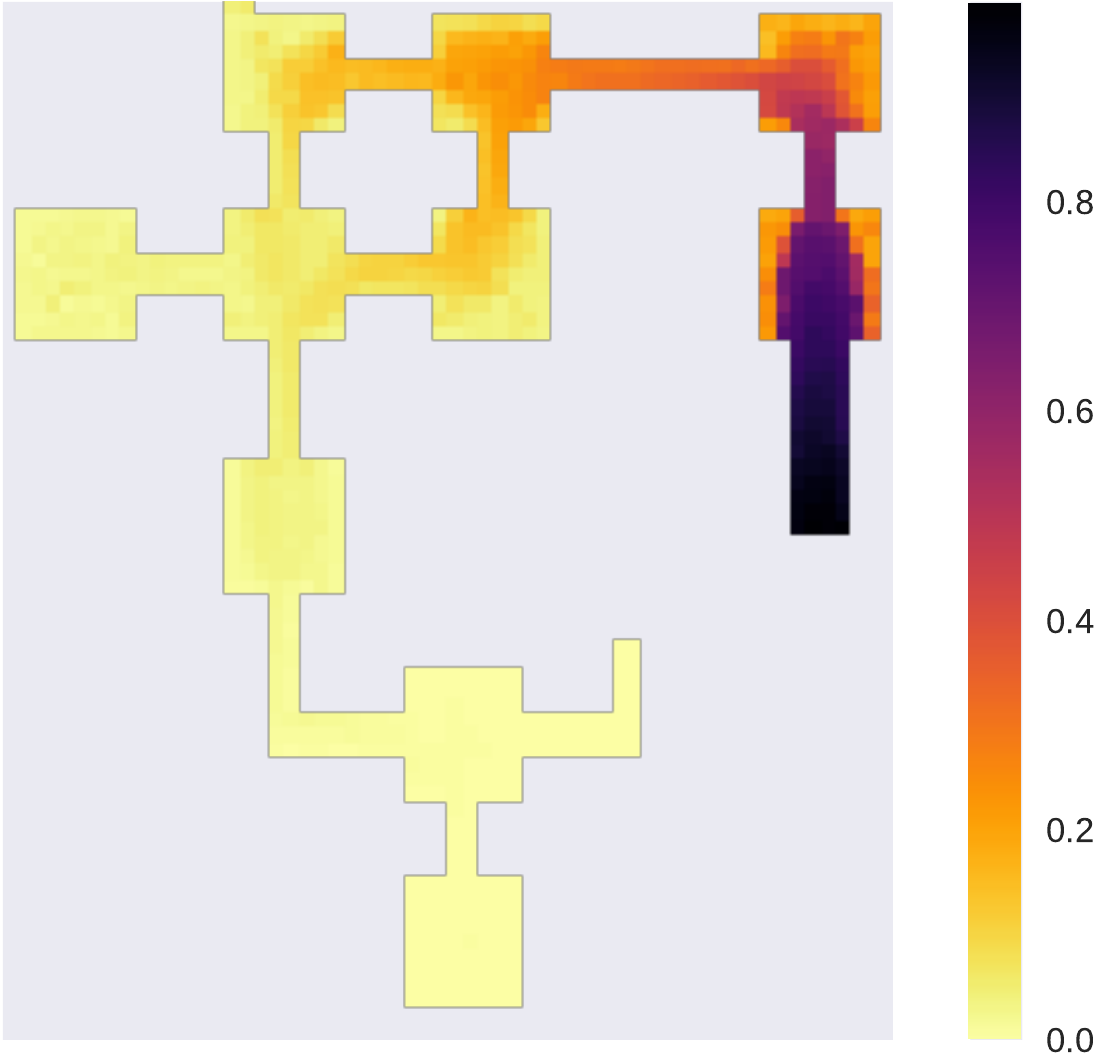}}\hfill
    \subfloat[]{\includegraphics[width=0.5\columnwidth]{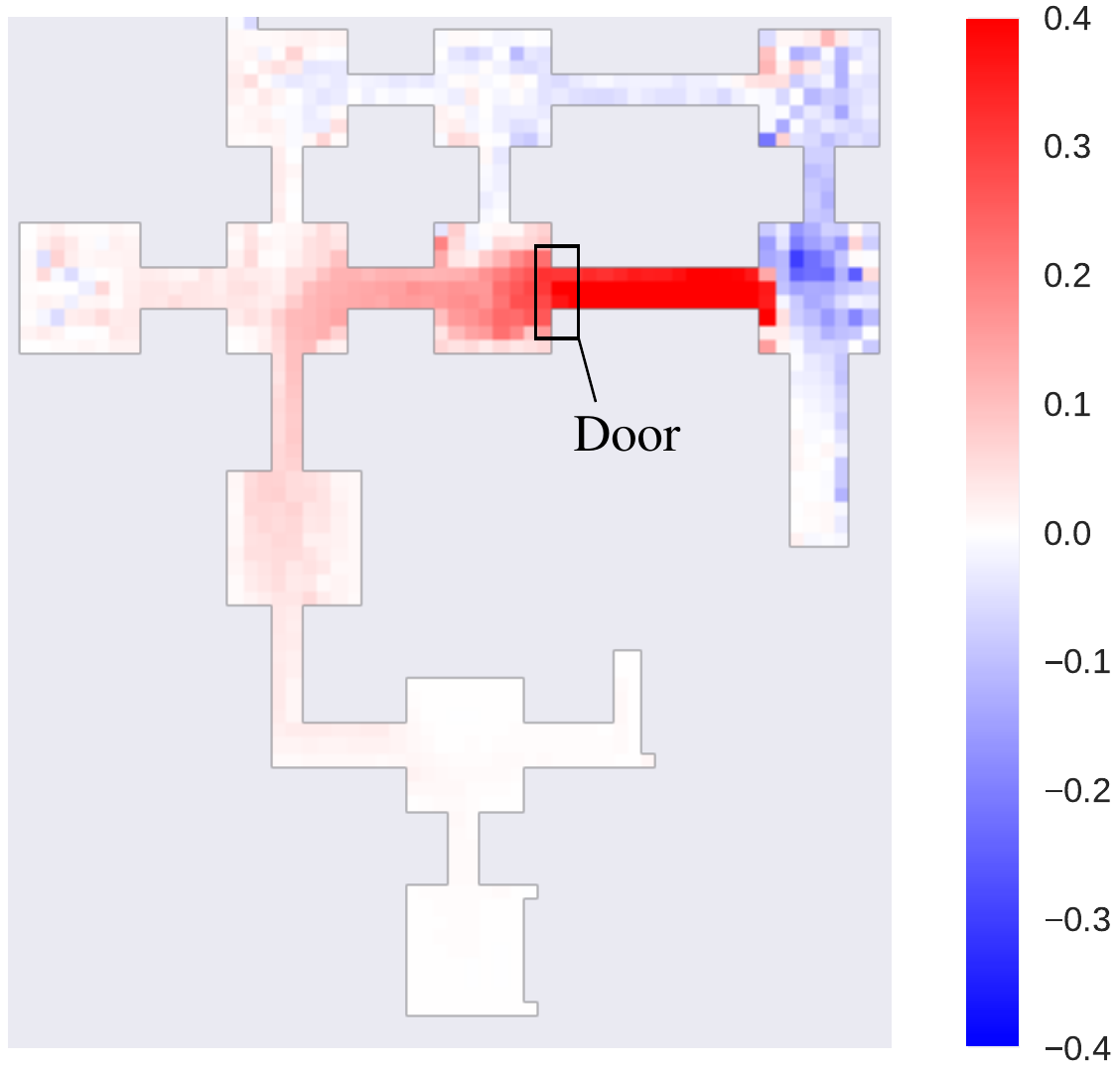}}
    \caption{State-value estimates over (a) a modified ViZDooM's \emph{My Way Home} scenario for a skilled agent that (b) can and (c) cannot open a door that gives way to a corridor. (d) Divergence map between the state-value estimates of both agents. Locations where the skilled agent has better value estimates are highlighted in red, whereas in the spots colored in blue the other agent has better estimates. In those locations where the policy distributions are supposed to be similar, the value estimates are almost white, while differences increase when their optimal paths diverge.}
    \label{fig:heatmap_workers_problem_statement}
\end{figure*}

Furthermore, we clarify with informed evidence the effectiveness of different exploration strategies from the perspective of intrinsic rewards, where the choice and sharing of different parameters may have a big impact when having heterogeneous action spaces. More importantly, we analyze the impact of considering not only the state $s$, but also the action $a$ as well. It is worth mentioning that in single-agent problems, considering the action $a$ in the mechanism used to induce intrinsic motivation has been reported not to provide any additional advantage \cite{tang2017exploration}. Differently than in our work, multi-agent problems with sparse rewards have been approached mainly by assuming that decisions taken by one agent influence anyhow the environment they share. In other words, they learn from the same instance of the environment. Given the assumptions made in our setup, where the agents run over independent copies of the same environment, it is not clear how to design an effective intrinsic motivation strategy.

Last but not least, we highlight that none of the agents has any prior knowledge about the task and the environment, and all depart from the same level of intelligence. All this without losing focus on the main objective: \textit{to expedite the learning process of the involved agents with respect to performing it independently from each other}.

\section{Problem Statement} \label{sec:problem_statement}

The exploration-exploitation dilemma is not new in RL. Using intrinsic motivation as a signal to encourage exploration unleashes new challenges: i) \textit{how to generate the intrinsic reward}; and ii) \textit{how to correctly balance between extrinsic and intrinsic rewards}. As already seen in Section \ref{sec:related_work} there are many different options to generate such exploration bonuses, each with their advantages and drawbacks. Moreover, establishing a proper balance between the intrinsic bonus and the extrinsic rewards returned by the environment is not easy to accomplish, being a central matter of research in several works \cite{badia2020never,badia2020agent57}. These problems are even more difficult to be tackled in multi-agent settings, where the exploration of a given state can be subject to other agents' behavior. Consequently, an important issue in those problems is \textit{how and when should one agent's exploration information be combined with other agents' working in the same environment}.

The problem tackled in this manuscript is neither a single-agent nor a multi-agent problem, but requires accounting for the problems related to both scenarios. Agents learn how to interact with the environment at the same time, and share knowledge in an \texttt{online} fashion. As such, although both agents have equal but different copies of the same environment, the training process of an agent influences the other and vice versa. What is more, our agents are heterogeneous because they have slightly different action spaces that make them diverge in their optimal solution space. Hence, how to share the knowledge is not clear as it can lead to no advantage or even to negative transfer. In summary, the problem can be stated as follows: \textit{How can we share information between heterogeneous agents, avoiding negative transfer and achieving a better convergence in comparison to the case where no knowledge is shared among them?} 

In order to illustrate the issues arising from the problem under study, we depict an example over the modified ViZDooM's \emph{My Way Home} RL scenario depicted in Figure \ref{fig:heatmap_workers_problem_statement}.a. In such an scenario, a door is placed at a point in which, if made open, the entrance to a corridor is enabled making the path to the target $20$\% shorter than going through elsewhere. We assume two differently skilled agents: one can open the door and access the corridor thanks to a certain action it can perform exclusively, whereas the other lacks this action and hence is not allowed to open the door nor to traverse the corridor. Due to the different action spaces, their optimal path solutions diverge. However, a unique reward is only provided when reaching the final goal. Therefore, there is no extrinsic signal that provides the agents with information to ascertain whether they should share information, when to do it and how. As a result of pursuing the same objective, this could lead one agent to be pushed through a path that can never be accomplished, or through a solution that is far from its optimal trajectory towards the target. 
\begin{figure}[h!]
    \centering
\begin{tabular}{c}
     \includegraphics[width=0.8\columnwidth]{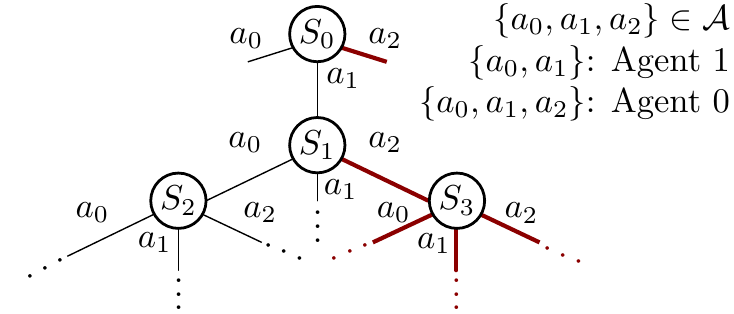} \\
     (a)\\
     \\
     \includegraphics[width=\columnwidth]{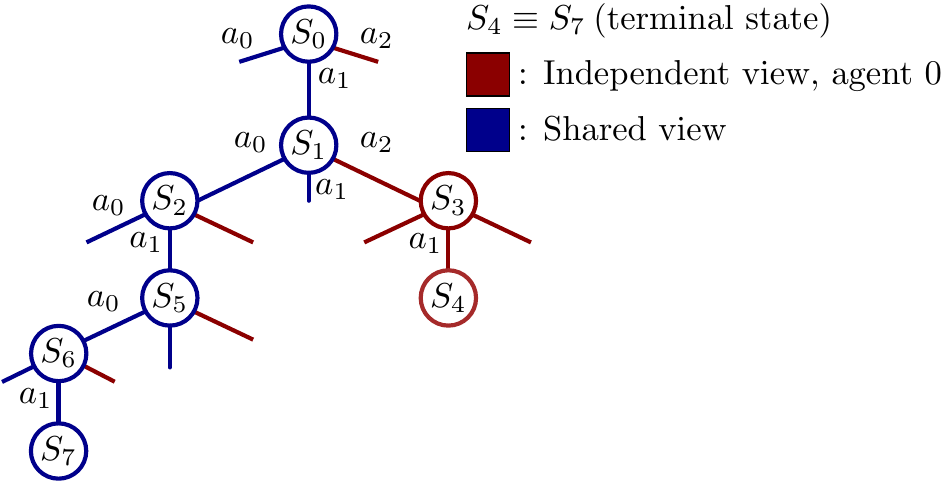} \\
     (b) 
\end{tabular}
    \caption{Example of a Markov Decision Process as a tree where nodes are states and edges denote actions that lead to another state that expands the tree. The tree can be completely visited if the agent is capable of executing all the actions. (a) represents the case where both agents possess different action spaces (some tree parts cannot be explored by one of them); (b) shows common parts shared between agents (shared-view) and how their optimal paths diverge at some point (independent-view). For the sake of clarity, here each node represents a unique state independently of the timestep. Hence, one state (i.e. $S_4$) which is apparently impossible to be reached by one of the agents (i.e. A1) might be tractable from other given state.}
    \label{fig:mdp_tree}
    \vspace{-4mm}
\end{figure}

By training those agents independently, they are able to achieve the goal through their respective optimal paths. Nevertheless, when training in a collaborative way and as a consequence of those different optima and state spaces, the agent reaching the target first will encourage the other to follow its steps. In that sense, both agents can achieve the goal by chance although the non-skilled agent may get easier due its reduced search space. This implies a two-sided fight where the non highly-skilled agent drives the skilled one into a longest path solution and the highly-skilled agent pushing the former to take the shortcut. Consequently, a negative transfer problem may arise. This situation can be deduced from their value estimates which, as shown in the plots nested in Figure \ref{fig:heatmap_workers_problem_statement}, differ remarkably from each other. 

These issues can be understood even clearer if the problem is represented as a Markov Decision Process (MDP) tree (Figure \ref{fig:mdp_tree}). The agent that lacks the action needed for opening the door is not able to accomplish the expected path through the corridor either. Hence, it has another optimal independent view from one point (one state) onward. Moreover, decisions/actions that are strengthened in each state may influence the other agent as described before, leading to negative transfer by either non-reproducibility or pushing across suboptimality.

These problems are not limited to the example shown in the plot, but also to any scenario where agents own distinct action spaces which yield different optimal paths. Due to these unknowns, our research aims to expose this problem and sketch effective collaborative learning strategies under such circumstances.


\section{Preliminaries}\label{sec:preliminary}

Before delving into the framework proposed to deal with the above problem, some core ideas needed for a proper understanding of the framework and a soft introduction to notation and concepts later referred are now posed: RL as a partially observable MDP (POMDP, Subsection \ref{ssec:pomdp}), Proximal Policy Optimization (PPO, Subsection \ref{ssec:ppo}) and intrinsic rewards (Subsection \ref{ssec:intrinsic_rewards}).

\subsection{RL as a POMDP} \label{ssec:pomdp}

In essence, RL aims to learn an agent capable of performing a task through its interaction with an environment over which the task is defined. At each time step $t$, the agent perceives the scenario as a state $s_t \in \mathcal{S}$, and executes an action $a_t \in \mathcal{A}$, which yields a reward $r_t$ and a transition to a new state $s_{t+1}\sim{P(s_t,a_t)}$ based on the transition probability function of the scenario. This problem formulation is usually formalized as a MDP, which consists of a tuple $\langle\mathcal{S},\mathcal{A},\mathcal{R},\mathcal{P}\rangle$.

However, in some cases the state variable cannot summarize completely the information of the environment, as the agent can only perceive part of the underlying system state. Hence, the Markov property is not fulfilled. In those cases, the problem is rather framed as a Partially Observable MDP (POMDP), in which the agent collects observations $o_t \in \Omega$ rather than the true state $s_t$; thus, a POMDP tuple is composed by the tuple $\langle\mathcal{S},\mathcal{A},\mathcal{R},\mathcal{P},\mathcal{O},\Omega\rangle$, where $o \sim O(s)$ is the probability distribution of observations. 

The result of the learning is always a deterministic or a stochastic policy that maps each observation to a given probabilistic distribution over the action space, $a_t \sim \pi$. In order to discover the best possible policy, the agent is trained to maximize the expected discount return: 
\begin{equation}
    \label{eq:expected_discounted_return}
    G_t = \mathbb{E}\left[\sum_{t=0}^{\infty} \gamma^t r_t \right],
\end{equation}
where $\mathbb{E}[\cdot]$ denotes expectation, and $\gamma \in [0,1]$ is the discount factor that is used to weight the importance of immediate rewards over their long-term counterparts. The way in which the obtained policy is evaluated is usually based on the average extrinsic rewards/returns obtained through an entire episode.

\subsection{Proximal Policy Optimization} \label{ssec:ppo}

Proximal Policy Optimization (PPO) \cite{schulman2017proximal} is a policy gradient method that collects data on-policy; thereby, it only operates with data that are representative with the current policy state. Commonly, policy-gradient algorithms calculate their gradient estimator as:
\begin{equation}
\hat{g}=\hat{\mathbb{E}_t} [\nabla_\theta log \pi_\theta (a_t\vert s_t) \hat{A_t}], 
\end{equation}
where $\pi_\theta$ refers to a stochastic policy and $\hat{A_t}$ is an advantage estimator at time step t. The estimator $\hat{g}$ is obtained by differentiating the following loss function:
\begin{equation}
L^{PG}(\theta)= \hat{\mathbb{E}}_t \left[\log \pi_\theta (a_t\vert s_t) \hat{A_t} \right].
\end{equation}

However, using the same batch of experiences only once is not sample-efficient; conversely, using it to perform multiple optimization steps can lead to destructive policy updates. PPO was created to take the biggest possible improvement step in the policy without moving so far that the performance of the policy can eventually collapse. With the purpose of avoiding large policy updates, the latter is updated with a clipped function that prevents it from moving too far away from unity under a given probability ratio $\Gamma_t(\theta) = \pi(a\vert s)/\pi_{old}(a\vert s)$, i.e.:
\begin{align}
    &L^{clip}(\theta)\!=\! \mathbb{E}\! \left[\min \{\Gamma_t(\theta)\hat{A_t}, \mbox{clip}(\Gamma_t(\theta),1\mbox{-}\epsilon,1\!+\!\epsilon) \hat{A_t} \}\right]\!
    \label{eq:ppo_clip}
\end{align}

Another important component of the above formula is the advantage estimator $\hat{A_t}$, which can be calculated in different ways. In the case of the proposed framework, we opt for a variance-reduced technique coined as Generalized Advantage Estimation (GAE, \cite{schulman2015high}), which ensures a good trade-off between bias and variance.

\subsection{Intrinsic Rewards} \label{ssec:intrinsic_rewards}

The generation of intrinsic rewards can be designed as per multiple strategies and agent architectures, where their knowledge acquisition can be subject to prediction errors, state novelty or information gain \cite{aubret2019survey}. There are different ways in which the intrinsic reward $r_i$ can be defined through a function as described in Section \ref{sec:related_work}. For instance, when using \emph{visitation counts} a common choice is to generate the reward as $1/\sqrt{N(s)}$, where $N(s)$ computes the number of times a state $s$ has been visited \cite{bellemare2016unifying,ostrovski2017count}. 

One of the main challenges when dealing with intrinsic rewards is how to blend those rewards with the extrinsic ones obtained from the environment ($r_e$). A widely embraced workaround is to combine them directly as a linear sum at a given time step $t$, using a parameter $\beta$ to tailor the importance granted to each reward term:
\begin{equation} \label{eq:reward_composition_formulation_1}
r_t = r_{t_e} + \beta r_{t_i}.
\end{equation}

Another alternative is the one presented in \cite{burda2018exploration} where, instead of performing a sum directly over the reward functions, the combination is made at the advantage level after separately calculating their respective value estimates. This implies that both extrinsic and intrinsic streams are treated with their respective independent rewards. As depicted at \cite{burda2018exploration}, we can combine them through the next operator
\begin{equation} \label{eq:advantages_mix_calculation}
A(s,a) = A(s,a)_{ext} + \beta A(s,a)_{int},
\end{equation}
i.e, as a weighted combination of the advantages associated to each state-action pair. This simple change allows for a higher flexibility to mix episodic and non-episodic returns, and also enables the use of different discount factors. Moreover, it is intuitively more suitable to separate both streams that are indeed stationary and non-stationary in nature. Following \cite{burda2018exploration}, observations are normalized, and so are intrinsic rewards by dividing them by the standard deviation of the discounted intrinsic returns.

\begin{figure}[h!]
    \centering
    \includegraphics[width=\columnwidth]{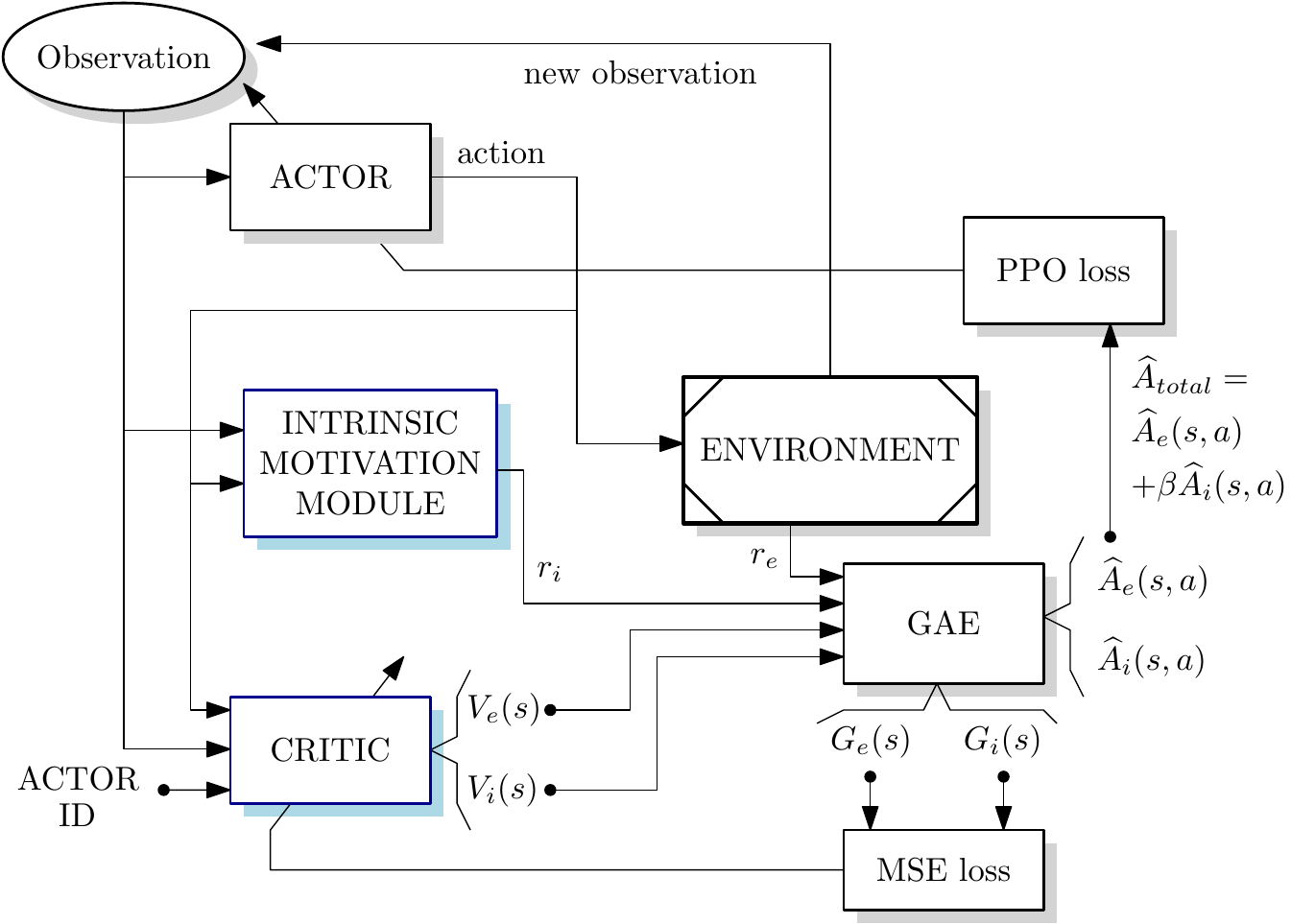}
    \caption{Flowchart of the collaborative framework, where those modules that are usually performed independently for each agent, and that can be shared in our framework, are highlighted in blue.}
    \label{fig:architectura_bloques}
    \vspace{-5mm}
\end{figure}

Intrinsic rewards explained heretofore could be part of the flowchart of our proposed framework when addressing a single-agent scenario. This flowchart is depicted in Figure \ref{fig:architectura_bloques}. However, when considering two or more agents (which is our identified problem), these steps can slightly change. 

We can assume that each agent can have its own decentralized intrinsic reward module in charge of computing their intrinsic rewards. Afterwards, those values can be mixed together through an operator $G(\cdot, \cdot)$, which can be used to combine and balance each reward (i.e. addition, variance-based). This leads to an update in Expressions \eqref{eq:reward_composition_formulation_1} or \eqref{eq:advantages_mix_calculation}:
\begin{equation} \label{eq:reward_composition_formulation_2}
    r_{t}^0 = r_{t_{e}}^0 + \beta \left[ G \left( r_{t_{i}}^0,r_{t_{i}}^1 \right) \right],
\end{equation}
where $r_t^0$ ($r_t^1$) denotes the total reward of agent $0$ (corr. 1) at time step $t$, and $r_{t_i}^0$ ($r_{t_i}^1$) the intrinsic motivation for the observation at that moment. In this way agents $0$ and $1$ explore the environment independently from each other, but they combine the gathered information to get a joint intrinsic reward \cite{iqbal2019coordinated}.

On the other hand, we can have just a single intrinsic reward module for all the agents (\emph{centralized curiosity}), which is trained by taking into account all the collected experiences. This minor change induces a faster decrease of the novelty\footnote{This decrease can be slowed to render a similar behavior by using a update proportion equivalent to the number of agents/workers.}. Moreover, those experiences used are going to be more diverse by the virtue of being collected by diverse agents with their own action distributions (in the case of two agents, $\pi_0$ and $\pi_1$). Thus, a centralized curiosity module is indirectly biased to the policy distributions of all the involved agents over time, whereas the decentralized curiosity is biased only to their associated agent. In the present work we use the independent approach when referring to the generation of decentralized intrinsic rewards (referred as the \emph{independent approach}), whereas \emph{centralized curiosity} will stand for the case when the intrinsic motivation of agents are shared and combined together. Note that both cases relate to how intrinsic rewards are computed and updated over the curse of the agents' training.


\section{Proposed Framework}\label{sec:methodology}

The design of our framework roots on the fact that there can be observations where the policy distributions of heterogeneous agents can be very similar to each other.In some cases, both agents can push towards the same direction, i.e. $\pi_0 \equiv \pi_1$. However, in some places of the environment those distributions can differ from each other because each agent pushes in a different direction based on their optimal solution. In this envisaged situation, we aim to strengthen the shared knowledge between both of them, yet at the same time, to avoid negative transfer in places where the optimal solutions of each agents become in conflict. Consequently, the goal of the framework is to learn a shared-knowledge view while respecting those subspaces in the environment where the interest of agents are not the same.

As we have already explained in previous sections, in problems characterized by sparse rewards the main issue to deal with is an efficient exploration of the environment. The application of intrinsic motivation and on-policy techniques does not permit to interfere in the action-sampling process directly, as the training experiences have to be representative of the current policy, i.e., $a \sim \pi(s)$. Hence, the use of past experiences or even samples collected by other policies are not tractable\footnote{Not tractable at least theoretically without any type of correction such as importance sampling.}. In this case, the policy is optimized as per Expression \ref{eq:ppo_clip} where, aside from the inherent mechanism of the algorithm itself, the advantage estimator $\widehat{A_t}$ is the main factor that eases and pushes the learning. The latter advantage estimator can be estimated in different ways, but almost all of them are correlated to the reward and the value function through the TD-error:
\begin{equation}
\delta = r_t + \gamma V(s_{t+1}) - V(s_t),
\end{equation}
whose value changes iteratively as soon as $V(s)$ gets updated for other states. This process can be said to converge when $V(s)$ represents the true value estimate of state $s$. However, the speed at which this is achieved depends on multiple factors. All this coupled with the fact of transient intrinsic rewards and sparse extrinsic feedback, raised the importance of introducing Monte Carlo updates to latch on to these signals rapidly \cite{bellemare2016unifying,ostrovski2017count}. In our framework, this is instead circumvented by using GAE \cite{schulman2015high}.

The framework described in what follows aims at accelerating the learning process focusing on the exploration part, more concretely in how to generate better advantages. For that purpose, we propose a framework driven by two different design objectives (DO):
\begin{itemize}
    \item DO1: How to generate better and faster value estimates $V(s)$.
    \item DO2: How to modify the intrinsic reward generation process to be tackled more efficiently when dealing with heterogeneous action-spaces.
\end{itemize}

Next, we propose methods to answer the above questions, so that the ablation study discussed in Sections \ref{sec:experiments} and \ref{sec:results} informs about the best performing approach as per the experiments carried out in this study.

\subsection{Centralized Critic Module}

Our framework adopts an actor-critic policy gradient architecture with two separated networks:
\begin{itemize}
    \item An actor whose policy (for each agent) is fed just with its local observations
    \item A critic with two output heads related to the extrinsic and intrinsic signals that is trained with the observations gathered by all the agents.
\end{itemize} 

The core idea is to have a unique and centralized critic so that its capabilities can be augmented with additional information corresponding to the different agents \cite{lowe2017multi}. The reason behind is to ease the training phase but to not interfere in the final result (the policies), which at test/deploy time operate independently regardless of the information given by the critic. With this design we aim on accelerating the critic's learning to generate more accurate and faster value estimates (DO1). Moreover, it gives rise to a scalable architecture which can easily take into account more agents with little additional complexity. 

The framework makes use of a centralized critic that unites all the parameters in a single neural network. The design is based on an universal value function approximator (UVFA) \cite{schaul2015universal} that, instead of making parts of the architecture subject to the goal, we make it to be hold to the agent's capabilities. The design is inspired by the idea that the feature extraction of an observation should not be linked to an agent. However, the information gathered through a sequence (from the recurrent module in advance) might be inconsistent as the trajectories of the agents (assuming an exploration task) may diverge in some parts due to the differences in their skills.

In order to solve this inconsistency during training stage and to aid the net in gaining insights about what knowledge must be shared and what must be preserved for individual use, we give information about the skills to the network as an input. Other parameters such as the trade-off between intrinsic-extrinsic streams or the collected rewards could also be advantageous \cite{badia2020never}, but we limit the number of parameters to be analyzed to avoid over-parametrized critic architectures.  

Moreover, due to the difference in the action spaces of the agents and aiming at aiding the net to distinguished between shared versus individual views, value estimates should not only depend on the state, but also on the action taken by the agent. Hence, instead of $V(s)$, the critic used in our framework is based on an action-value function $Q(s,a)$, which will take as many outputs as the number result from the union of all actions of all agents. In this sense, some actions will be more updated than others at each state $s$, gathering better estimates and being more accurate. These changes impact on how advantages are computed, as we must consider the probability distribution of each agent to transform those output values into a single value. This is done by iterating over all actions available at agent $x$ as
\begin{equation}
    V_x(s) = \sum_{a\in\mathcal{A}_x} \pi_x(a\vert s)\cdot Q(s,a),
\end{equation}
where subindex $x$ indicates the agent for which the value estimate is computed.
\begin{figure}[h!]
    \includegraphics[width=\columnwidth]{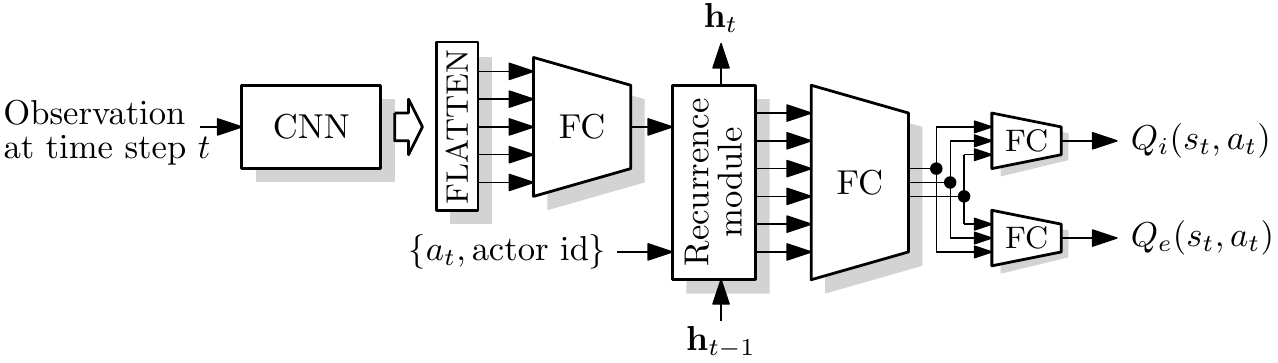}
    \caption{Architecture of the critic module. Convolutional layers extract features that are common to both agents. Then, before the recurrent layers, the predicted $Q$ values of the critic are made subject to the agents' skills.}
    \label{fig:critic_architecture}
\end{figure}

With the design of the centralized critic we aim to have a more robust and stable learning, where the shared-view value-estimates of the environment should be easier to be obtained. Furthermore, by virtue of the UVFA-based design of the critic, it manages to provide independent-view value estimates when the optimal solutions (paths) of the agents diverge.

\subsection{Centralized Intrinsic Curiosity Module} \label{subsec:centralised_curiosity}

The most straightforward strategy to make the exploration of one agent be subject to the exploration performed by others is to combine them by using a centralized module, which is directly related to the intrinsic reward generation (DO2). This idea is governed by the principle of \textit{divide and conquer}, where a observation/location should be discouraged to be visited if the other agent has already been there, promoting the exploration of uncharted areas. The problem of this assumption is that if agents have different knowledge and/or capabilities, one agent may get discouraged to explore areas that are indeed crucial for finding its own optimal solution and enforced to visit unpromising areas instead. Even if the curiosity module is intended to decrease the novelty for both agents, notice that the learning stage once the episode is over is only carried out by the collector of those experiences. The rest of the agents will make use of such information when their own trajectories get to those same states.
\begin{figure}[h!]
    \centering
    \vspace{-5mm}
    \begin{tabular}{cccc}
        \includegraphics[width=0.28\columnwidth]{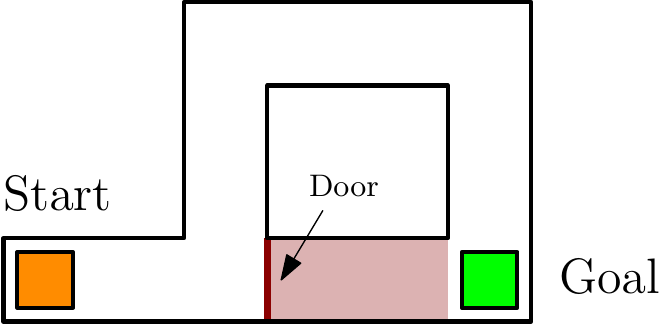} &
        \includegraphics[width=0.25\columnwidth]{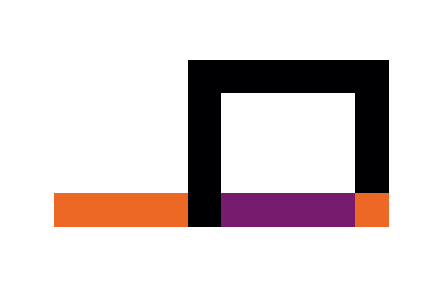} &  \includegraphics[width=0.25\columnwidth]{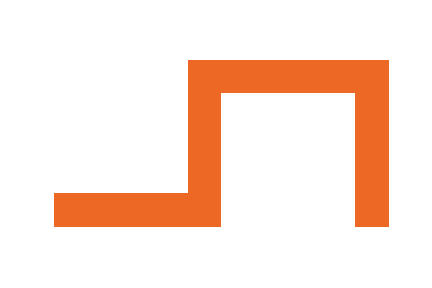} & 
        \includegraphics[width=0.04\linewidth]{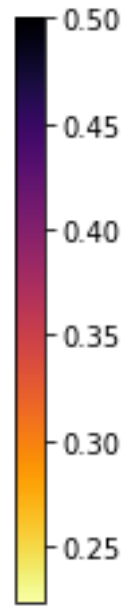} \\
        & (a) & (b) & \\
        & & & \\
        \includegraphics[width=0.25\columnwidth]{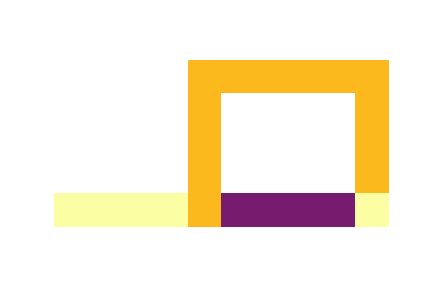} &
        \includegraphics[width=0.25\columnwidth]{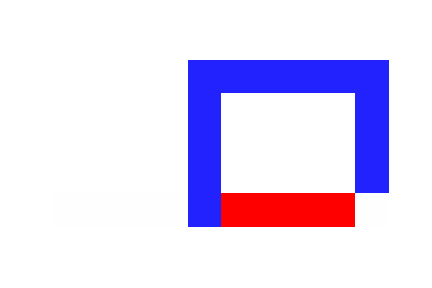} &
        \includegraphics[width=0.25\columnwidth]{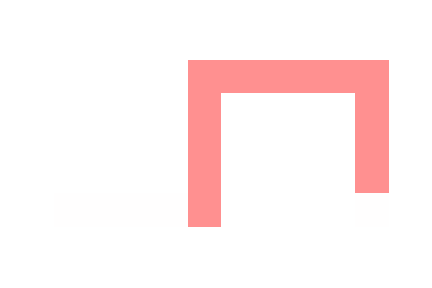} &
        \includegraphics[width=.04\linewidth]{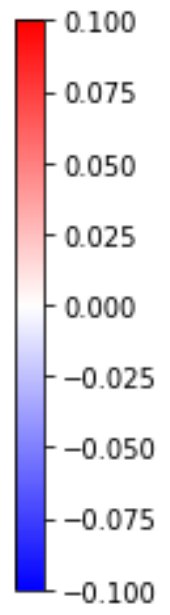}  \\
        (c) & (d) & (e) & 
    \end{tabular}
    \caption{Evolution of the intrinsic reward in a simplistic RL environment after 10 executions according to the number of visits, i.e. $r = 1/\sqrt{N(s)}$. The agent is initialized at the bottom-left corner and its goal is to arrive to the destination located at the bottom right. Going straight, in the middle is a door that obstructs the path, which can be only be opened by a skilled agent. 
    (a) Rewards of a skilled agent able to traverse the corridor through the door and go straight. (b) A non-skilled agent not capable of opening the door, hence arriving at the target through the larger path. (c) Combination of both different agents' visits in order to get the rewards for a total of 10 executions per agent (20 in total). (d) Relative difference of rewards using the centralized novelty, \textit{(c)}, with respect to using two skilled agents for the same amount of interactions. (e) Relative difference of rewards using the centralized novelty, \textit{(c)}, with respect to using two normal agents for the same amount of interactions. In (a,b,c) darker colors mean higher reward, brighter the opposite. In (d,e) red means that the centralization with heterogeneous agents encourages visiting those locations more often respect to using homogeneous agents, yielding higher rewards in that location by virtue of having heterogeneous actions (blue the opposite).}
    \label{fig:example_of_rewards_evolution}
\end{figure}

In practice, by using a centralized curiosity approach, the agents will receive less intrinsic bonuses for those states that can be reached by both agents (Figure \ref{fig:example_of_rewards_evolution}.c, yellow areas). By the same token, intrinsic returns should be higher along those trajectories in which the agent visits more novel states. Furthermore, this behavior is exacerbated in those states that are only accessible by one of the agents (skilled-agent, Figure \ref{fig:example_of_rewards_evolution}.a, corridor colored in purple), as they can only be visited by him and its novelty compared to the rest of the possible states decrease at a slower pace (Figure \ref{fig:example_of_rewards_evolution}.d, red). Therefore, the skilled agent will end up getting more encouraged to visit restricted areas when compared to the behavior in the decentralized intrinsic module approach.

When it comes to the other agent (non-skilled), these changes have little impact as the novelty distribution would indeed undergo no changes for them. Indeed, the parts leveraged by the skilled agent (in this case, the door and the corridor) do no influence the exploration of the other (Figure \ref{fig:example_of_rewards_evolution}.e, corridor). Moreover, the other observation space will be similarly visited for both agents, although if we assume that the skilled agent will be encouraged to visit more times those experiences that lead him to the corridor, inversely the non-skilled agent will be discouraged to go over those locations. Eventually, it will be pushed towards exploring other alternatives. This can be observed in Figure \ref{fig:example_of_rewards_evolution}.e, in which the non-skilled agent will be more encouraged to explore through the larger path (as told by the higher/red rewards) when combining its rewards with a skilled-agent respect to do it completely independent Figure \ref{fig:example_of_rewards_evolution}.b.

In conclusion, this approach can be directly beneficial for those observations involved in the same rollout as the observations that can only be achieved by those actions that are rarely executed (i.e. open the door and access the corridor), because they will have a higher intrinsic return.
Besides, all observations visited before in that episode are also indirectly (and positively) influenced, as the critic will learn higher value estimates for those observations close to the corridor. Eventually, those will push up in farther locations through backpropagation\footnote{This is the case when the rollout size is smaller than the episode size, and there is no extrinsic signal in the whole rollout.}. Moreover, it also discourages the agent who cannot go through those state spaces intractable for it (i.e. corridor), as the other agent is supposed to go herein, being also beneficial for the other agent to explore more promising zones for him.

\subsubsection{Action-Based Curiosity Module}
Manifold means of calculating the novelty of a given state have been proposed in the literature. Mechanisms to deal with novelty are based on using either $s_t$ \cite{bellemare2016unifying}, $s_{t+1}$ \cite{burda2018exploration} or even the information related to the transition between successive states $\{s_t,s_{t+1}\}$ \cite{pathak2017curiosity}\footnote{The intrinsic reward is generated just with $s_{t+1}$, but the update of the whole ICM architecture requires of $a,s_t,s_{t+1}$.}. In this vein, when having multiple agents using this module in a centralized manner, they update it more frequently with the experiences sampled by their own independent action distributions leading to different visitation strategies as those depicted in Figure \ref{fig:example_of_rewards_evolution}. Previous works have reported that no difference arises from considering the action \cite{tang2017exploration}, speculating that the policy itself was sufficiently random (had sufficient entropy) to entrust the exploration at each state. This hypothesis, however, was posed over RL environments with single agents whose individual exploration does not interfere with others'. 

In addition, state novelty becomes an analogous mechanism to a visiting log yielding a vague estimation of the attractiveness of the state, instead of representing the real opportunities to open new and promising paths to reach the goal \textit{through specific actions}. Concretely, the agent will be discouraged to visit states already inspected regardless the actions taken before. This is, the agent will have the same curiosity to visit a state and execute an action frequently selected (at that state) as selecting another action that has been barely chosen. Hence, the agent's critic and actor will not gain any insight into the action that should be explored. 

Therefore, we modify those intrinsic related appro\-aches, that just rely on the state/observation, in order to account for the action as well, so that the intrinsic reward generation becomes more informative for the critic (DO2). In fact, following a \textit{strategy that takes into account both the action and the state when computing the novelty will encourage a more homogeneous action selection and a deeper exploration in terms of tuples $\{s,a\}$}. This difference may not hinder convergence in single-agent problems, but can be problematic when having agents with different action spaces. In this latter case, actions that can only be executed by just one agent will become more affected.


\subsubsection{Tree Filtering} \label{sssec:filter}

Previous exploration strategies aim at sharing as much information as possible between the agents. However, there might be states embedded in a trajectory that are not accessible by some agents where specific chunks of the trajectory might, in turn, be replicable. In such cases where a given episode is not fully reproducible by the other agent, a question arises: does it make sense to share the novelty along the whole path between all the actors? Is this advantageous? Do they need indeed to reach consensus about the interestingness of all the states in the environment? Note that in this context we have 2 agents ($W_0$ and $W_1$), being both two copies in terms of skills, but one of them endowed with a supplementary sophisticated skill ($W_0$). 


On the one hand, we could deem a trajectory as a shareable trajectory for both agents if the actions selected by the agent responsible for gathering the experiences belong to the mutual action space\footnote{This also applies when selecting an action out of that mutual action space which has no effect on the environment, or which is interchangeable by one of the actions of the mutual action space.}. Hence, that agent can be conceived as a virtual runner of all those agents possessing those common skills. 
\begin{figure}[ht]
    \centering
    \includegraphics[width=\columnwidth]{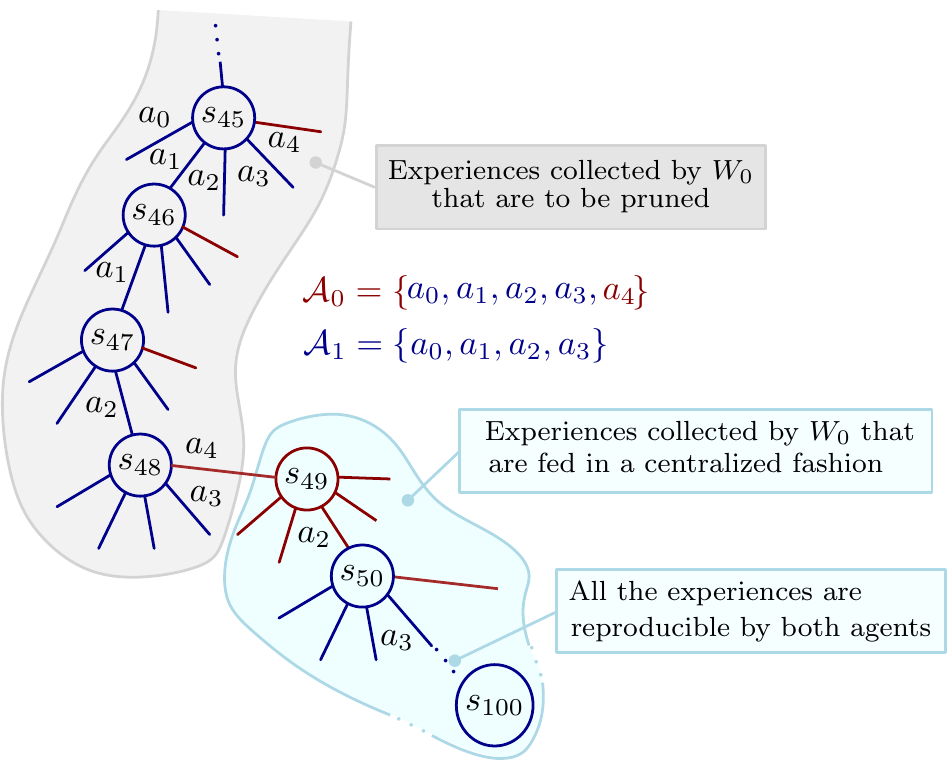}
    \caption{Diagram representing the filtering process of the exploration novelty based on action reproducibility performed by the proposed framework. At a given point, $W_1$ is not able to reproduce the experience $\{(s_{48},a_4)\}$ gathered by $W_0$. Because of that, all the experienced state-actions (shaded tree) up to that point are not taken into account for updating $W_1$'s curiosity.}
    \label{fig:Tree-Filter-by-action}
\end{figure}

On the other hand, let us consider a trajectory gathered by the sophisticated agent ($W_0$) that is not fully reproducible by the other agent ($W_1$). For the sake of clarity, we consider Figure \ref{fig:Tree-Filter-by-action}, where we distinguish several different subsequences in the displayed episode\footnote{We note that in this plot, unlike the MDP tree in Figure \ref{fig:mdp_tree}, nodes represent the sequence of visited states rather than the \emph{possible states} reachable by the agent}:
\begin{itemize}
    \item $\{(s_{49},a_2),(s_{50},a_3),\ldots\}$:
    \vspace{3mm}
    \item[] From $s_{49}$ onwards, the whole trajectory is assumed to be reproducible by agent $W_1$. Hence, the novelty of those experiences is updated for both agents. At this point, two possible situations could eventually occur:
    \begin{itemize}
        \item[$\bullet$] The non-sophisticated agent ($W_1$) is not capable of reaching those observations. Consequently, it will never be able to exploit such updated novelty, this having no effect on learning.
        \item[$\bullet$] The non-sophisticated agent ($W_1$) manages to reach those observations by its own means. Then, in the training stage, the critic will learn from those states that are less attractive, which looks like a natural procedure because a similar agent with at least the same capabilities as him has previously explored those points (pretending now that it itself was the agent who visited these states).
    \end{itemize}
    
    \item $\{\ldots (s_{45},a_1),(s_{46},a_1),(s_{47},a_2),(s_{48},a_4)\}$:
    \vspace{3mm}
    
    \item[] At state $s_{48}$, $W_0$ executed an action out of the mutual action space ($a_4$) which is not reproducible by $W_1$. Should we then decrease the novelty of all those tuples (state, action) for $W_1$? If so, that novelty reduction will be noticed when $W_1$ collects a trajectory containing any of those tuples and updates its critic. Regarding $(s_{48}, a_4)$, no impact will be made since this (state,action) is indeed impossible to be experienced in any trajectory performed by $W_1$. No\-netheless, for the rest of feasible tuples: 
    $$\{\ldots (s_{45},a_1),(s_{46},a_1),(s_{47},a_2),(s_{48},a_4)\},$$
    the intrinsic reward signal will be lowered, discouraging $W_1$ from developing its own exploration strategy on account of an external update of $W_0$ not playing the role of an equally skilled agent.
    
    
    Thus, in order to encourage $W_1$ to create its own personal experience, the novelty update of the tuples from $s_{48}$ back to the initial state is not performed on $W_1$, allowing it to keep on working on its independent individual view. 
    \end{itemize}

As a result of this filtering process, we propose a first contribution in considering \emph{novelty along paths} rather than \emph{novelty as attractiveness on isolated stepped-on states}. In practice, the novelty of a path is calculated as the discounted intrinsic return for each the experiences belonging to that trajectory, which indeed is a sum of independent intrinsic bonus as in Expression \eqref{eq:expected_discounted_return}. Here, we aim to minimize the error between the globally generated novelty estimation of paths taking into account the intrinsic rewards generated at each experience and also their reproducibility, thus polishing the intrinsic reward recollection by allowing room for independent views on the environment (DO2). Ideally, the novelty through a path would be handled by a intrinsic curiosity module that takes into account sequences rather than single experiences. However, as we will further elaborate in Section \ref{sec:discussion}, the design of such a novelty reward function is not trivial at all.

\subsection{Summary of the Proposed Framework}
To sum up, the proposed collaborative framework is composed by a centralized critic and intelligent exploration strategies, where:
\begin{itemize}
    \item The use of a centralized critic enhances the learning process by ensuring more diverse experiences while, at the same time, constructing a robust knowledge view capable of distinguishing between different types of agents and their respective action-solution domains. This is, it helps generate better and faster value estimates (DO1).
    \item The use of a centralized exploration strategy helps generate more suitable intrinsic rewards, taking into account the diversity between agents (DO2). When dealing with heterogeneous agents, we claim that the consideration of the action is necessary to generate the rewards so as to harness interesting albeit diverging behaviors throughout the experience of the agents with the environment. Furthermore, the tree-filtering strategy described previously aims to minimize undesired exploration behaviors due to the non-reproducibility of (state,action) experiences.
\end{itemize}


\section{Experimental Setup}\label{sec:experiments}

In order to assess the performance of the proposed collaborative learning framework, we now present an extensive experimental setup defined over a target finding RL task over the ViZDooM's \textit{My Way Home} environment \cite{wydmuch2018vizdoom}. As shown in Figure \ref{fig:heatmap_workers_problem_statement}.a, we have slightly modified the environment so that we can gauge the impact of having heterogeneous agents with different action spaces. The goal of these modifications is to make differences in the action spaces render different optimal solutions of the agents. 

For that purpose, we set a corridor that is closed behind a door, which can be only opened by using a certain action (\texttt{OPEN}), which is useful just when being close to the door. We further assume two agents/workers, which have $4$ actions in common: \texttt{MOVE FORWARD}, \texttt{TURN RIGHT}, \texttt{TURN LEFT} and \texttt{NO ACTION}. One of the agents ($W_0$) is endowed with the ability to \texttt{OPEN} the door, which allows for a better (\emph{shorter}) optimal path to the target. The other agent ($W_1$) can only perform the aforementioned 4 actions. Hence, a subspace of the overall environment is only reachable for $W_0$, which in turn has an immediate impact on the quality of the learned policy. At this point it is important to note that this environment differs from the one considered in our preliminary work \cite{9534146}, where the action required to access the corridor was a combination of actions (\texttt{MOVE FORWARD} and \texttt{CROUCH}). As opposed to the scenario tackled in this manuscript, the combined action used in \cite{9534146} generates slightly different observations at any point of the map due to the fact that the agent was crouched down. Consequently, it might also get advantage from just moving straight.
\begin{figure}[h!]
\vspace{-5mm}
    \centering
    \includegraphics[width=0.6\columnwidth]{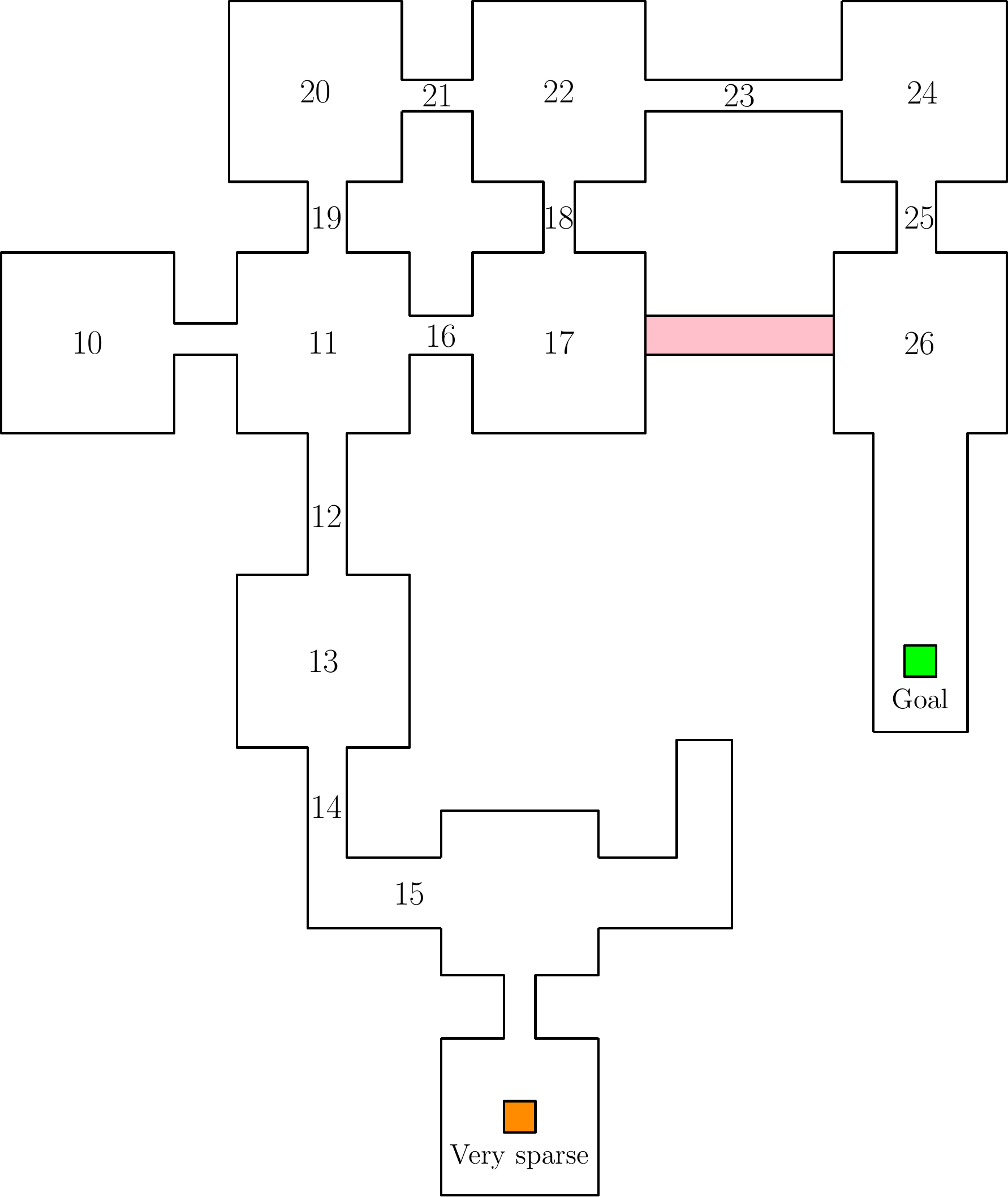}
    \caption{ViZDooM's modified \emph{My Way Home} environment with increased difficulty due to the large amount of steps that are required to obtain a reward.} 
    \label{fig:vizdoom_my_way_home}
    \vspace{-5mm}
\end{figure}

Secondly, we have increased the complexity of the most difficult known setting (\textit{very sparse}) where the number of steps to achieve the goal are estimated to be around 350 by following an optimal policy \cite{pathak2017curiosity}. In our modified environment, this number amounts to around 430 steps. Moreover, the probability of a random agent to arrive to the goal is $4.6$\% for $W_0$, where within this percentage, $60$\% of the time the agent manages to achieve the target through the corridor, this is, a random agent with $W_0$'s action space is able to achieve the goal through the corridor with a $2.8$\% of chance, and by the other alternative path with a $1.8$\%. In the case of not having the ability to open the door (worker $W_1$), its probability decreases slightly down to $3.6$\%. The reason for the modification is not only to face a more challenging scenario, but also to ensure that agents have an equal action distribution over a large part of the environment, despite having different action spaces (namely, from the initial position until room 11, as shown in Figure \ref{fig:vizdoom_my_way_home}).

\subsection{Training details}

All neural networks embedded in the agents are trained only by the observations captured from the environment, this is, with only images as inputs. The setup is similar to the one used in \cite{pathak2017curiosity,9534146}, where gray-scale observations of $42\times 42$ pixels are used. In order to address issues derived from the temporal dependency between frames, networks are fed with a representation of the current observation and the concatenation of the previous 3 frames \cite{mnih2015human}. We have also used an action-repeat factor of $4$ and a rollout size of $50$ steps, which has been proved empirically to attain a well-balanced trade-off between bias and variance for the current problem. All experiments are run for a total of $6000$ episodes, each comprising a maximum of $2600$ steps.

The use of multiple parallel environments is avoided at the cost of some instability (larger variance) in the computation of gradients. This is done for the same reasons explained at \cite{9534146}, where the use of multiple environments implies having a higher probability of arriving at the target destination and receive a non-zero extrinsic signal. Since one of our goals is to assess the differences due to different curiosity strategies, we minimize the number of parallel agents to be deployed, so that gaps can be attributed to the curiosity strategy rather than to the availability of more computational resources. Moreover, avoiding parallel environments can be also a requirement imposed in those cases when there is no chance for deploying several parallel agents (i.e. insufficient computational resources). Hence, in the results we have just used 3 parallel runners per agent, which we found out experimentally to be the minimum number of agents to yield consistent results.

Related to PPO, we employ GAE with $\lambda=0.95$ to calculate the advantages of both extrinsic and intrinsic streams, which are balanced with a relation of $3$ vs $1$, i.e., $\beta=1/3$ in Expression \eqref{eq:advantages_mix_calculation}. Moreover, we use a discount factor $\gamma$ equal to $0.99$ for both discounted returns, an epsilon value of $0.2$ for clipping, an entropy coefficient of $0.01$, a learning rate of $10^{-4}$, and $4$ epochs per training step. Finally, network architectures of the actor and critic modules are summarized in Table \ref{table:network_architectures}. 

\subsection{Evaluation Metrics}

In general, the main goal of knowledge reuse in RL is to accelerate the learning process. In order to analyze the benefits of using knowledge transfer, different metrics can be used \cite{taylor2009transfer}. However, a framework could report similar performance metrics to other possible options, but remain of interest due to other factors related to the training procedure, such as the number of required samples, the training time for a given computational power, and model complexity/size, among others.  
\begin{table}
    \centering
    \caption{Conv2D(A1,A2,B,C,D,E): Convolutional layer with A1 input channels and A2 output channels, B kernel size B, stride C, padding D and activation function E (ELU: Exponential Linear Unit)}
    \resizebox{\columnwidth}{!}{\begin{tabular}{C{1cm}L{4.5cm}L{3.2cm}}
     \toprule
     &\multicolumn{1}{c}{Network Architecture} & \multicolumn{1}{c}{Training Parameters} \\
     \midrule
     Actor & \makecell[l]{Conv2D(4,32,3,2,1,ELU)+\\Conv2D(32,32,3,2,1,ELU)+\\Conv2D(32,32,3,2,1,ELU)+\\Conv2D(32,32,3,2,1,ELU)+\\Dense(256,ELU)+\\Dense(\# actions, softmax)} & \makecell[l]{Orthogonal initialization\\Adam optimizer\\PPO loss}\\
     \midrule
     Critic & \makecell[l]{Conv2D(4,32,3,2,1,ELU)+\\Conv2D(32,32,3,2,1,ELU)+\\Conv2D(32,32,3,2,1,ELU)+\\Conv2D(32,32,3,2,1,ELU)+\\Dense(256,ELU)+LSTM(128)+\\Dense(256,ELU)+$\ldots$+\\Dense(5) [extrinsic] \& \\Dense(5) [intrinsic]} & \makecell[l]{Orthogonal initialization\\Adam optimizer\\MSE loss in both\\critic heads}\\
     \bottomrule
    \end{tabular}}
 \label{table:network_architectures}
\end{table}

Discussions on the experimental results later held in this work consider two performance scores:
\begin{itemize}
    \item \textit{Average extrinsic result} (also referred to as \emph{Success Rate}, SR), which is calculated as the average extrinsic score obtained through a window of the last 100 episodes.
    \item \textit{Number of steps to achieve the goal}, measured from the starting point of the scenario until the agent reaches the target.
\end{itemize}

The reason for considering these two scores is that, by only inspecting the SR metric, the discussion only regards whether agents have achieved the goal, disregarding the number of steps required for the purpose (which represent the quality of the learned policy). Other works using this environment assume that no rewards are given except when arriving to the goal, when they actually give a small penalization referred to as \emph{living reward}, equal to -0.0001 for each step. This small modification yields an optimal average extrinsic return of $0.97$ approximately for 270 steps. In our work we instead fix a null living reward, and give a reward equal to 1 when achieving the goal. In this way, we stand strict in regards to the sparse reward problem formulation. Moreover, the environment itself is slightly different depending on the action space of each agent. Hence, in this case $W_0$ has different possibilities to achieve the target, being optimal the one that involves going through the corridor (referred as $\_OPT$). Therefore, we track not only if every agent reaches the target, but also if they navigate through their optimal paths. 

\subsection{Experimental Ablations}

As already explained in Section \ref{sec:methodology}, the way in which both intrinsic and extrinsic bonuses are combined together can be made through the sum of both advantage streams, which can be positively influenced by either a better state value estimate $V(s)$ or by a more suitable intrinsic reward generation $r_i$. This renders different combinations of the algorithmic ingredients of the proposed framework, which are summarized in Table \ref{table:ablations} and next described:
\begin{itemize}
    \item Full Independent PPO (\texttt{IC\_IC\_3r}): the baseline PPO algorithm with independent curiosity (\texttt{IC}), independent critics (\texttt{IC}) and 3 parallel environments (\texttt{3r}).
    \item Full Independent PPO (\texttt{IC\_IC\_6r}): the baseline PPO algorithm with independent curiosity, independent critics and 6 parallel environments.
    \item Centralized Critic + Independent Curiosity (\texttt{CC\_IC}): both agents share a unique/centralized critic, but they remain independent in what refers to the generation of their intrinsic rewards.
    \item Centralized Critic + Centralized Curiosity (\texttt{CC\_CC\_sh}): both agents share all parameters of both the critic and the curiosity modules to generate the intrinsic rewards.
    \item Centralized Critic + Centralized-Action-based Curiosity (\texttt{CC\_CC\_sh\_action}): both agents share all parameters of both the critic and the curiosity modules to generate the intrinsic rewards. However, in this case the intrinsic bonus is made dependent on the state and the action, instead of just uniquely on the state. 
    \item Centralized Critic + Centralized-Action Curiosity + Tree Filtering (\texttt{CC\_CC\_sh\_action\_filter}): this sche\-me is equal to the previous one, but during the generation of the rewards it prunes those rollouts whose experiences are not reproducible by the non-skilled agent ($W_1$, see Section \ref{sec:methodology}). We discard those exploration visits just for $W_1$. For simplicity, we assume an oracle that informs whether the action executed by $W_0$ is reproducible by $W_1$. Further thoughts on the on-line estimation of these discarded rollouts will be given in Section \ref{sec:discussion}. 
\end{itemize}
\begin{table}[t]
    \centering
    \caption{Configurations of the modules and components of the proposed framework that are considered in the experiments performed in this work.}
    \resizebox{\columnwidth}{!}{\begin{tabular}{L{1.5cm}C{1.5cm}C{1.5cm}cC{1.5cm}C{1.5cm}C{1.8cm}}
     \toprule
     & \multicolumn{2}{c}{Critic} & & \multicolumn{3}{c}{Curiosity Module} \\
     \cmidrule{2-3} \cmidrule{5-7}
     & \makecell[cb]{Independent} & \makecell[cb]{Centralized} & & \makecell{Independent\\(state)} & \makecell{Centralized\\(state)} & \makecell{Centralized\\(state-action)} \\
     \midrule
     \texttt{IC\_IC\_3r} & \checkmark & & & \checkmark  & & \\
     \texttt{IC\_IC\_6r} & \checkmark & & & \checkmark  & & \\
     \texttt{CC\_IC} & & \checkmark & & \checkmark  & & \\
     \texttt{CC\_CC\_sh} & & \checkmark & & & \checkmark & \\
     \texttt{CC\_CC\_sh\_action} & & \checkmark & & & & Naive\\
     \texttt{CC\_CC\_sh\_action\_filter} & & \checkmark & & & & Filter \\
     \bottomrule
    \end{tabular}}
 \label{table:ablations}
\end{table}

In the experiments a count-based strategy as in \cite{iqbal2019coordinated} is adopted, so that the exploration bonus is made inversely proportional to the number of counts for every state, i.e., $1/\sqrt{N(s)}$ \cite{bellemare2016unifying}. To realize this, we discretize the environment space in bins based on the coordinates $(x,y)$, giving rise to a total of $30$ and $26$ bins along the $x$ and $y$ axis, respectively. When using action-based approaches, the count is made subject also to the action (namely, $1/\sqrt{N(s,a)}$), hence increasing the total number of bins. It is important to note that these bins are only used to compute the intrinsic reward, but are not fed to the policy nor to the critic as an additional source of information.


\section{Results and Discussion} \label{sec:results}

Results elicited by the experimental setup are discussed in this section. For the sake of clarity in the discussion, results are commented based on the following research questions (RQ):
\begin{itemize}
    \item RQ1: \emph{Does a centralized critic provide any gain when compared to completely independent agents?}
    \item RQ2: \emph{Does a centralized curiosity yield better performance levels than maintaining the curiosity locally at every agent?}
    \item RQ3: \emph{Should we compute curiosity incentives based on the (state,action) pair rather than only the state itself?}
    \item RQ4: \emph{Should agents get updated their intrinsic rewards only by experiences that are reproducible as per their action spaces?}
\end{itemize}

We now analyze experimental results aiming to obtain informed responses to the above questions, using to this end the different configurations of the proposed collaborative framework that are described in Table \ref{table:ablations}. The reported results are over 3 independent runs in order to account for their statistical variability. All scripts and herein shown results are released in a public GitHub repository\footnote{\url{https://github.com/aklein1995/heterogeneous_agents_curiosity_vizdoom}} to ensure reproducibility and support follow-up studies.

\subsection*{RQ1: \emph{Does a centralized critic provide any gain when compared to completely independent agents?}}

\begin{figure}[h!]
    \centering
    \begin{tabular}{c}
        \includegraphics[width=.7\columnwidth]{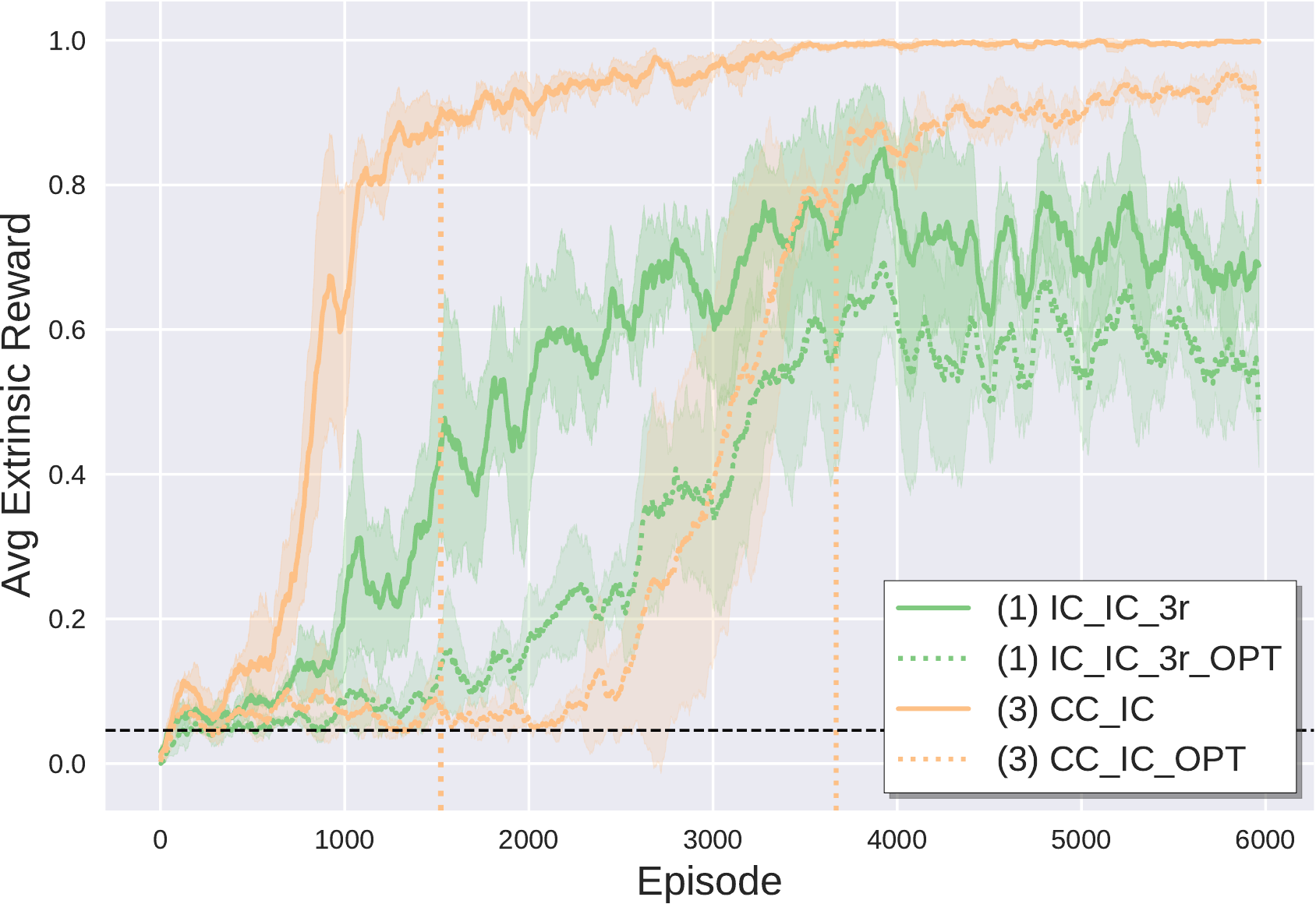} \\
        (a) \\
        \includegraphics[width=.7\columnwidth]{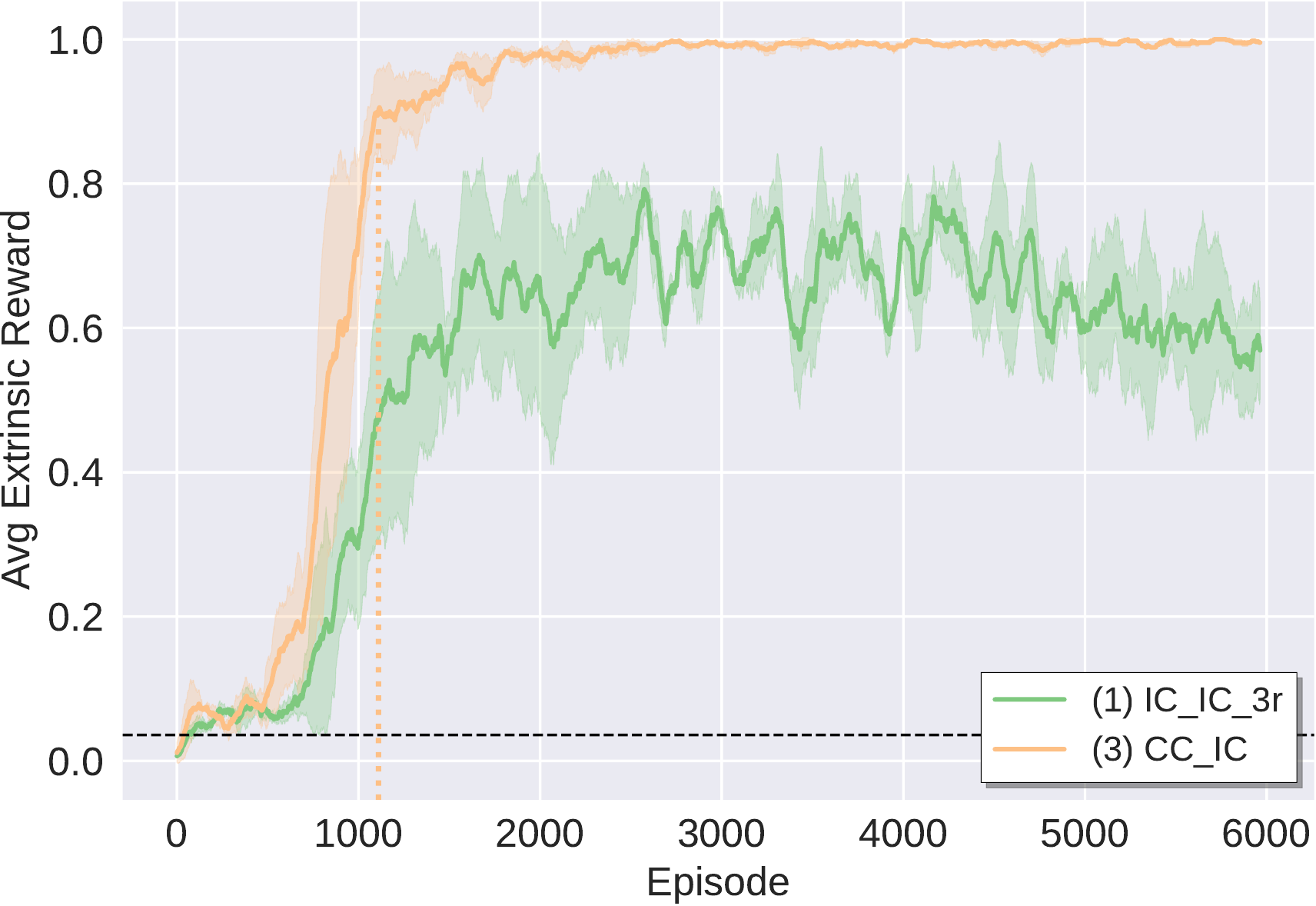} \\
        (b)
    \end{tabular}
    \caption{Success rate (SR) achieved by (a) $W_0$; and (b) $W_1$ by using an individual approach with curiosity for encouraging exploration (\texttt{IC\_CC\_3r}, green lines), compared to the case when a single centralized critic is trained for both agents (\texttt{CC\_IC}, orange lines).}
    \label{fig:icic3r_centralized}
\end{figure}

We begin our discussion by examining whether a centralized critic performs better than completely independent agents in the RL scenario under consideration. Responses to this question can be found in Figure \ref{fig:icic3r_centralized}, which evinces that a centralized critic reaches better performance levels. With a centralized critic (\texttt{CC\_IC}) both agents achieve a 100\% SR, and also guarantees that both workers reach the target through their optimal paths (SR indicated with dashed lines). By contrast, agents featuring individual critic modules (\texttt{IC\_IC\_3r}) get stacked in around 60-80\% of SR. 

Intuitively one can postulate that the advantage of using a centralized critic is that, for the same/unique neural network, more number of experiences are collected and used, as we compute the gradients with the data gathered by the two agents. If this is the case, we can just increase the number of experiences collected by each worker by doubling the number of runners used in the last experiment. By inserting this change, we ensure that the agents gather the same amount of experiences when compared to those fed to a centralized critic.

\begin{figure}[t!]
    \centering
    \begin{tabular}{c}
        \includegraphics[width=.7\columnwidth]{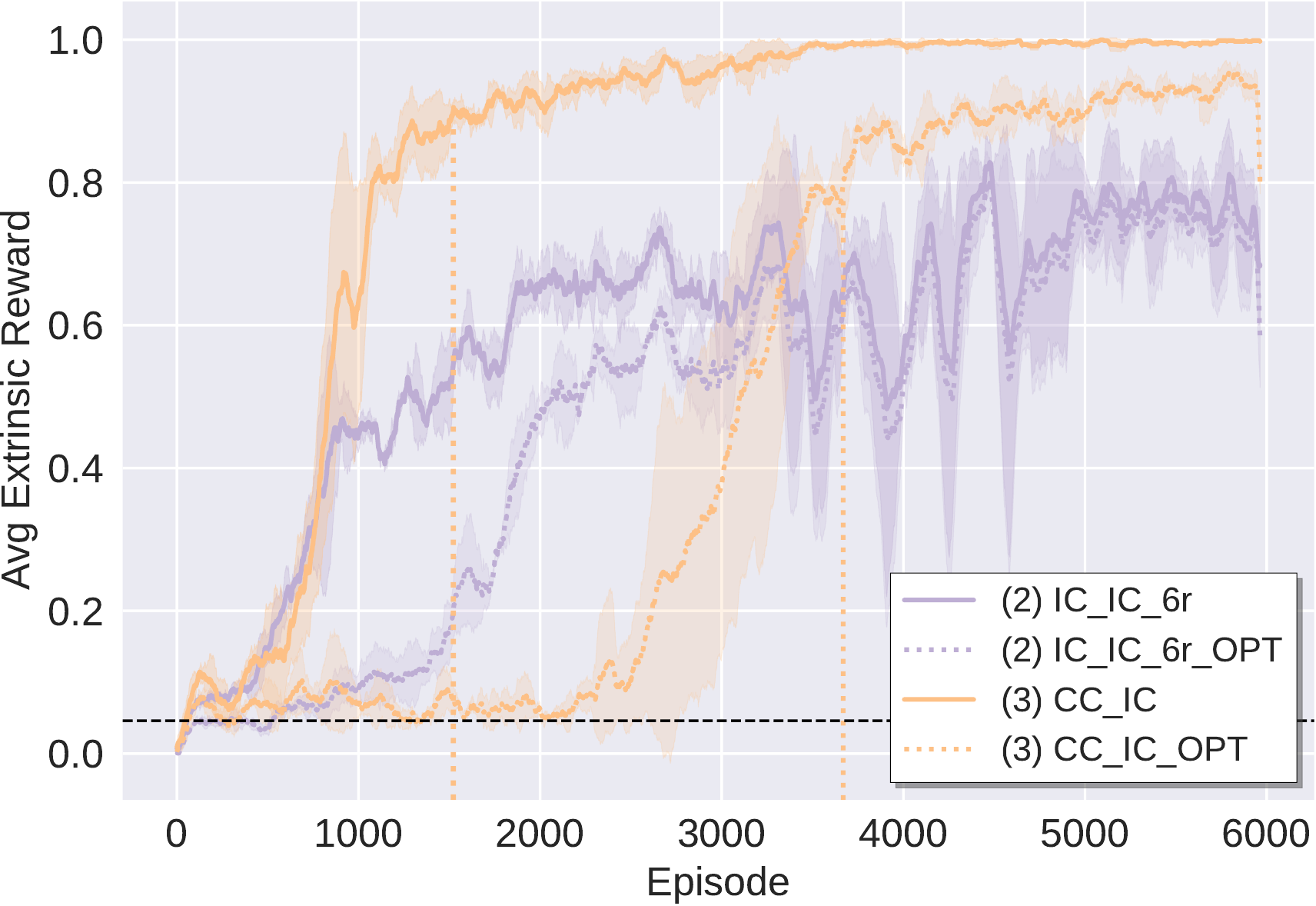} \\
        (a) \\
        \includegraphics[width=.7\columnwidth]{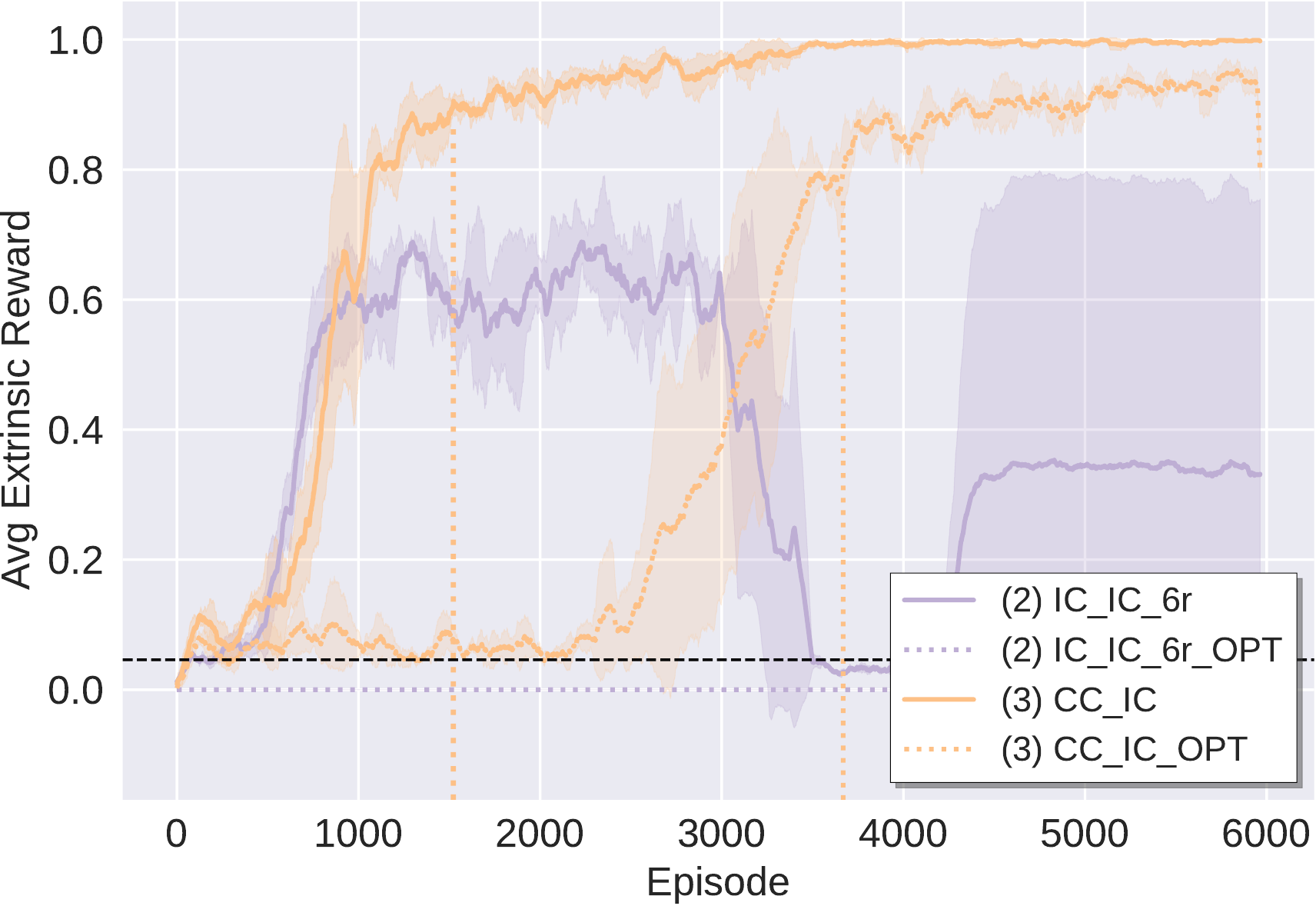} \\
        (b)
    \end{tabular}
    \caption{Success rate (SR) achieved by (a) $W_0$; and (b) $W_1$ by using an individual approach with curiosity for encouraging exploration (\texttt{IC\_CC\_6r}, purple lines), compared to the case when a single centralized critic is trained for both agents (\texttt{CC\_IC}, orange lines). Here, the individual approach utilize 6 parallel environments, whereas only 3 parallel environments are employed by the centralized critic approach.}
    \label{fig:icic6r_centralized}
\end{figure}

Results for this second experiment are shown in Figure \ref{fig:icic6r_centralized}. Even if the number of parallel runners is increased (which should decrease the variance of the gradients, and also feed the critic with the same amount of samples in comparison with centralized critic learning), we observe that \texttt{IC\_IC\_6r} is unable to perform similarly to \texttt{CC\_IC}, but also performs even worse than \texttt{IC\_IC\_3r}, specially for $W_1$. The difference is that \texttt{CC\_IC} is updated almost twice faster, as it executes an optimization step for the trajectories collected by each worker and it is a single centralized critic network for all the agents. On the contrary, in \texttt{IC\_IC\_3r} and \texttt{IC\_IC\_6r} each worker has its own critic module, which is updated once for the experiences collected by their respective actor. Nevertheless, if the number of optimization steps was the key to perform better, then at double the number of episodes, any individual approach should achieve similar performance levels than those by a centralized critic. However, this is not the case either, thereby arriving at the conclusion that a centralized critic performs better than individual critic modules. Consequently, subsequent experiments consider a centralized critic.

\subsection*{RQ2: \emph{Does a centralized curiosity yield better performance levels than maintaining the curiosity locally at every agent?}}

Once the superior performance of a centralized critic has been verified, we now inspect whether it is beneficial to share curiosity information among the agents.

\begin{figure}[t!]
    \centering
    \begin{tabular}{cc}
        \includegraphics[width=.455\columnwidth]{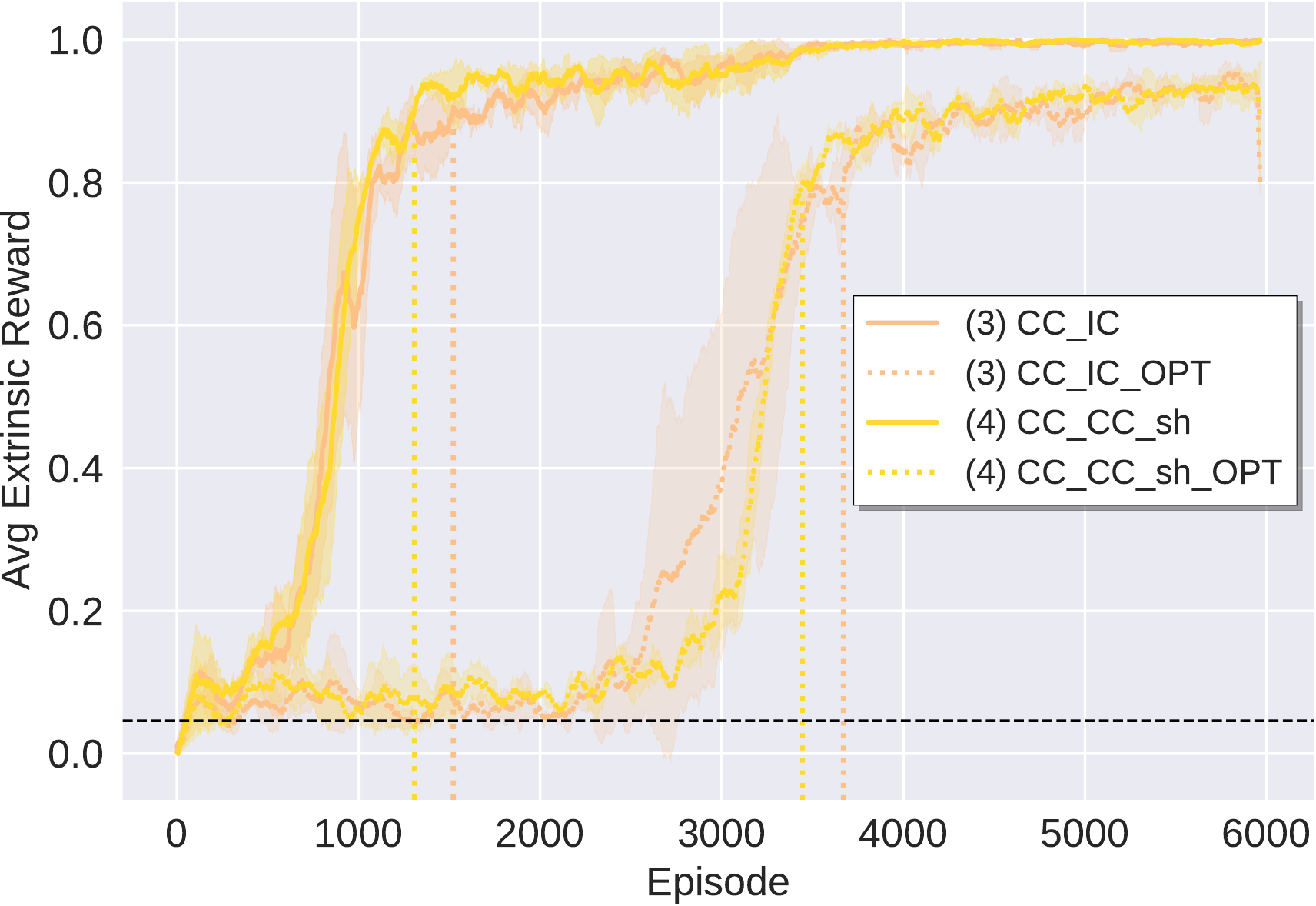} & 
        \includegraphics[width=.455\columnwidth]{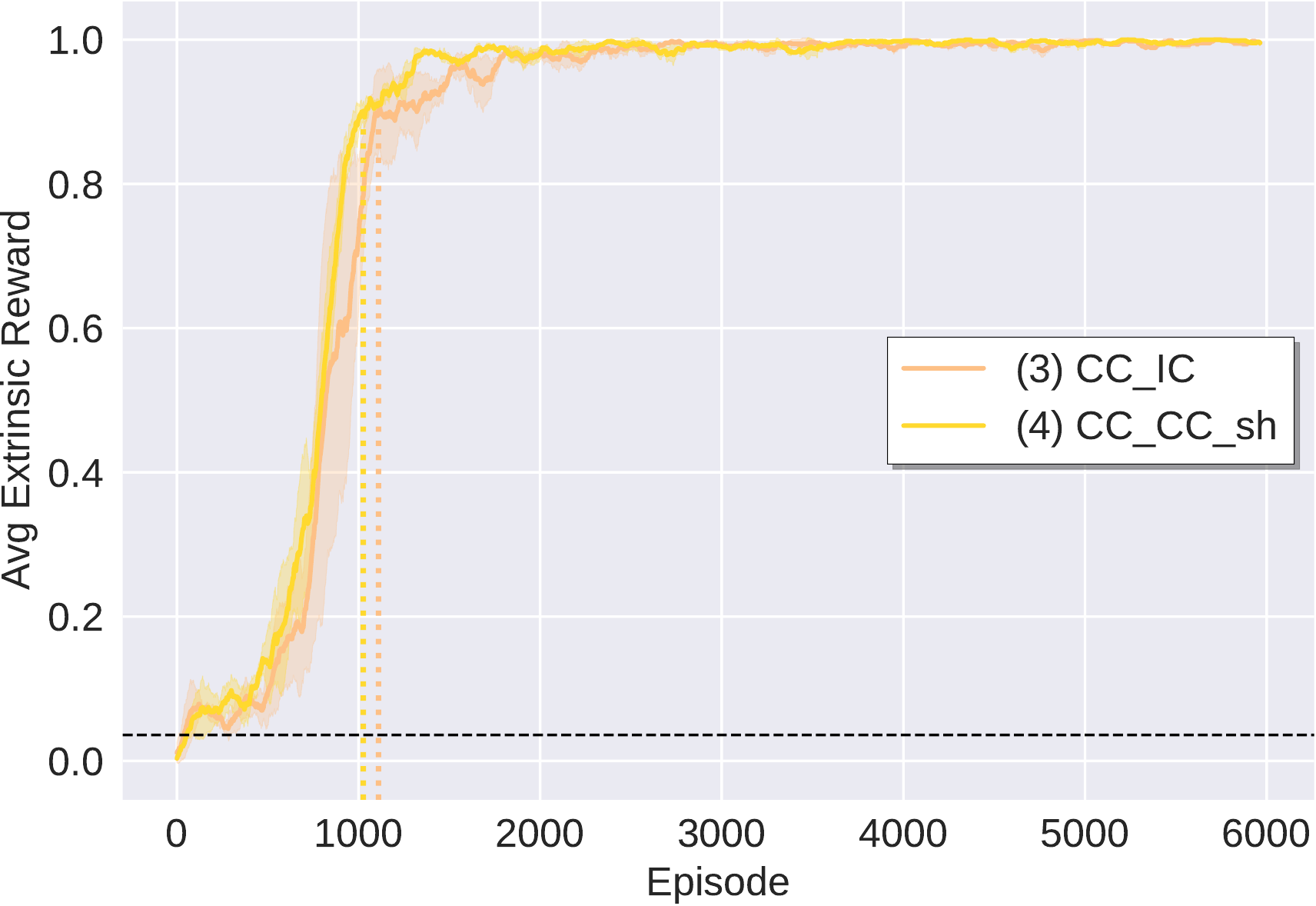} \\
        (a)  & (b) \\
        \includegraphics[width=.455\columnwidth]{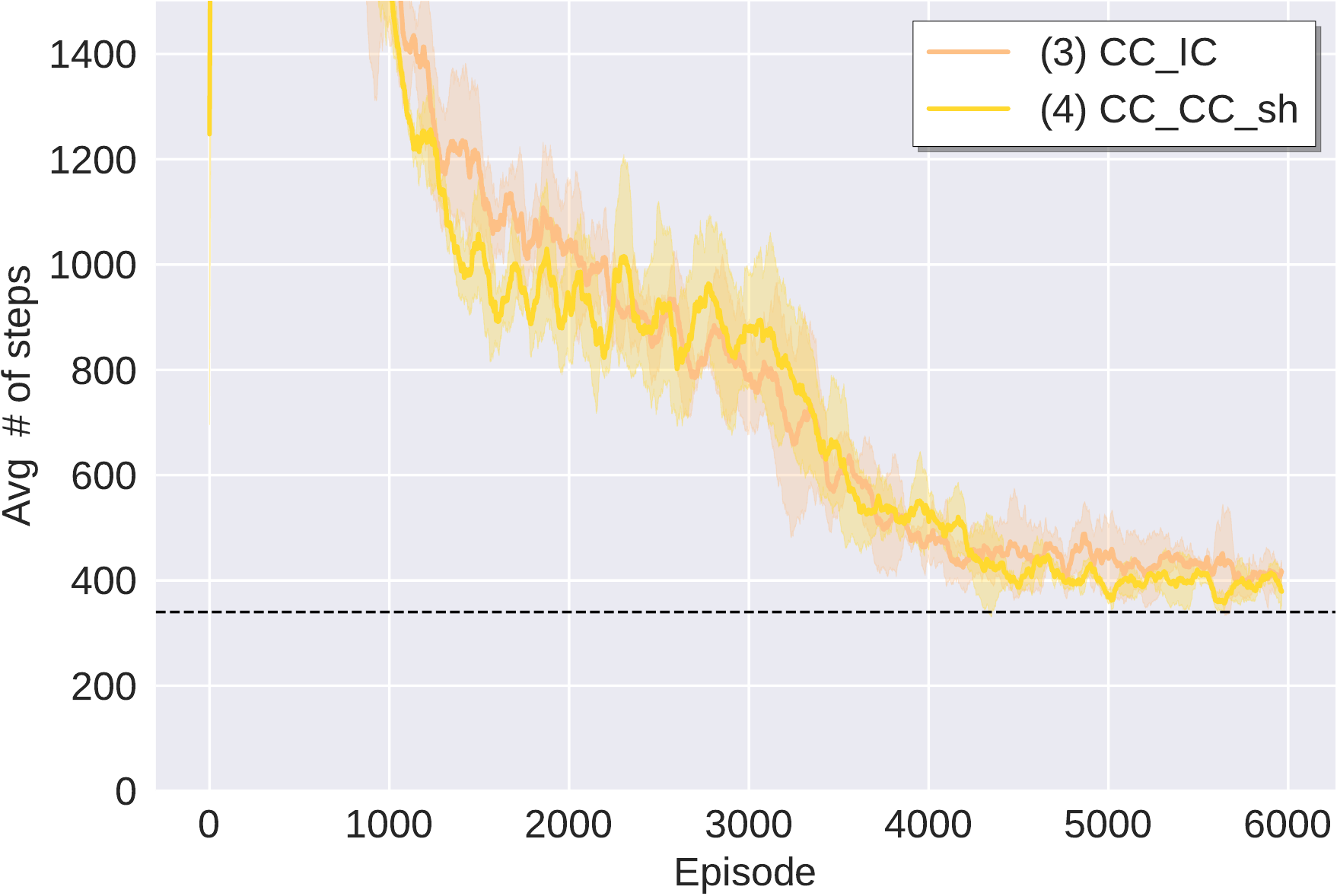} &
        \includegraphics[width=.455\columnwidth]{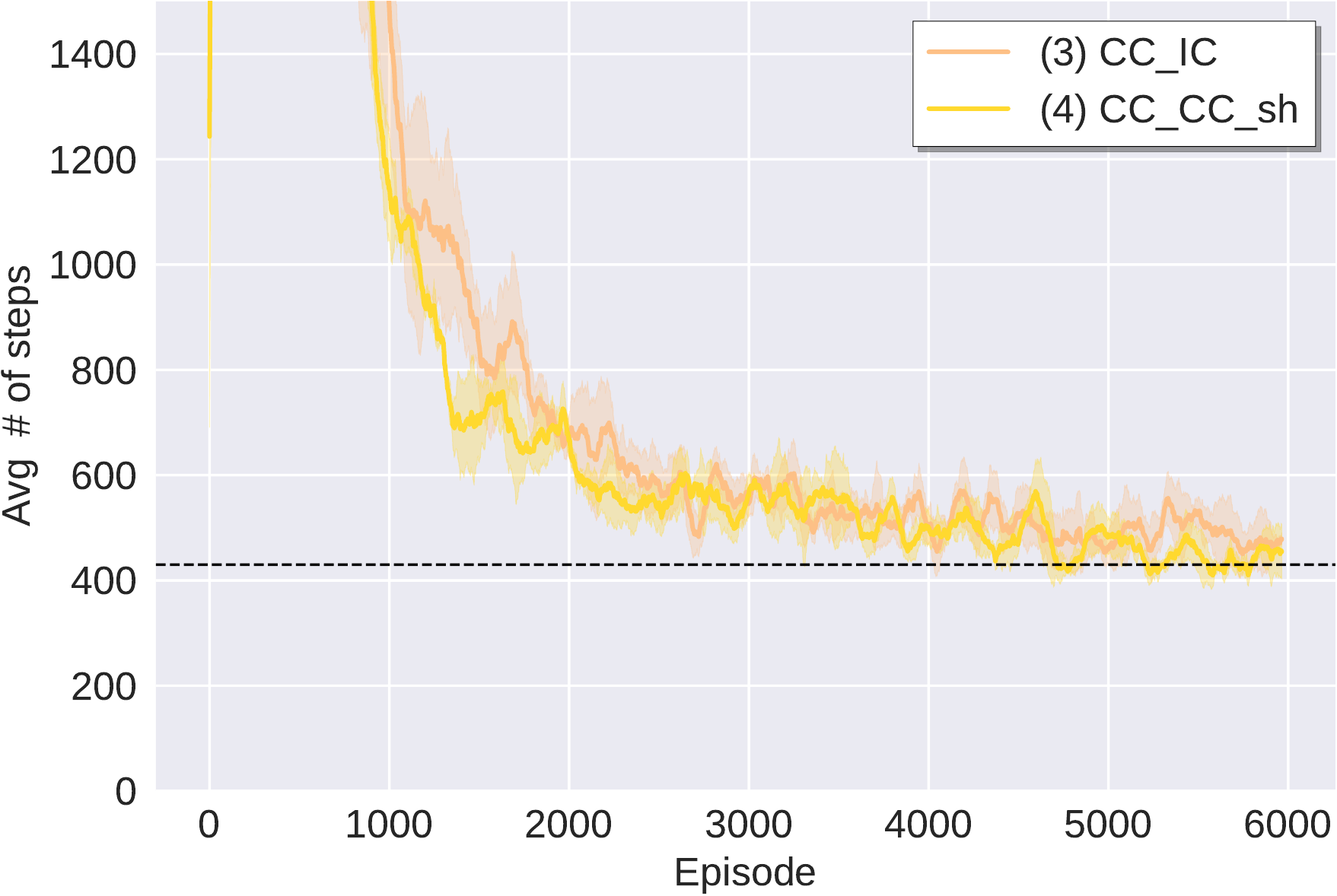} \\
        (c) & (d)
    \end{tabular}
    \caption{(a, b) Success rates and (c, d) average number of steps attained by $W_0$ (a and c) and $W_1$ (b and d) by using a centralized (\texttt{CC\_CC\_sh}) or an individual (\texttt{CC\_IC}) curiosity module with a centralized critic module.}
    \label{fig:centralized_vs_individual_curiosity}
\end{figure}

This second research question can be effectively responded by delving into the plots in Figure \ref{fig:centralized_vs_individual_curiosity}, which depicts the average success rate and number of steps of the framework configured with a centralized critic and individual (\texttt{CC\_IC}) or shared curiosity (\texttt{CC\_CC\_sh}). A quick glimpse at these plots reveals that the results attained by both workers are slightly better for the full centralized approach of the proposed framework. The need for having a large number of episodes to actually see that $W_0$ is capable of traversing the corridor hides any improvements between experiments. By zooming into these results, for $W_0$ the \texttt{CC\_CC\_sh} approach achieves a 90\% of SR with 1309 episodes on average, whereas \texttt{CC\_IC} requires 1522 (an improvement of 14\%). This can be also observed when $W_0$ achieves the destination through the corridor over 80\% of the total test episodes. At this point of the learning process, the fully centralized approach requires 6\% less episodes. In the case of $W_1$, differences appear to be visually negligible, but they actually represent an improvement of 8\%. Furthermore, \texttt{CC\_IC} finishes with a slightly better policy that requires less steps to achieve the goal.

\subsection*{RQ3: \emph{Should we compute curiosity incentives based on the (state,action) pair rather than only the state itself?}}

In the previous discussion we have concluded that sharing curiosity information between agents yields advantages in terms of success rate and number of steps to reach the target. Now we turn the focus towards evaluating whether the intrinsic reward should be made dependent on both the state and action rather than just the state. Previously, the work in \cite{tang2017exploration} showed no empirical differences between both approaches. However, in the cases under study they were not dealing with heterogeneous agents, where the novelty may be influenced by the actions available at each agent. Thus, as foretold in Section \ref{sec:methodology}, our hypothesis is that by making it subject to the (state,action) tuple, different exploration behaviors can be induced into the agents, making it easier for $W_0$ to open the door when being in the required room, but also making it more prone to selecting the \texttt{OPEN} action in unnecessary locations (which can slow down the learning process).
\begin{figure}[t!]
    \centering
    \begin{tabular}{cc}
    \includegraphics[width=0.455\columnwidth]{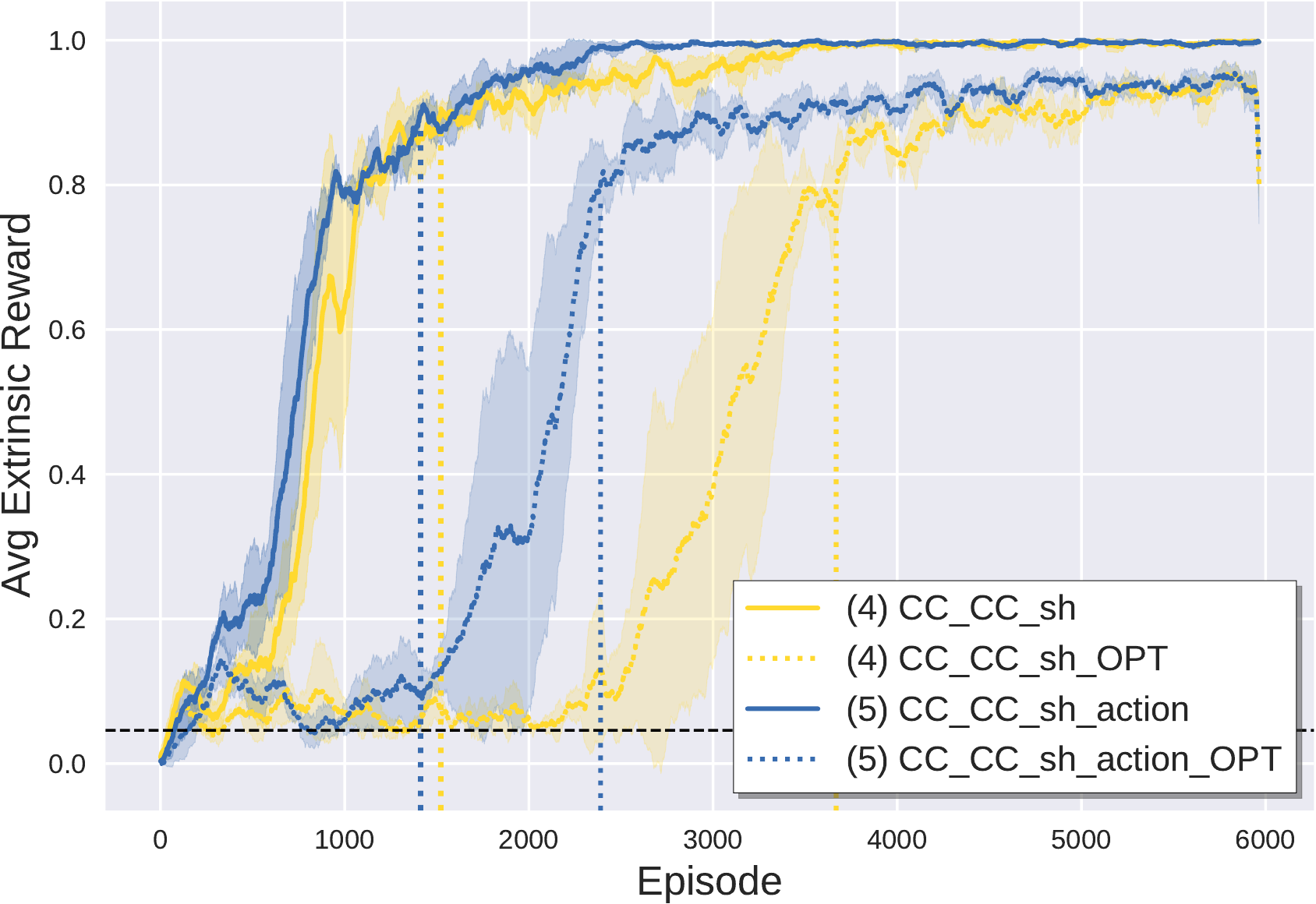}
    &
        \includegraphics[width=0.455\columnwidth]{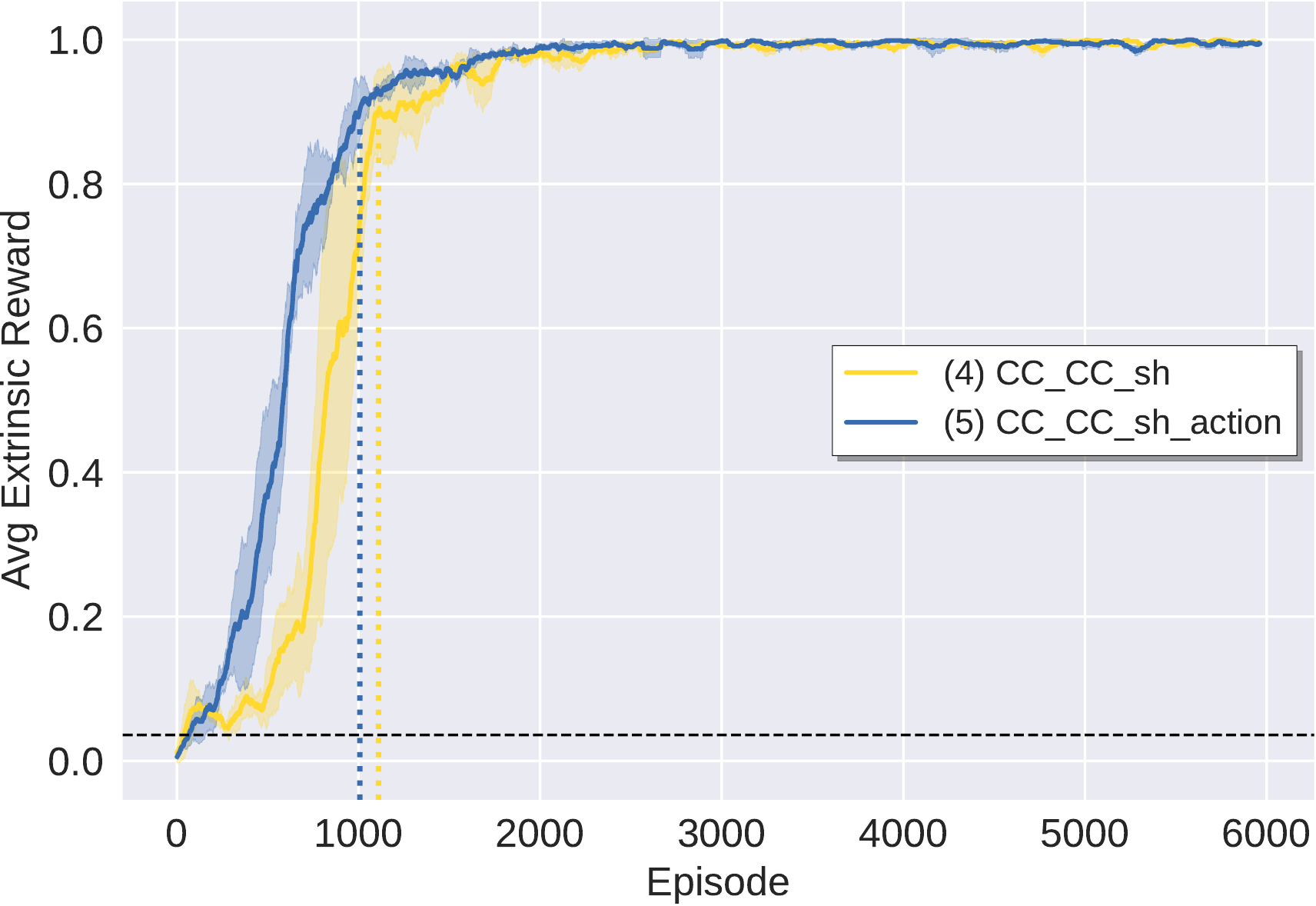} \\
    (a) & (b) \\
        \includegraphics[width=0.455\columnwidth]{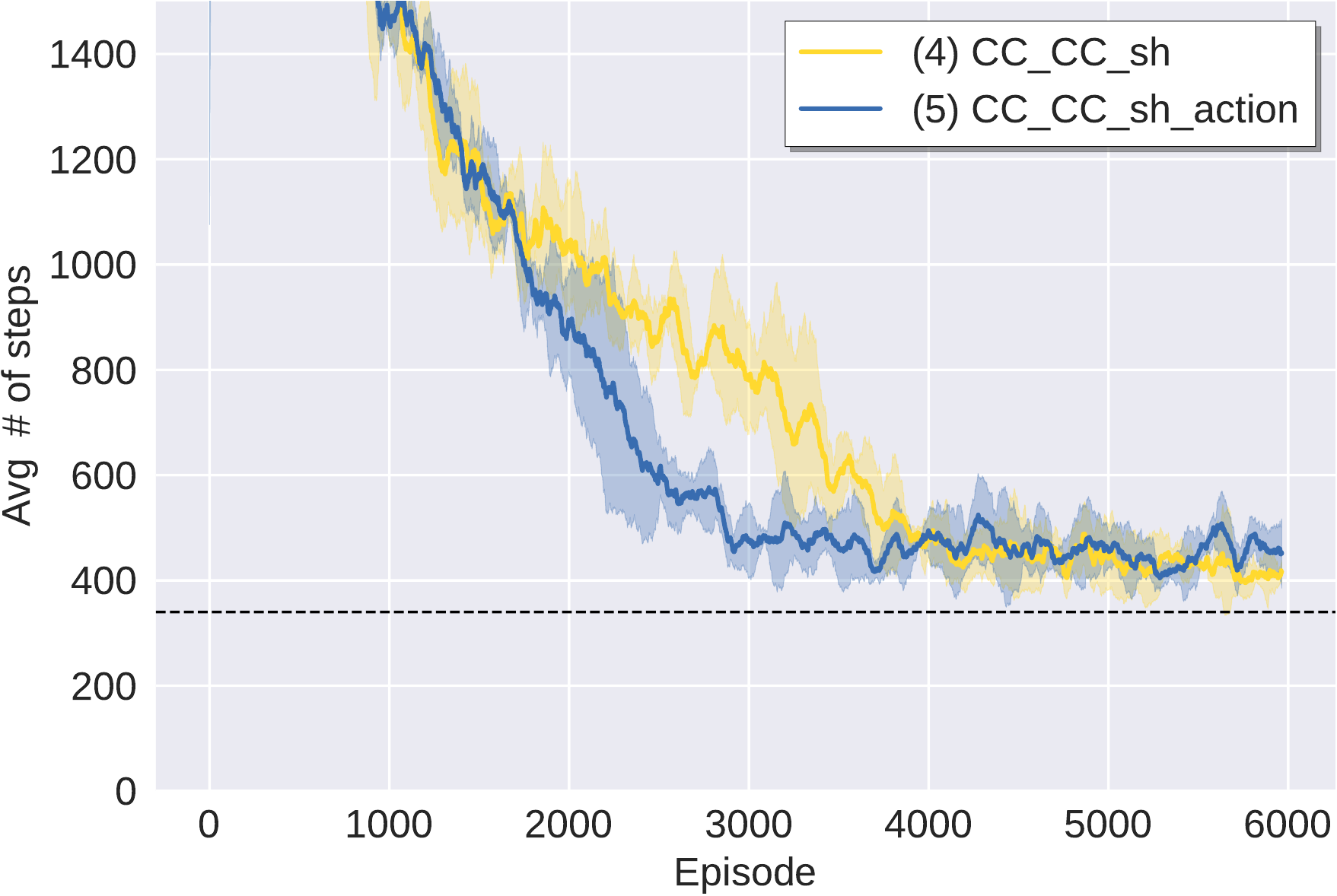}
    &
        \includegraphics[width=0.455\columnwidth]{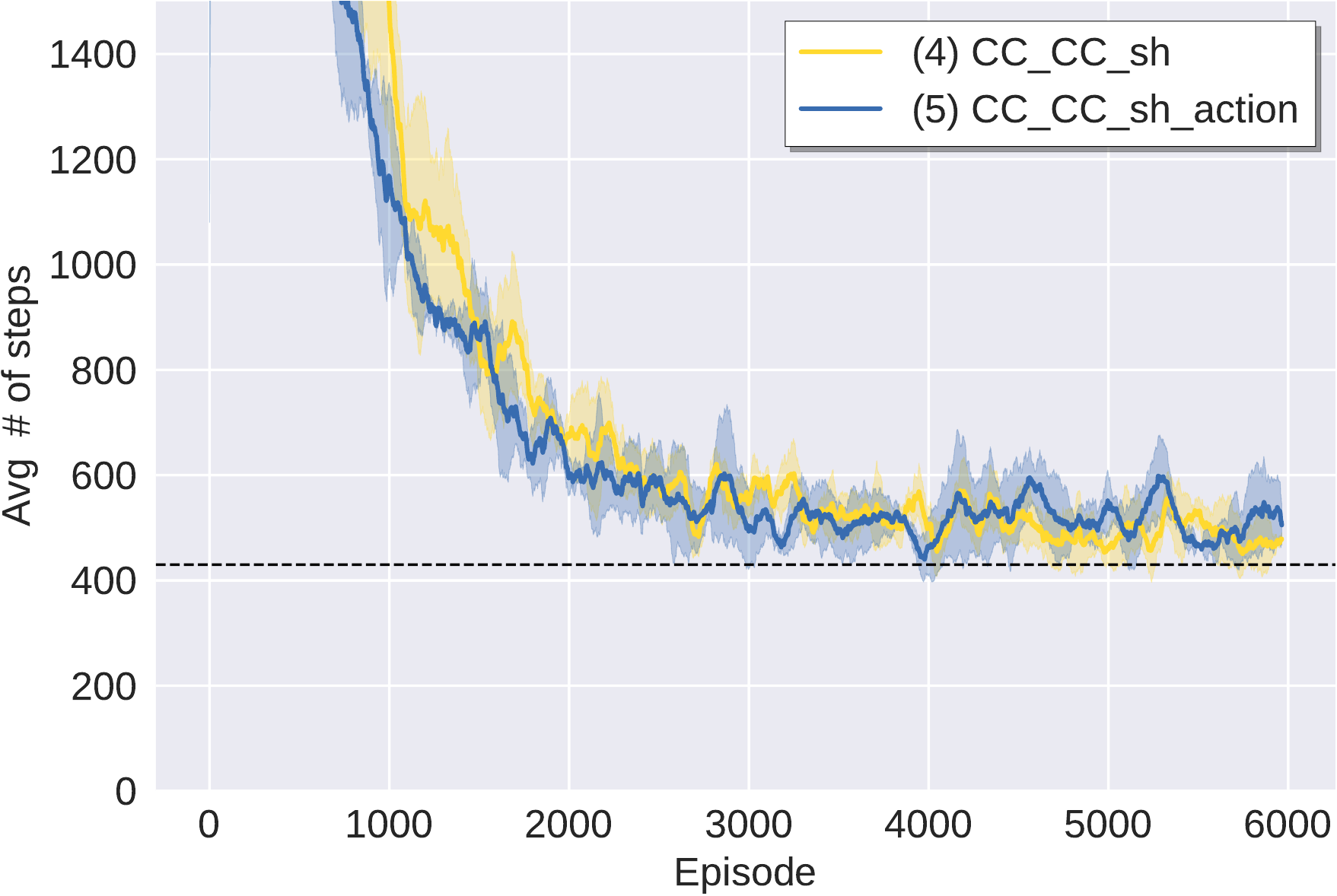} \\
    (c) & (d)
    \end{tabular}
    \caption{(a, b) Success rates and (c, d) average number of steps attained by $W_0$ (a and c) and $W_1$ (b and d) by using a centralized critic and a centralized curiosity module based on state (\texttt{CC\_CC\_sh}) or (state,action) (\texttt{CC\_CC\_sh\_action}).}
    \label{fig:action_based_curiosity}
\end{figure}

In light of the results depicted in Figure \ref{fig:action_based_curiosity}, it is fair to claim that our hypothesis holds. On the one side, $W_0$ exhibits a success rate convergence improvement of almost 1000 episodes when considering success as traversing the corridor to reach the target. This enhancement can be attributed to a smoother exploration bonus, which is representative on how the required steps decay more abruptly after finding out that path. On the other side, once that the path is discovered, it gets stacked with a policy that is slightly worse than the two before analyzed approaches. We speculate that the reason for this effect is the same that leads the agent to find the path faster: the exploration component (\emph{intrinsic reward}) is high when compared to the extrinsic bonuses, which makes the agent undergo noise in its learning process (higher entropy). The same behavior is also distilled into the policy learned by $W_1$, which scores worse despite converging faster.

\subsection*{RQ4: \emph{Should agents get updated their intrinsic rewards only by experiences that are reproducible as per their action spaces?}}

Finally, we evaluate our proposed collaborative framework configured with a centralized critic and a centralized action-based curiosity, but filtering the updates at novelty of each agent subject to being reproducible. Differences should appear mainly for agent $W_1$, so that its learning process changes by deleting those experiences that modify its curiosity inappropriately.
\begin{figure}[h!]
    \centering
    \begin{tabular}{cc}
        \includegraphics[width=0.455\columnwidth]{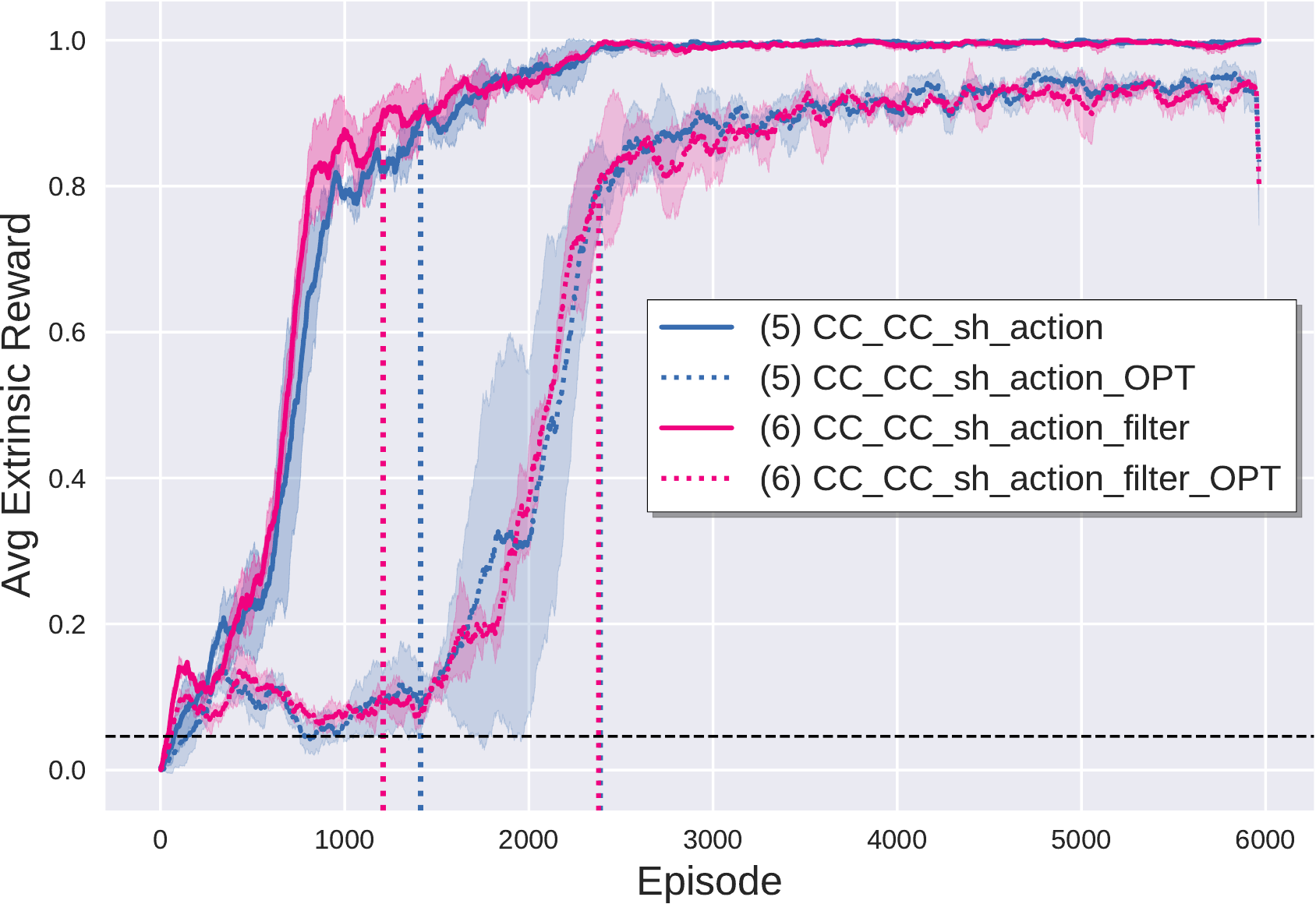} &
        \includegraphics[width=0.455\columnwidth]{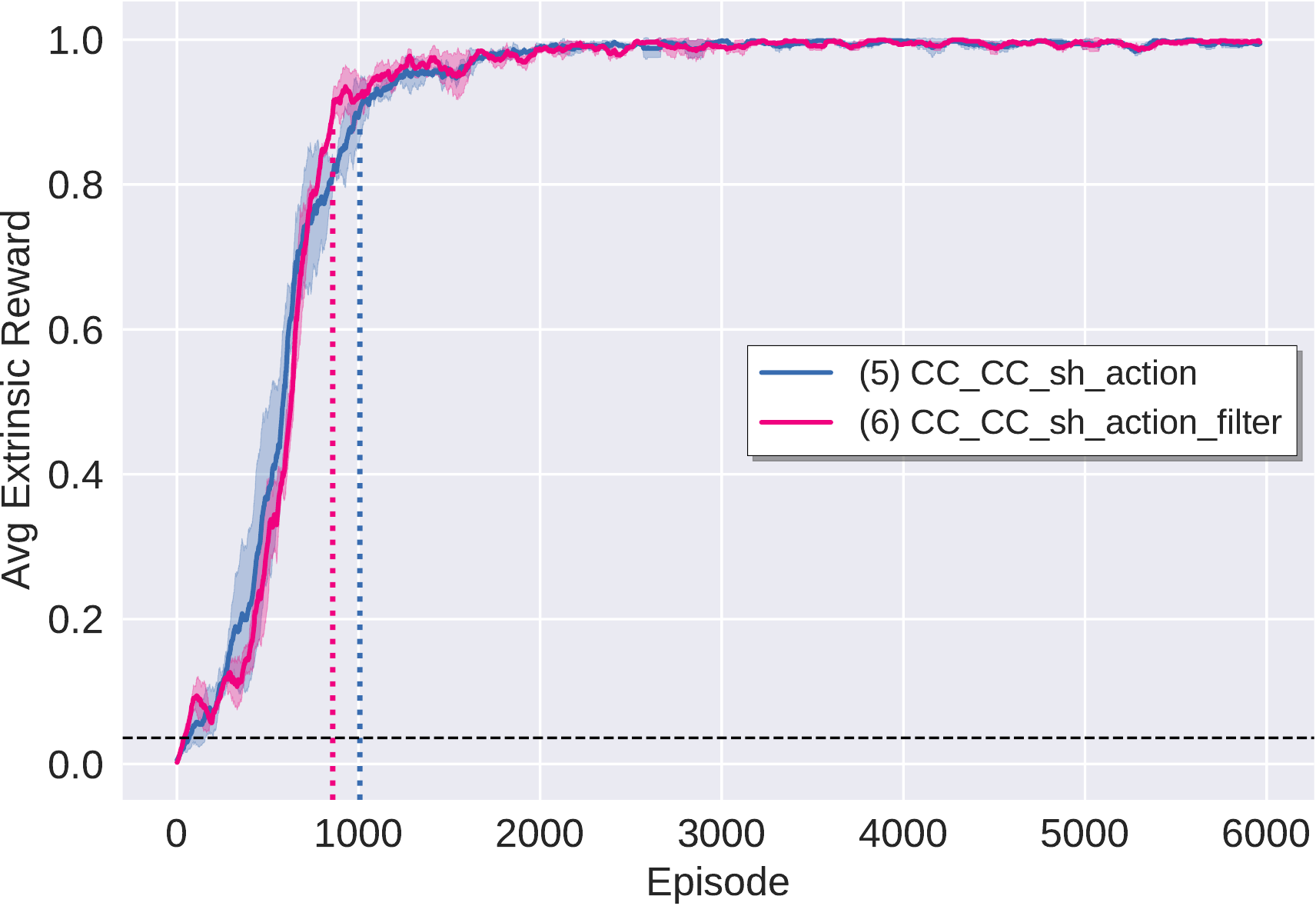} \\
    (a) & (b) \\
    \includegraphics[width=0.455\columnwidth]{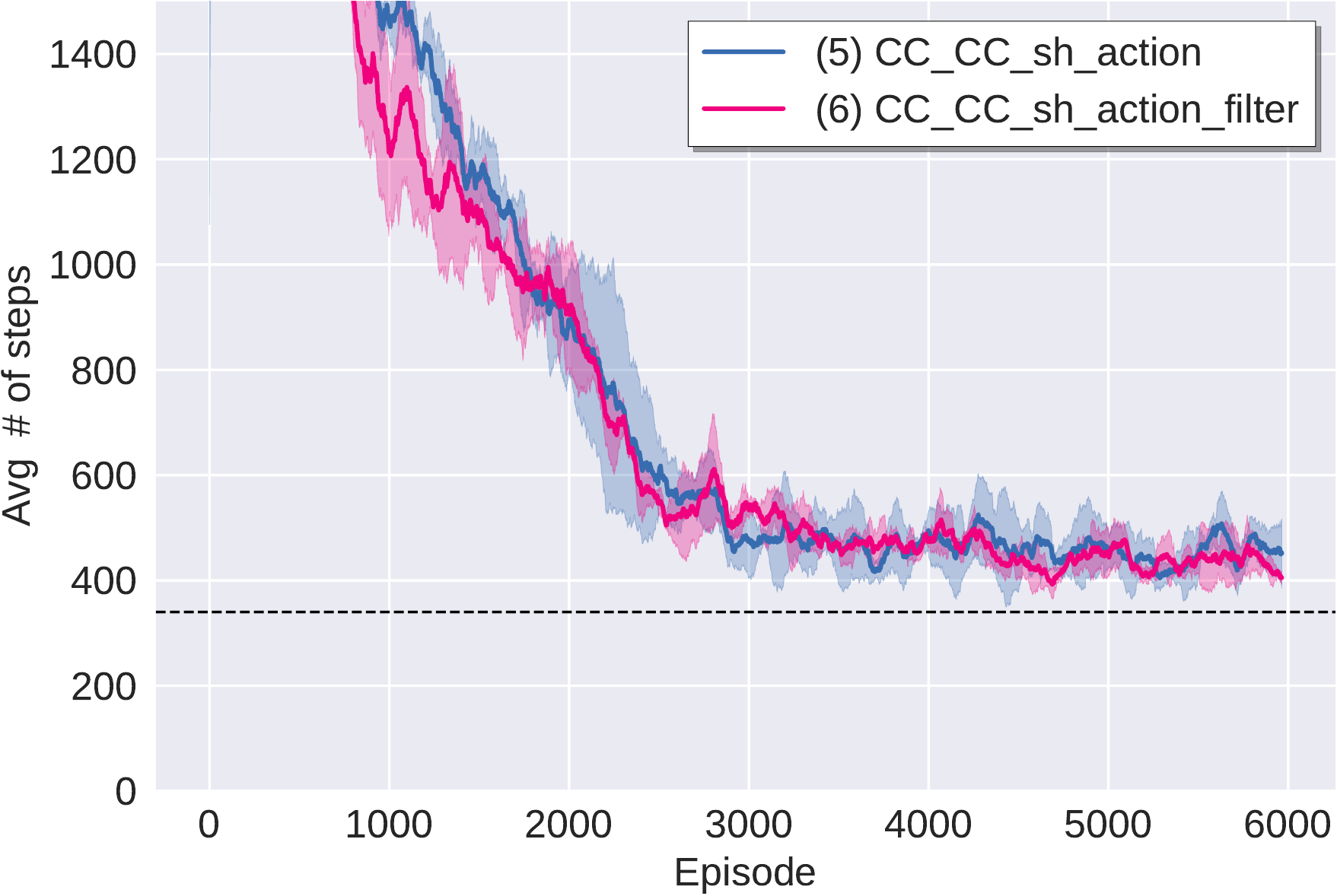} & \includegraphics[width=0.455\columnwidth]{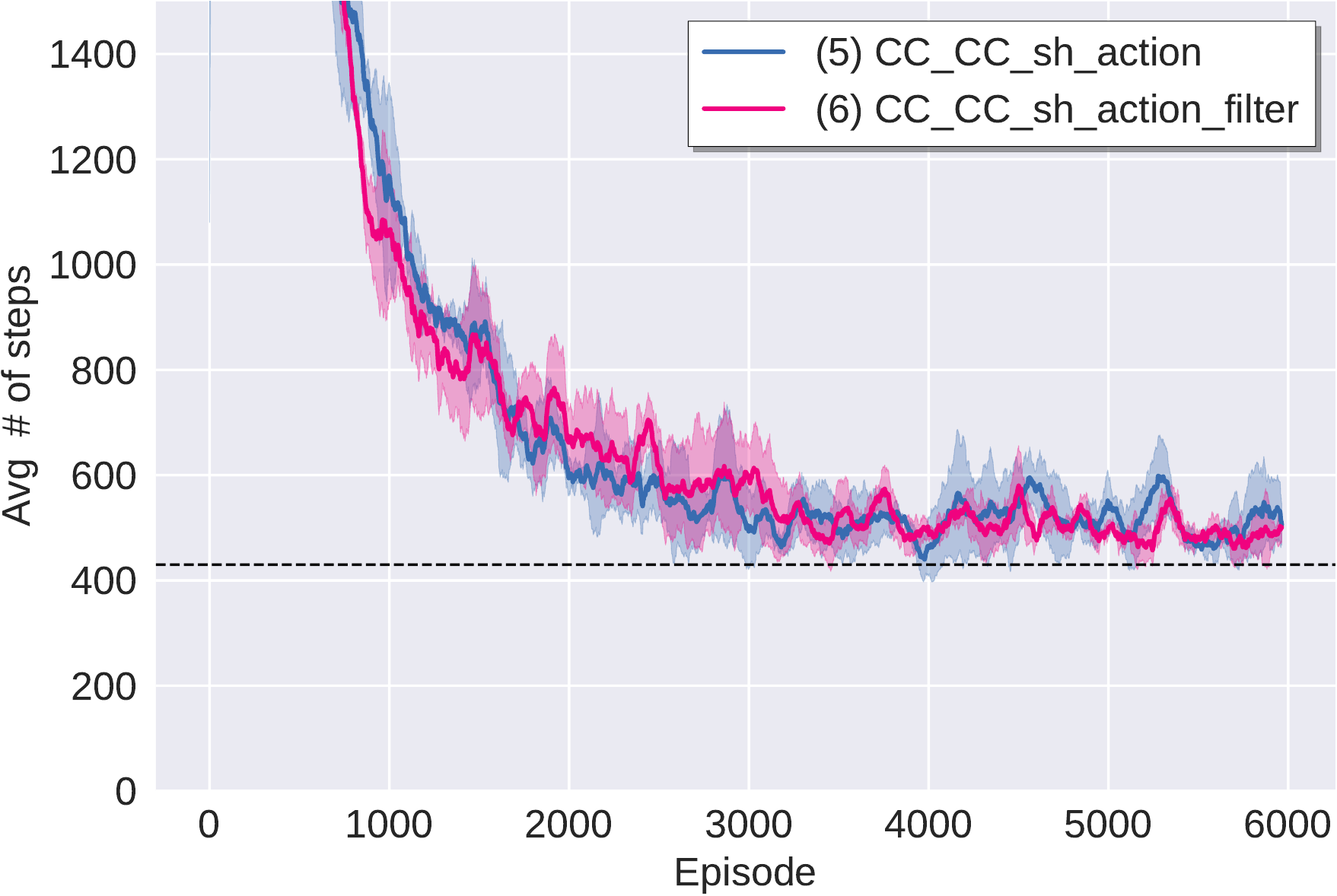} \\
    (c) & (d)
    \end{tabular}
    \caption{(a, b) Success rates and (c, d) average number of steps attained by $W_0$ (a and c) and $W_1$ (b and d) by using a centralized critic, a centralized curiosity module based on (state,action), and with (\texttt{CC\_CC\_sh\_action\_filter}) and without (\texttt{CC\_CC\_sh\_action}) filtering episodes in which the action to open the door has been used.}
    \label{fig:action_based_curiosity_filter}
\end{figure}

Plots nested in Figure \ref{fig:action_based_curiosity_filter} expose that indeed, a narrow performance gap arises between the two compared approaches \texttt{CC\_CC\_sh\_action\_filter} and \texttt{CC\_CC\_sh\_act}\-\texttt{ion}. Both workers converge to a SR of 90\% faster when compared to any of the previously analyzed configurations of the framework, attaining an improvement of $7.7$\% (for $W_0$) and 15\% (for $W_1$) in comparison to the second-best solution.  

\subsection{Exploration versus Exploitation: When?}

One of the major issues arising from the analysis of the results is that the number of steps of the optimal policy is far from the number of steps taken by executing the minimum number of actions\footnote{Experiments have considered a frame skip equal to 4, hence the optimal solution with 1 frame per step should require less interactions of the agent with the environment.}. The reason is that, even at the final stages of the training process, the learned policy is too stochastic and still features much variability. Depending on the problem, this might be a good result as it allows the agent to adapt to changes more easily \cite{haarnoja2017reinforcement}. However, if the aim is to learn to perform the task over the given environment as efficiently as possible, the optimal policy should be the one that converges with the minimum required steps towards the target.

The problem is that there is not such a given objective embedded in the reward function that establishes how to solve the problem with the minimum number of steps. The refinement of the policy is due to the agent/algorithm itself through the backpropagation of the final reward to all the previously visited states, with the help of the discounted expected return. This is done through the value-function estimates $V(s)$. At the same time, the advantages used to compute the gradients of the policy depend on those value estimates. Hence, if those advantages are not well balanced (i.e., if they do not represent the actual values), this can make the agent explore when it should be actually exploiting information (and vice versa). Moreover, this problem gets even more involved when considering that the accuracy of the extrinsic-intrinsic streams depicted in Expression \eqref{eq:advantages_mix_calculation}, where depending on the maturity of the values estimates and how noisy are the advantages respect to their true value, there are going to be parts that converge faster (in train time). 

\begin{figure}[h!]
    \centering
    \begin{tabular}{ccc}
         \includegraphics[width=0.4\columnwidth]{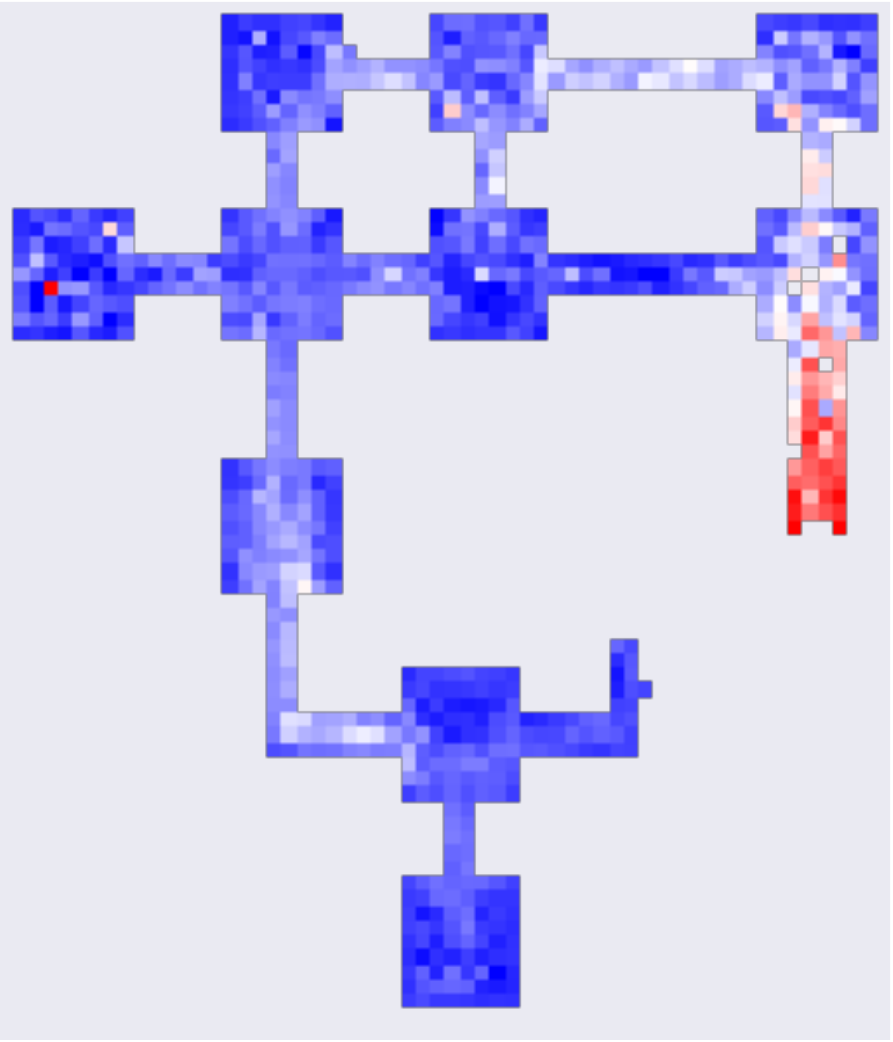} & \includegraphics[width=0.4\columnwidth]{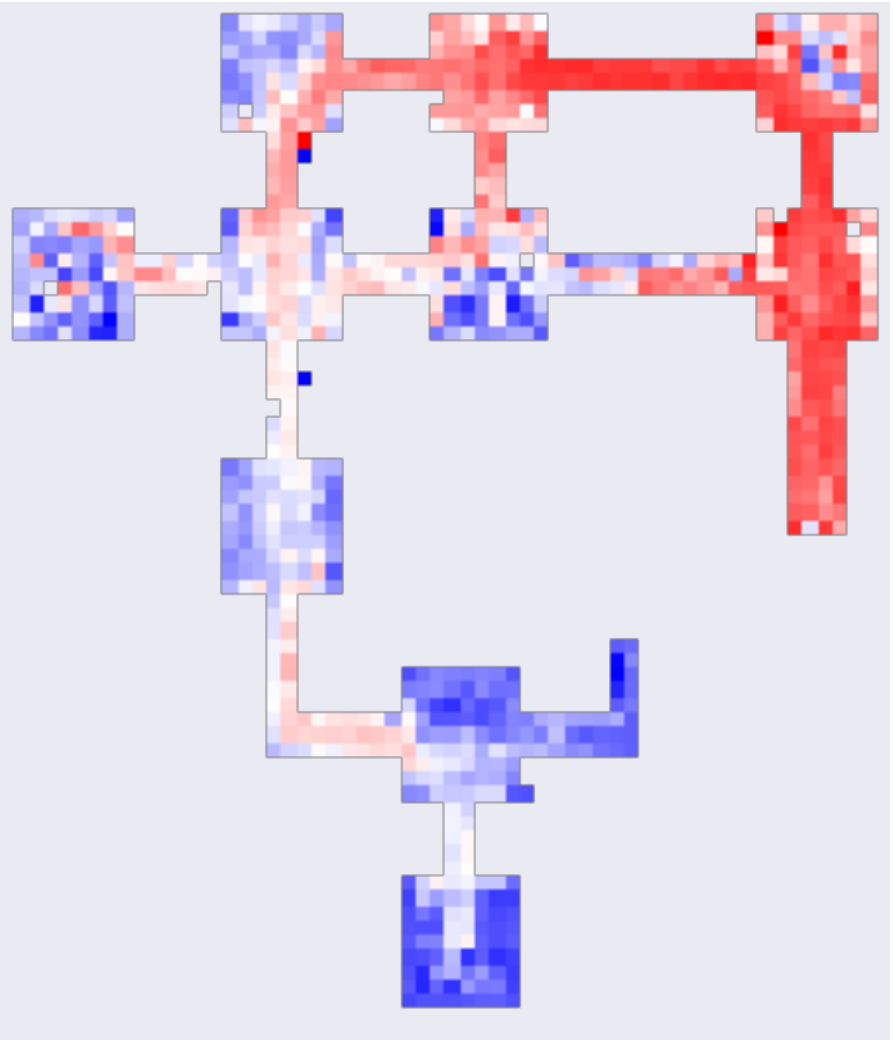} & \multirow{4}{*}{\includegraphics[width=0.07\columnwidth]{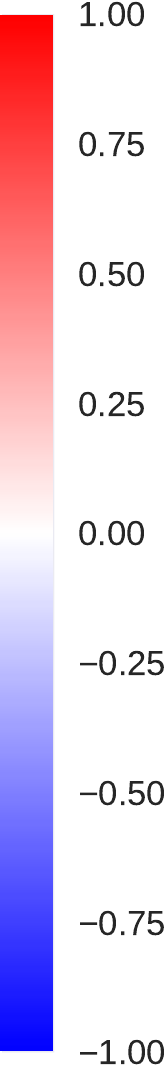}}\\
         (a) & (b) & \\
         \includegraphics[width=0.4\columnwidth]{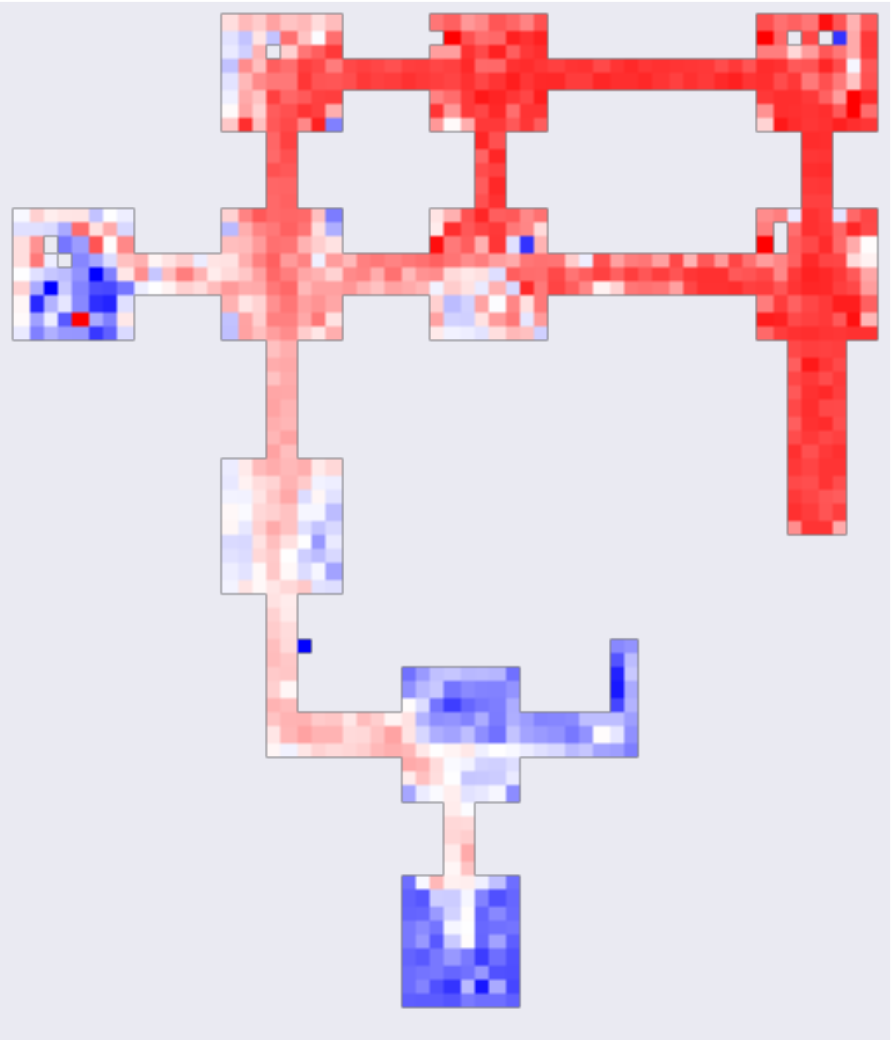} & \includegraphics[width=0.4\columnwidth]{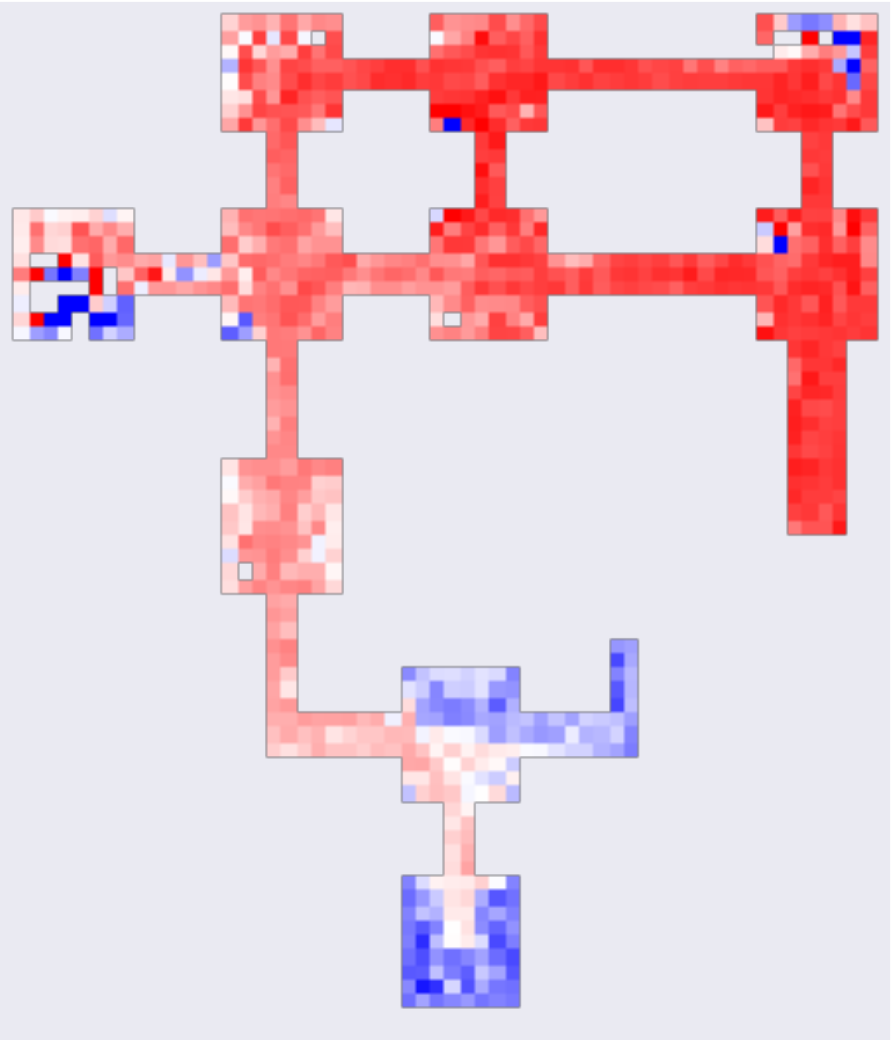} & \\
         (c) & (d) & \\
         \multicolumn{2}{c}{\includegraphics[width=0.6\columnwidth]{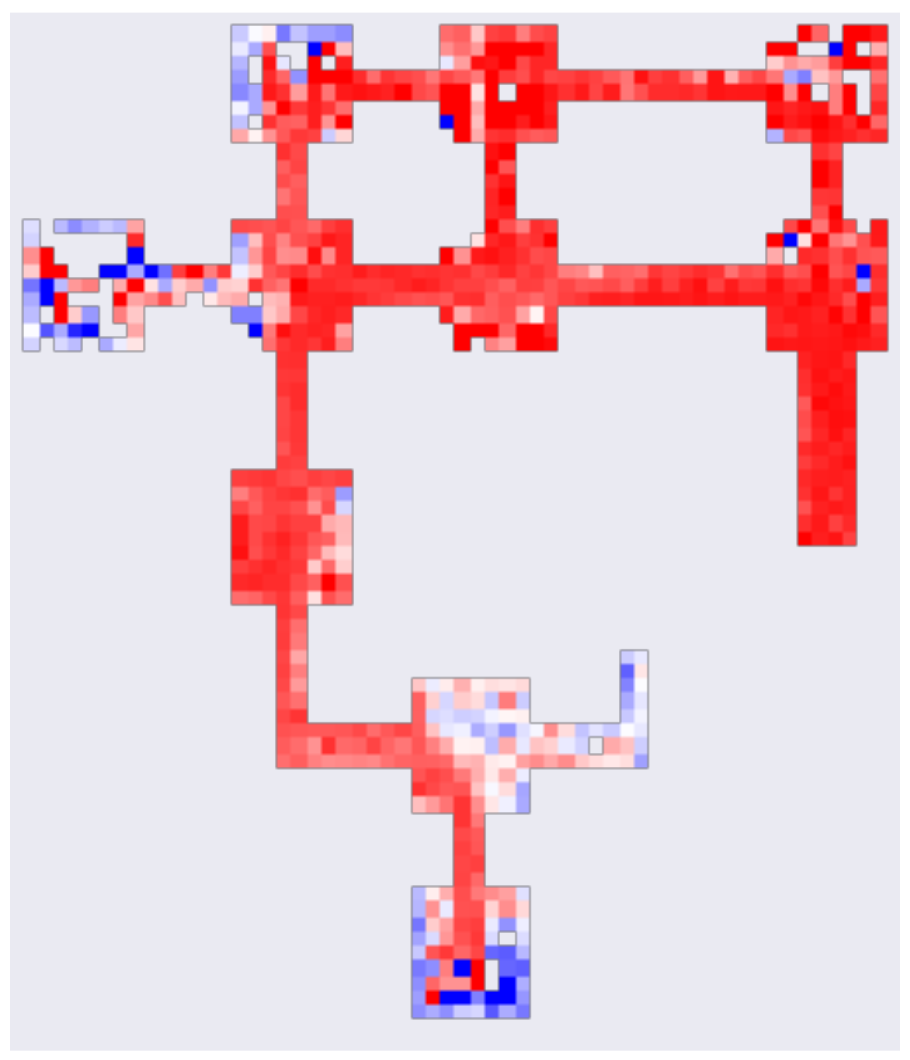}} & \\
         \multicolumn{2}{c}{(e)} &
    \end{tabular}
    \caption{Divergence maps of $W_0$ that shows when the total advantage estimator is guided by either the extrinsic (red) or intrinsic (blue) advantage term, at different stages of the training process: (a) 0-500 episodes; (b) 500-1000 episodes; (c) 1000-1500 episodes; (d) 1500-2000 episodes; (e) 5500-6000 episodes. The results are shown for one of the \texttt{CC\_CC\_sh\_action} simulations.}
    \label{fig:advantage_ext_predominance}
\end{figure}

This problem can be better understood by examining Figure \ref{fig:advantage_ext_predominance}, which depicts a heatmap showing the balance between the extrinsic and intrinsic parts in the computation of the total advantage at different stages of the training process. We observe that in Figures \ref{fig:advantage_ext_predominance}.a and \ref{fig:advantage_ext_predominance}.b, the agent rarely updates its policy by following the main extrinsic goal, although it becomes more influential as the training process evolves. Interestingly, in Figure \ref{fig:advantage_ext_predominance}.c it can be noticed that the agent, when being in the room where the door is located, cannot manage to go through the corridor. In this location, when the extrinsic advantage dominates, it stimulates the agent to go to the room located above (room 22). Contrarily, when updates are carried out at white-blue spots, the agent's decision might not be aligned with the environment goal, as it is mostly influenced by the intrinsic stream. Nevertheless, as the training process continues, almost all decisions taken are updated accordingly to what the extrinsic stream suggests (Figure \ref{fig:advantage_ext_predominance}.d). This analysis can be contrasted with the results reported in Figure \ref{fig:action_based_curiosity}, where $W_0$ begins to take the path through the corridor more greedily in the range between 1500 and 2000 training episodes. Afterwards, the agent enters an exploitation-driven phase, wherein the advantage extrinsic stream becomes more relevant, indicating that the agent is updated towards mainly accomplishing the environment's goal (Figure \ref{fig:advantage_ext_predominance}.e). 

\begin{figure}[t!]
    \centering
    \begin{tabular}{cc}
        \multicolumn{2}{c}{\includegraphics[width=0.65\columnwidth]{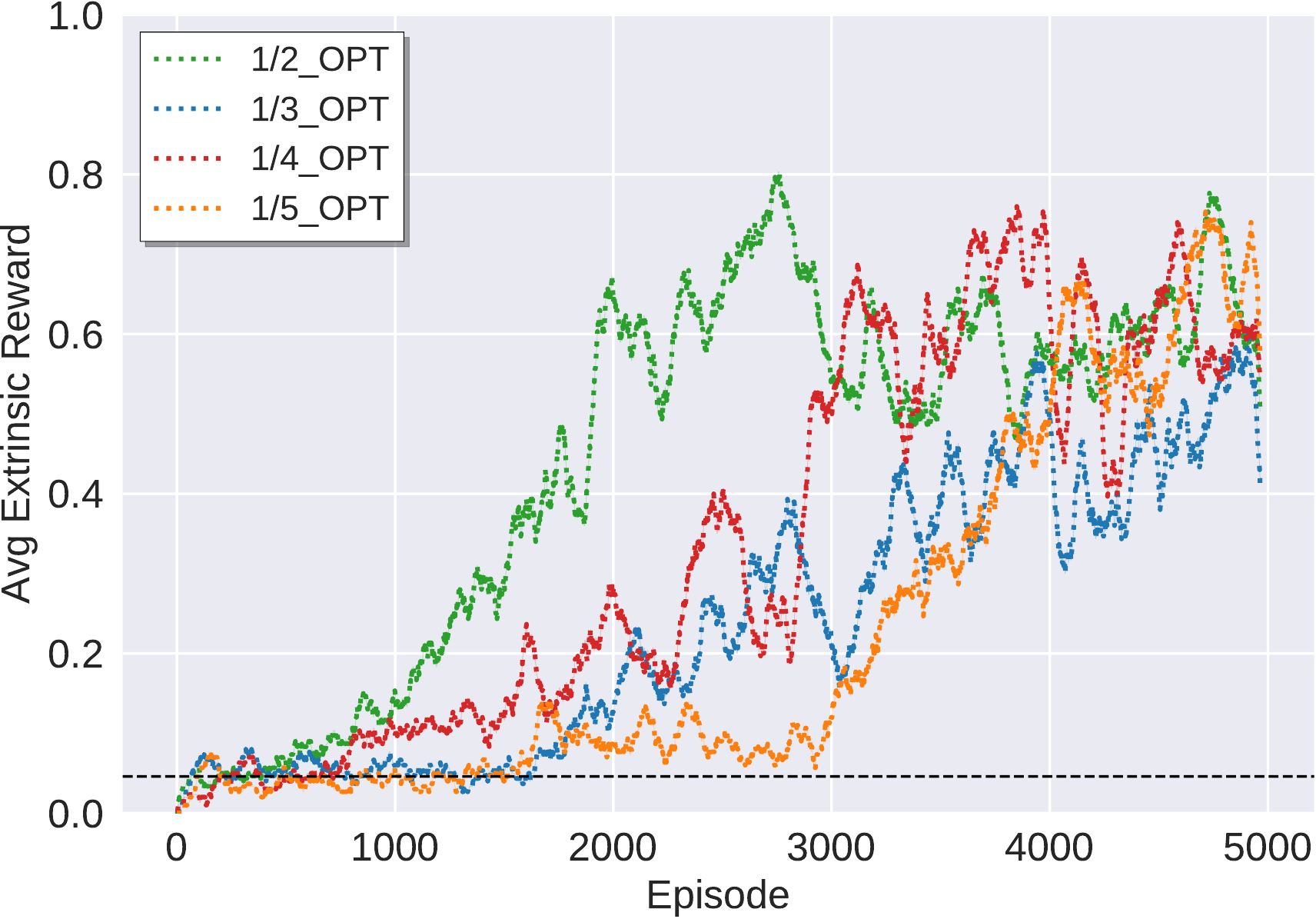}} \\
        \multicolumn{2}{c}{(a)}\\
        \includegraphics[width=0.455\columnwidth]{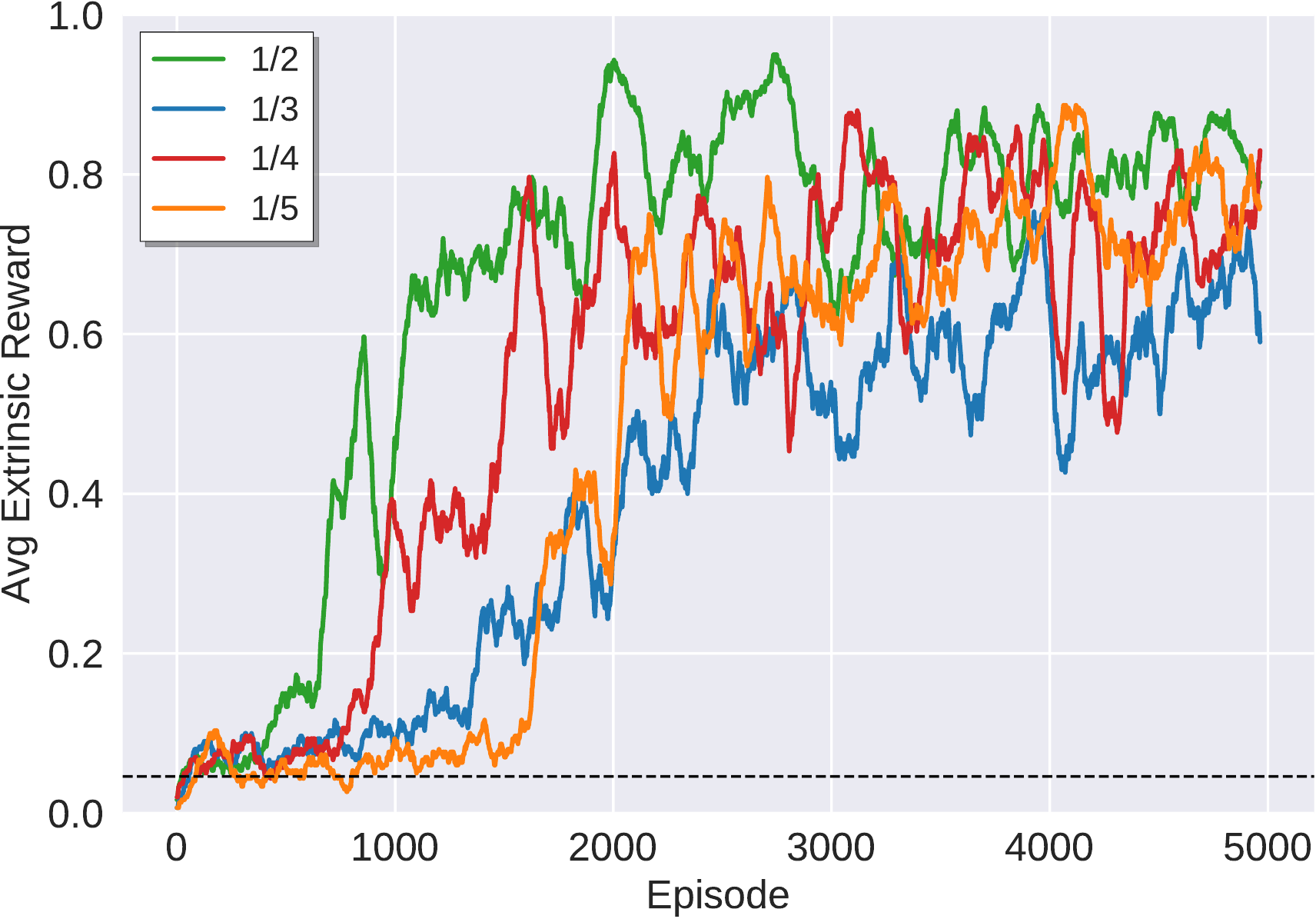} & \includegraphics[width=0.455\columnwidth]{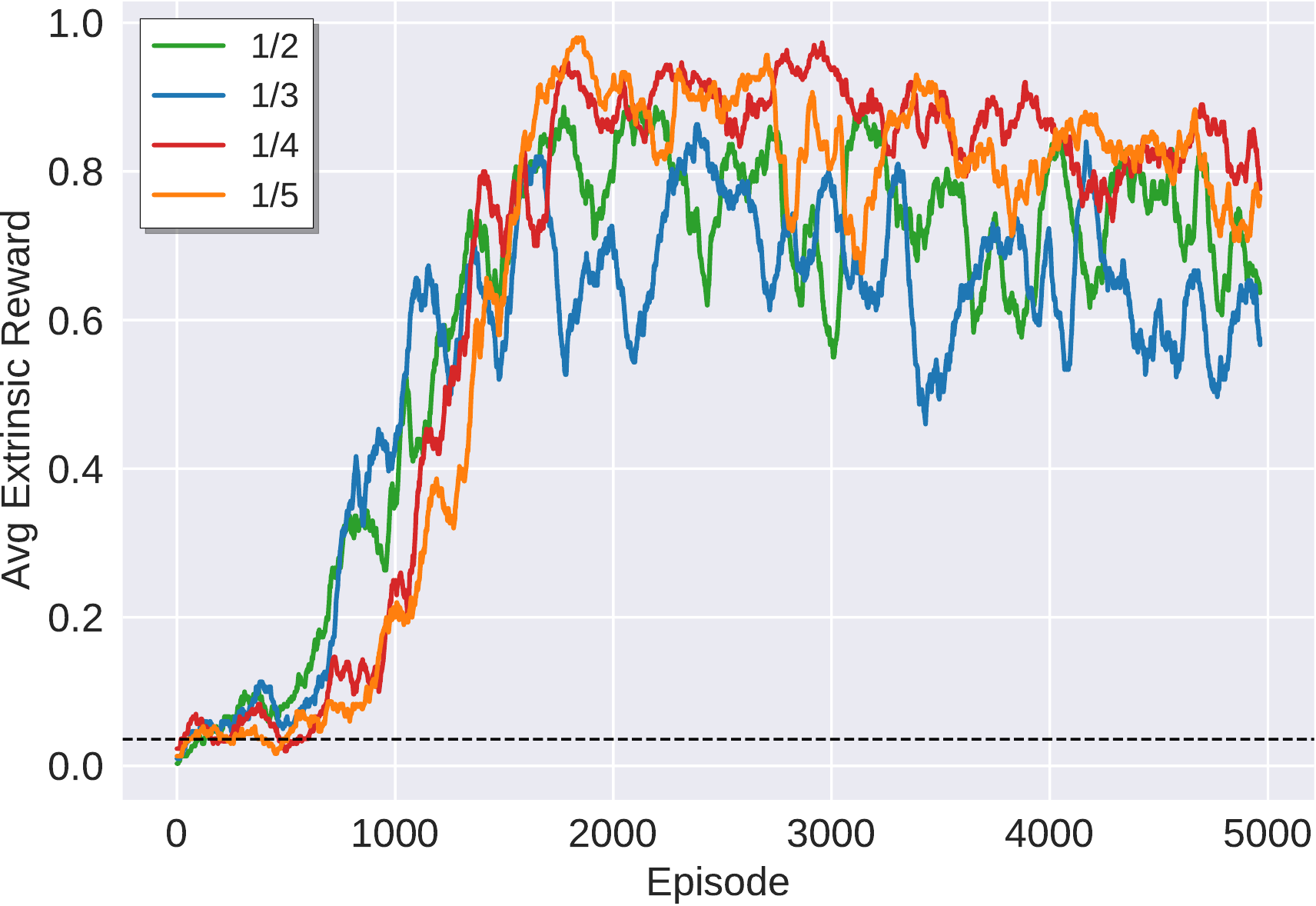} \\
        (b) & (c) \\
        \includegraphics[width=0.455\columnwidth]{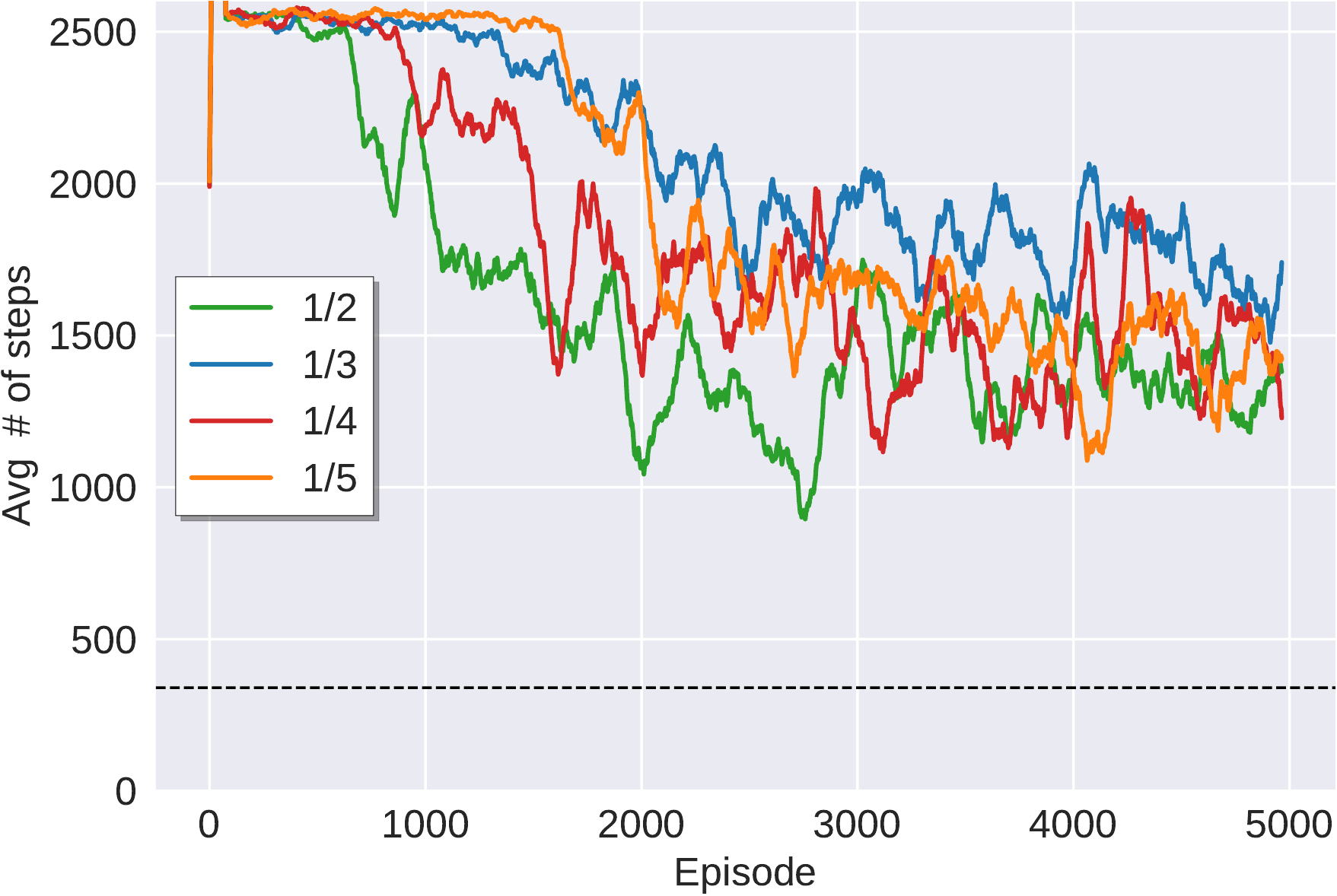} & 
        \includegraphics[width=0.455\columnwidth]{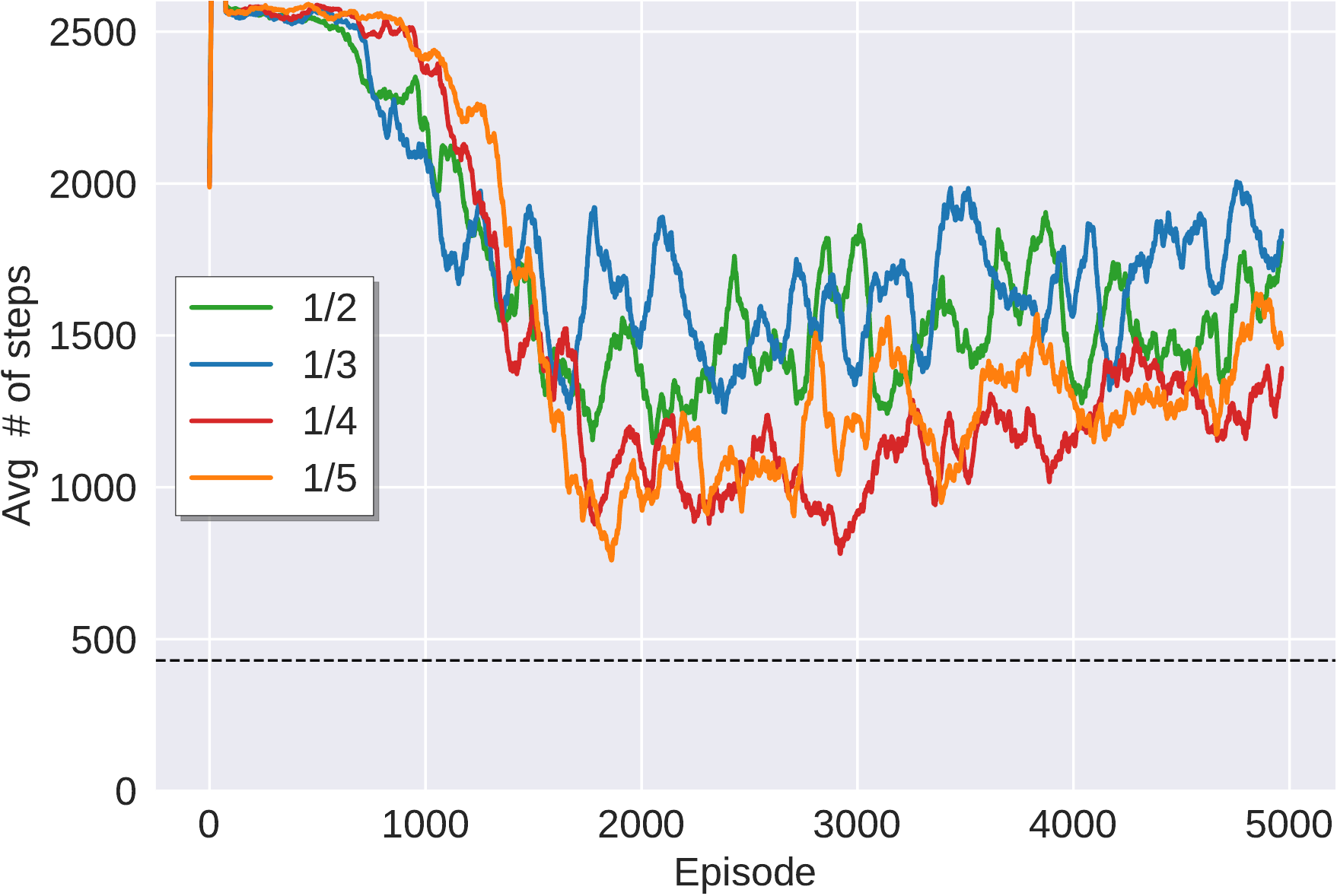} \\
         (d) & (e) 
    \end{tabular}
    \caption{For different values of $\beta$: (a) SR of $W_0$; (b) SR of $W_0$ when success is counted only for those episodes where the goal is reached through the corridor; (c) SR of $W_1$; average number of steps to solve the task for (d) $W_0$ and (e) $W_1$.}
    \label{fig:worker01_beta_value_study}
\end{figure}

One way to overcome this problem is by properly balancing the composition of the total advantage through $\beta$, as using a fixed value of this parameter seem not to be the best choice in view of the results. This unexpected behavior has already been noted in other works \cite{rosser2021curiosity,badia2020never} and still lacks a clear solution in the related literature. To further delve into this matter, we compared the performance of the collaborative framework configured with independent critic and curiosity modules (\texttt{IC\_IC\_3r}) and with different $\beta$ values, so as to ascertain whether they yield any differences in the exploration and exploitation phases of the training process. Results of these side experiments are reported in Figure \ref{fig:worker01_beta_value_study}, where it can be noted that, under the basic $\beta$-weighted implementation of the total advantage, the instability of the learning process is not solved by considering different values of the $\beta$ parameter.

Further along this line, we have carried out two additional experiments by switching off the curiosity for a given number of episodes (i.e., $\beta=0$). With the purpose of just gauging its impact, we have manually tailored the learning process so that two simulations are run, one in which curiosity is deactivated after 1000 episodes, and another in which the deactivation is triggered after 3000 episodes, respectively. These values were selected because they showed up to be the moments when the agent is able to achieve the destination with good guarantees through any given path and by its optimal path, respectively. We will refer to these tailored approaches as \texttt{CC\_CC\_sh\_action\_1000} and \texttt{CC\_CC\_sh\_action\_3000}.

\begin{figure}[h!]
    \centering
    \begin{tabular}{cc}
        \includegraphics[width=0.435\columnwidth]{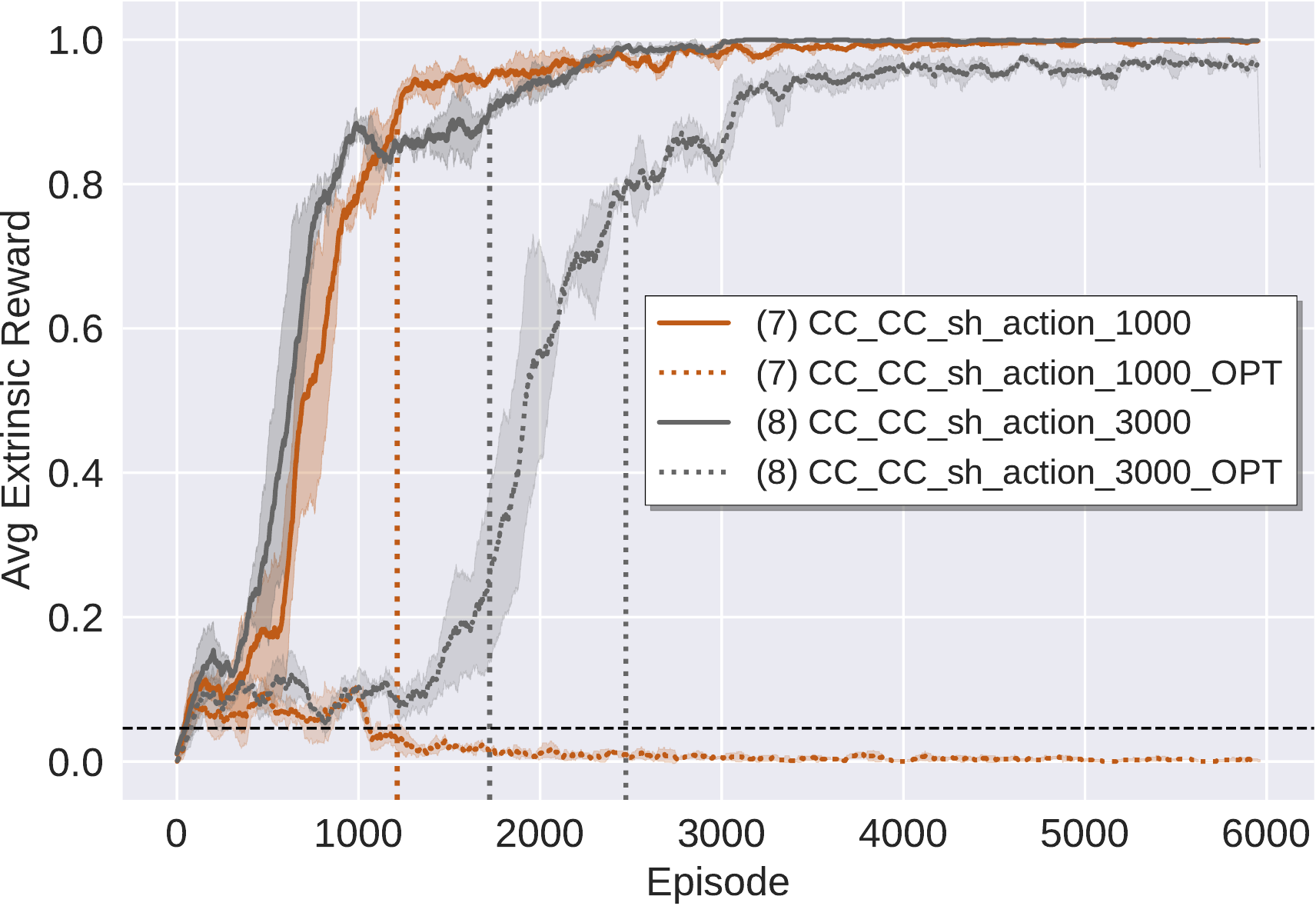} &
        \includegraphics[width=0.435\columnwidth]{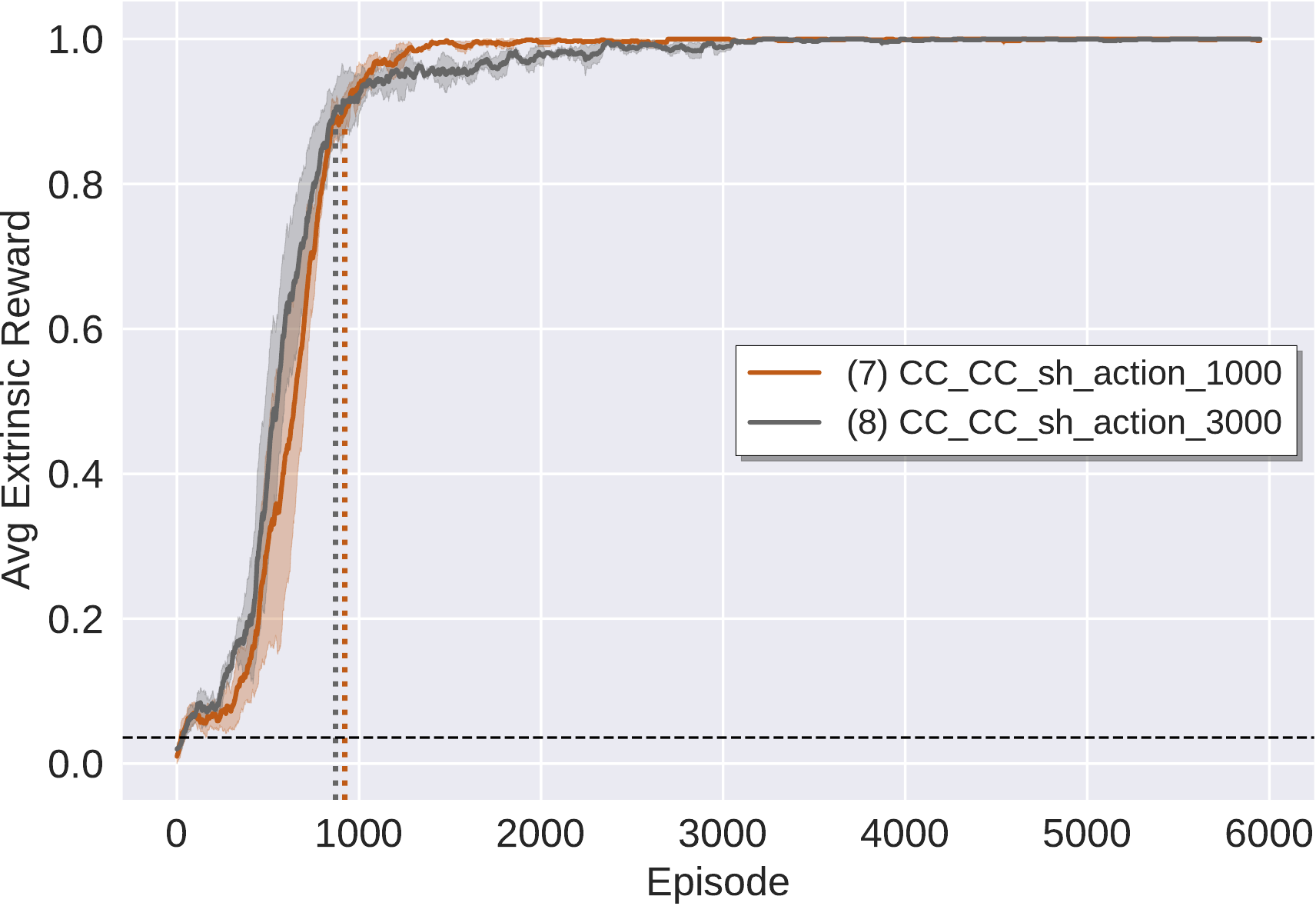} \\
        (a) & (b) \\
        \includegraphics[width=0.435\columnwidth]{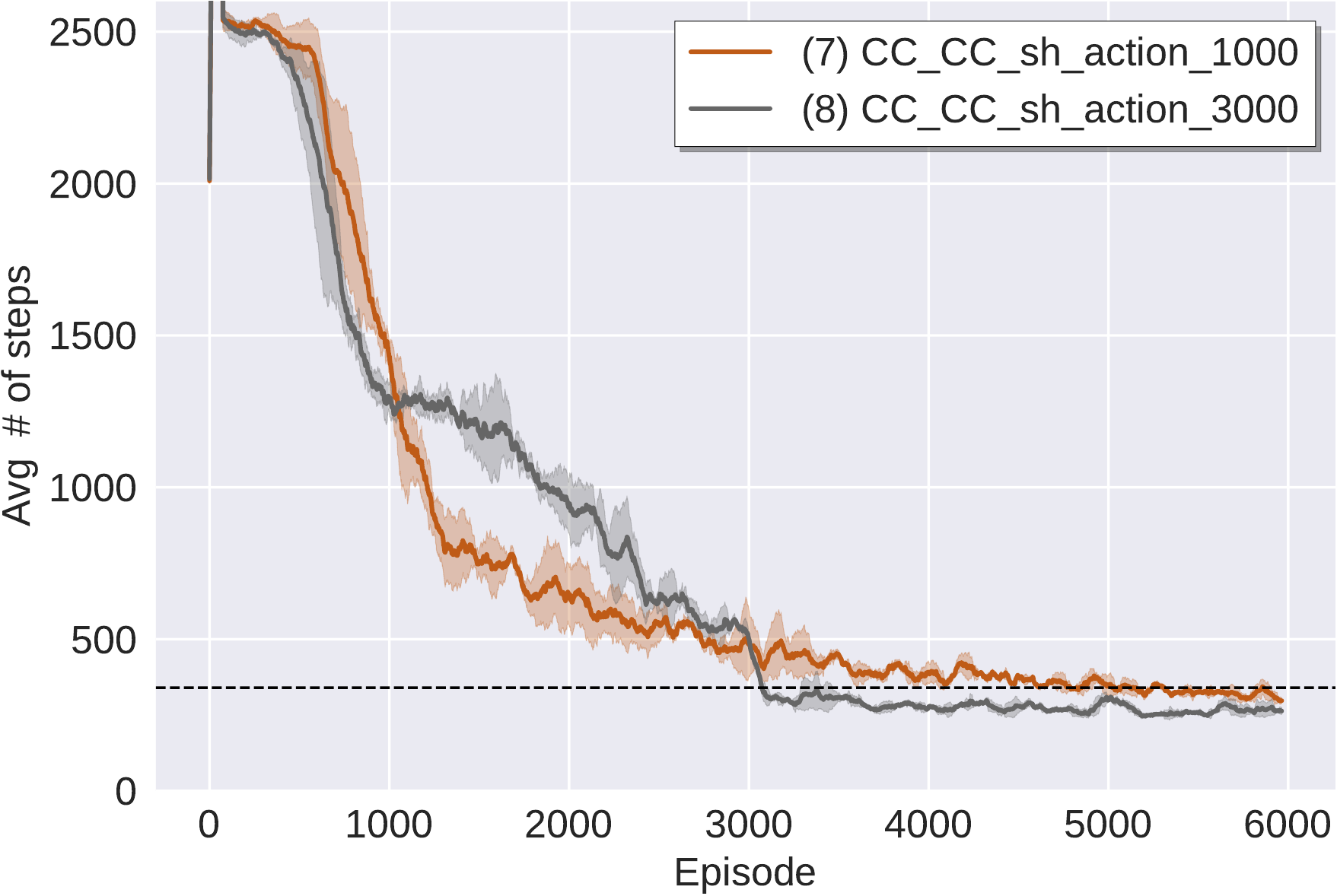} &
        \includegraphics[width=0.435\columnwidth]{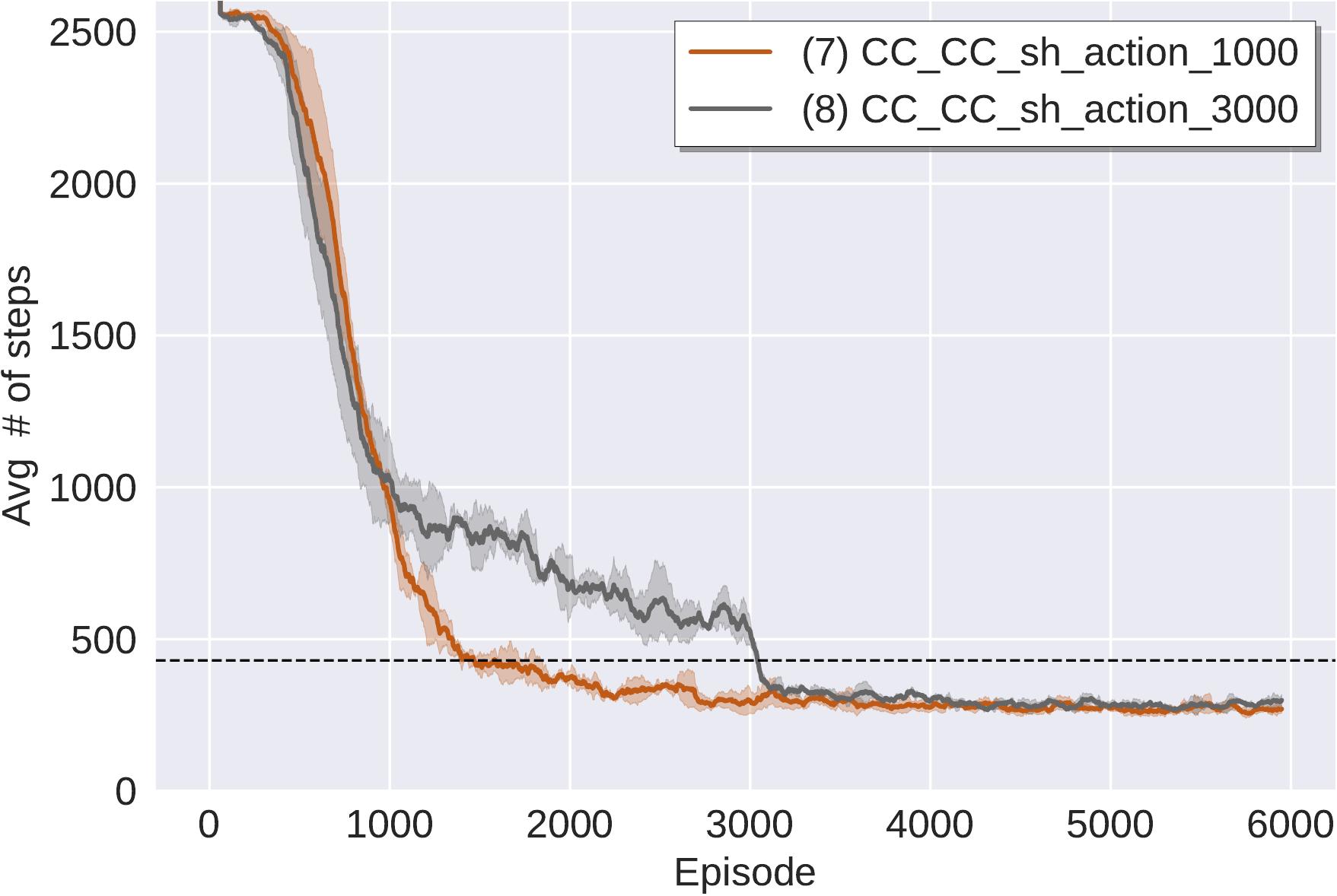} \\
        (c) & (d)
    \end{tabular}
    \caption{Performance comparison when switching off the intrinsic motivation stream at different stages of the training process(1000 and 3000 episodes). (a, b) Success rates and (c, d) average number of steps attained by $W_0$ (a and c) and $W_1$ (b and d).}
    \label{fig:premature_convergence_cut}
\end{figure}

As can be checked in Figure \ref{fig:premature_convergence_cut}, one of the main consequences of an early curiosity stopping criterion (namely, that imposed in \texttt{CC\_CC\_sh\_action\_1000}) is that agent $W_0$ is not able to discover its optimal path (the corridor). Consequently, the quality of the policy improves faster even across the suboptimal path. However, the best policy is achieved when ensuring that the agent traverses the corridor, and it is only then when the intrinsic motivation must be switched off (as done in \texttt{CC\_CC\_sh\_action\_3000}). This can be reflected at the number of required steps shown in Figure \ref{fig:premature_convergence_cut}.c where although when turning off the curiosity at episode 1000 improves earlier the quality of the policy, the final result is better when going through the corridor.

On the other hand, at the non-skilled agent $W_1$ the differences are more related on when it begins to optimize its quality to be aligned to the extrinsic goal. This is because for $W_1$ there is no distinction between optimal or suboptimal paths, they are in fact the same. Thus, the solution at episode 1000 is already optimal and from that point in advance it refines its policy requiring less training samples and yielding shorter training times in comparison to waiting until episode 3000.

Finally, the average extrinsic reward curves do not show any meaningful differences when compared to the case where intrinsic motivation is maintained during the entire training procedure (Figure \ref{fig:action_based_curiosity}).

\begin{table*}[ht]
    \centering
    \begin{tabular}{lccccccccc}
     \toprule
     &\multicolumn{5}{c}{Number of required episodes} & & & \\
     \cmidrule{2-6}
     Algorithm & \multicolumn{2}{c}{90\% SR} & & \multicolumn{2}{c}{80\% SR (corridor)} & & \multicolumn{2}{c}{Number of steps to goal} \\
     \cmidrule{2-3} \cmidrule{5-6} \cmidrule{8-9}
     & $W_0$ & $W_1$ & & $W_0$ & $W_1$ & & $W_0$ & $W_1$ \\
     \midrule
     \texttt{IC\_IC\_3r} & $\geq 6000$ & $\geq 6000$ & & $\geq 6000$ & - & & - & - \\
     \texttt{IC\_IC\_6r} & $\geq 6000$ & $\geq 6000$ & & $\geq 6000$ & - & & - & - \\
     \texttt{CC\_IC} & 1522 & 1110 & & 3668 & - & & $410.72\pm 12.15$ & $470.98\pm 19.15$ \\
     \texttt{CC\_CC\_sh} & 1309 & 1026 & & 3445 & - & & $403.07\pm 9.25$ & $455.88 \pm 5.05$ \\
     \texttt{CC\_CC\_sh\_action} & 1412 & 1007 & & 2389 & - & & $456.81\pm 6.24$ & $530.27\pm 8.47$ \\
     \texttt{CC\_CC\_sh\_action\_filter} & 1208 & 858 & & 2378 & - & & $418.47\pm 7.91$ & $492.2\pm 21.06$ \\
     \texttt{CC\_CC\_sh\_action\_1000} & 1213 & 937 & & $\geq 6000$ & - & & $314.45\pm 7.17$ & $267\pm 3.17$ \\
     \texttt{CC\_CC\_sh\_action\_3000} & 1722 & 886 & & 2472 & - & & $267.9 \pm 7.57$ & $294.64\pm 3.44$ \\
     \bottomrule
    \end{tabular}
 \caption{Number of required episodes needed by each evaluated configuration of the collaborative learning framework to achieve a SR equal to 90\% (first column) and a SR equal to 80\% when success is counted as such when the trajectory includes traversing the corridor (second column); quality of the trained policies in terms of the required number of steps to achieve the goal (statistics computed over the last 100 training episodes, removing those episodes in which the agent does not reach the target).}
 \label{table:ablations_results}
\end{table*}

On a concluding note from these experiments, the challenge remains in knowing when to switch off or increase $\beta$ in order to avoid undesired learning signals that are not aligned with the main objective of the problem. It is important not to deactivate curiosity incentives prematurely: in early learning stages, the optimal solution may not have been learned yet by an agent, so speeding up the training process by promoting exploitation can be counterproductive, specially when dealing with sparse rewards. As will be elaborated further in Section \ref{sec:discussion}, this challenge also holds in other RL problems, where it is unclear whether agents have explored sufficiently towards exploiting the already captured knowledge. A high $\beta$ value should be beneficial at early training stages to foster explorative interactions with the environment, while a lower $\beta$ value should inject less noise into the exploitation process at late stages of the policy learning process.


\subsection{Overall Comparison}

To sum up, an overall summary of the previously discussed results is given in Table \ref{table:ablations_results}. By first analyzing when the agents are capable of reaching the goal consistently, the introduction of a centralized critic is undoubtedly a differentiating factor. Another remarkable difference appears when using an intrinsic motivation strategy that takes into account both the state and the action, which imprints a better convergence through the optimal path through the corridor for $W_0$ (a reduction of $1000$ training episodes to achieve 90\% SR,an improvement of approximately 30\%). In this same line, the performance of $W_1$ is also improved first by the introduction of a centralized curiosity, and thereafter with the action-based intrinsic motivation, reducing by almost 10\% the total number of episodes to attain 90\% SR. 

One interesting fact in these summarized scores is that $W_1$ always converges faster, which makes sense as it has a smaller action and solution space. Consequently, the state value estimates $V(s)$ along the path followed by this agent will be more updated and will have better estimates when compared to those that are part of the optimal path of $W_0$ (which may have been negatively influenced to go through the path already paved by $W_1$). Surprisingly, despite the probability of getting through the corridor is low after $W_0$ has consistently found out the way to achieve the destination ($\leq 10\%$), the intrinsic motivation plays a big role to keep the agent exploring until it finally finds a better solution and increases the chances to go through the corridor. Hence, maintaining a certain degree of exploration is crucial to ensure that all the possible alternative paths have been sufficiently visited. This can be confirmed also with the former analysis of \texttt{CC\_CC\_sh\_action\_1000} in Figure \ref{fig:premature_convergence_cut} where, by turning off the curiosity reward (and hence decreasing the exploration), $W_0$ learns to reach the goal only across its suboptimal path. 

On the other side, once the agents are capable of reaching the destination, they begin to improve their decision making and consequently require progressively less number of steps. This is part of the feedback system where the critic improves his predictions. Thanks to this, the policy selects better decisions through advantages, so that the actor selects increasingly limited samples and the critic learns their differences more effectively. Remarkably, all experiments comprising configurations that have maintained the curiosity during the whole training barely achieve a performance close to a human-driven agent, even by using a frameskip of 4. Nevertheless, the number of steps required to achieve the target by $W_0$ are in general less than the ones required by $W_1$, which makes sense as $W_0$ affords a shorter path to the target. However, this is not the case of \texttt{CC\_CC\_sh\_action\_1000}, where $W_1$ achieves the goal faster than $W_0$ due to the combined effect of achieving a 100\% SR faster and a less diverse action space.


\section{Lessons Learned and Research Directions}\label{sec:discussion}

Grounded on the insights extracted from the previous analysis, in this section we sketch lessons learned and interesting directions for future research stemming from the development of this work. Some of the reflections offered in what follows relate to the heterogeneity between agents, whereas others related to issues that lie at the heart of RL and intrinsic motivation, which also affect the particular scenario tackled in this work.

\subsection{When to explore? On the Exploration-Exploitation Dilemma with Heterogeneous Agents}

A known challenge in RL is when to explore and when to exploit in environments comprising a single learning agent. Besides the strong dependence on the characteristics of the environment, there are different types of exploration strategies that can be followed with diverse results \cite{pislar2021should}. Even in the simpler single-agent scenario, it is not clear how to make the agent explore efficiently. When should a given agent explore? This problem is exacerbated in settings with sparse rewards, specially when the completion of the task can require long-term training horizons. By virtue of intrinsic motivation techniques, the agent can explore most of the time it interacts with the environment with a non-stationary novelty bonus that changes over time. This novelty bonus may not be aligned with the overall extrinsic goal, which in some training stages can be even worse than not using it whatsoever \cite{taiga2019benchmarking}.
    
Moreover, there is no direct correlation between the heterogeneous agents our framework aims to learn collaboratively. This clashes with the assumptions of multi-agent RL tasks, where they share at least a team reward, or the environment where they perform is the same for all agents. In fact, the observations and the decision making behind each policy are different in our case. Thus, when should we share a collaborative strategy with no link between agents even if we do not know when we should explore? It is hard to give an answer to that question even in single-agent scenarios. In our work we have assumed some kind of latent knowledge between agents and tasks, which is also the hypothesis behind transfer learning approaches. This, unfortunately, might not be enough in some RL scenarios.

\subsection{On the Detachment-Derailment Problem}

Approaches related to sparse RL tasks that rely on intrinsic motivation exhibit the so-called \textit{detachment-derailment} problem. Recently it has been shown that an effective manner to address this issue is by using cluster of representations, and reinitializating the agent smartly in the environment \cite{ecoffet2021first,ugadiarov2021long}. But this requires the environment to be \textit{reset-free}, namely, an environment in which the agent position and/or state perception can be manually selected without any constraints. This property grants flexibility to select new/desired start positions arbitrarily.

This problem arises when an agent has explored the environment correctly, becoming close to discovering a interesting state-space or to achieving the goal. At some point, however, the agent's learning gets stuck and the episode finishes. Next the episode is restarted, and all decisions that the agent took to reach those spots are now seen with less novelty (even being close to finding out promising locations). Consequently, the agent will be stimulated to check other alternatives, even if being in the right direction to discover what was really important in the environment, and ultimately degrading the effectiveness of the exploration.

The \textit{detachment-derailment} problem is further compounded if the time horizon required to achieve any meaningful feedback signal increases. In our work we have noted that, although the agent leverages the inclusion of intrinsic motivation, it is hard for it to visit back an already promising visited state. The reason being that the required steps to achieve the state may have not be learned consistently yet by the actor, so the agent gets lost in the early steps of the episode. 

In the scenario with heterogeneous agents tackled in this work, a similarity-based clustering of the state space might be suitable to identify promising states to explore. Clustering-based  proposals have been contributed in the literature for single-agent problems, which are mostly designed ad-hoc for the task at hand \cite{ecoffet2021first,ugadiarov2021long}. Unfortunately, state space clustering for POMDPs with first-person-view observations is not straightforward due to the dimensionality of the state space, nor does it effectively determine where to reinitialize each agent considering that they may have different stimuli and optimal paths for the same goal.

\subsection{On the Potential of Recurrent Rewards}

Another issue encountered during this research work springs from the fact that intrinsic bonuses are generated from a given experience tuple rather than a sequence of tuples. This issue affects not only the scenario tackled in this manuscript, but also other RL environments that generate intrinsic rewards based on single experiences. This mainly occurs when having a POMDP, as changes in the environment cannot be directly reflected even if those changes have a clear impact in the environment. 
\begin{figure}[h!]
    \centering
    \begin{tabular}{cc}
        \includegraphics[width=0.5\columnwidth]{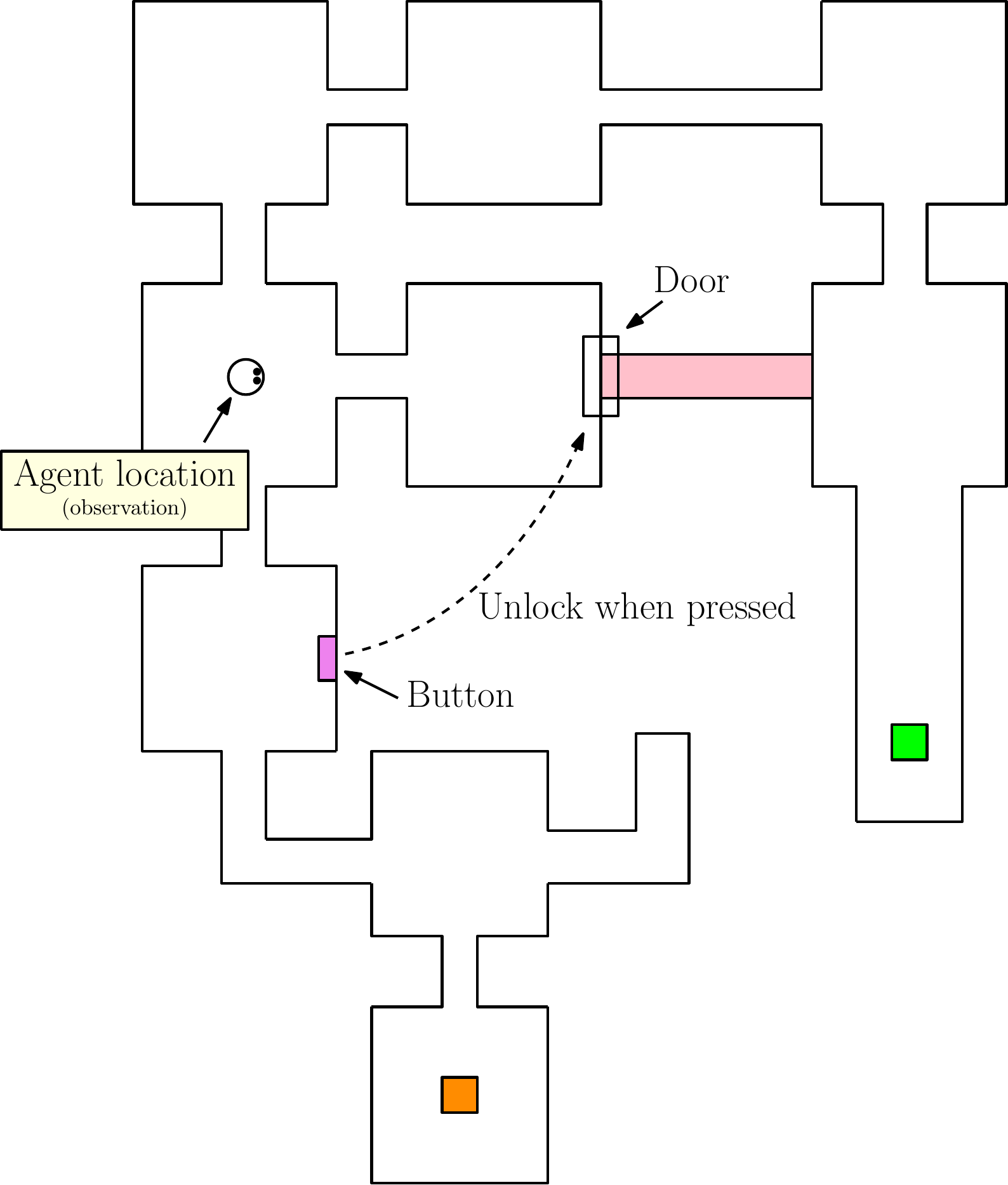} & \includegraphics[width=0.44\columnwidth]{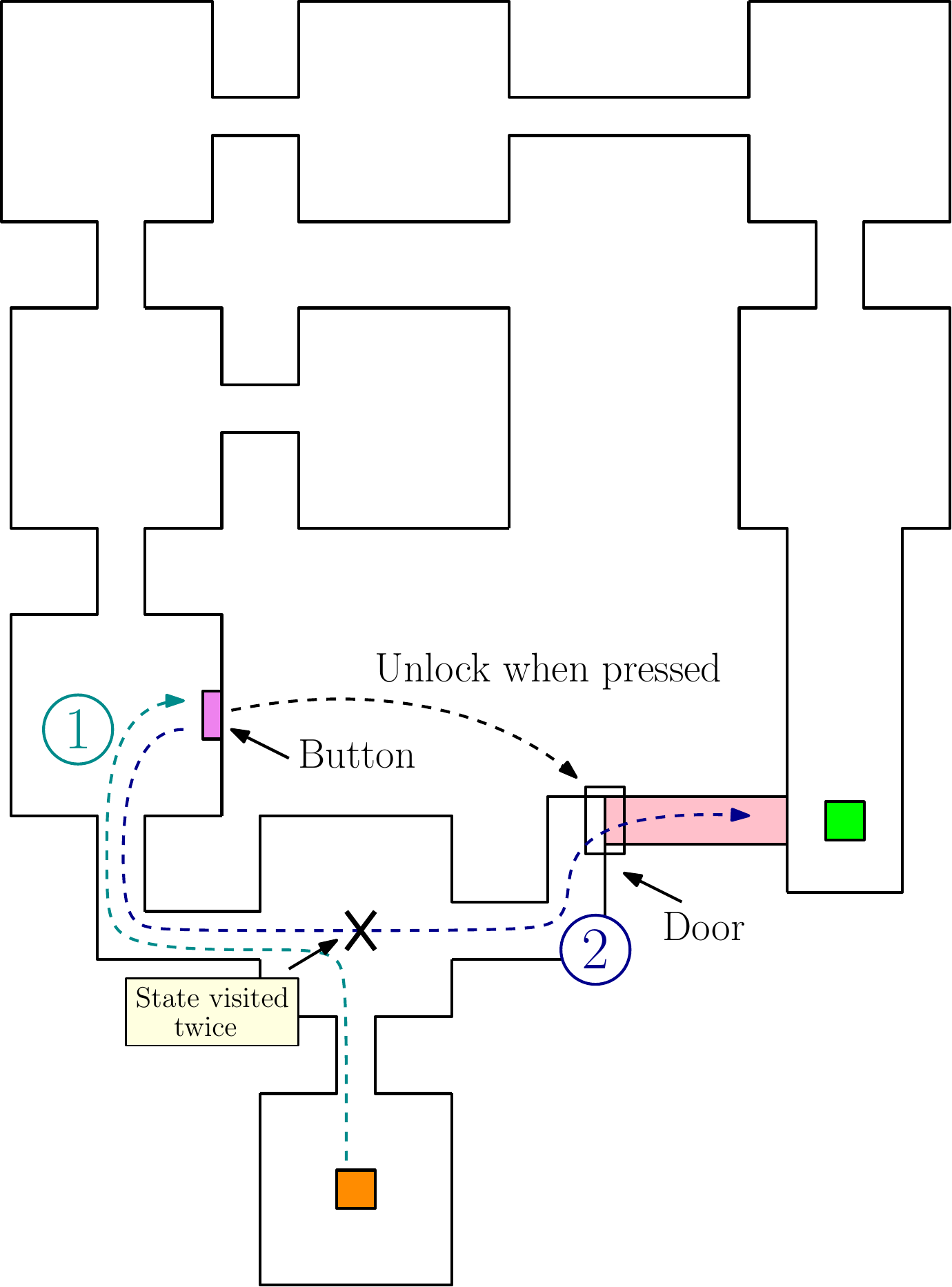} \\
        (a) & (b)
    \end{tabular}
    \caption{Case studies hypothesized to discuss on how to deal with long-term dependencies in (state, action) experiences.}
    \label{fig:case_study_discussion}
\end{figure}

We further expose this problem by briefly discussing on two different maps depicted in Figure \ref{fig:case_study_discussion}. In these maps an agent can unlock the colored passage by pushing the button that is located at a different and far location. The \texttt{OPEN} action is only necessary for unlocking the shortcut, with no additional utility in the rest of the environment. Agents will necessarily have to memorize in a long horizon that pushing the button will open a door, although this event is not \textit{noticeable} in the current and next observations. The value of reaching the location where the button is located will indeed differ depending whether the button is pushed and then used to go through the corridor or any other alternative. Consequently, the same observation have a strong temporal dependency that must be modeled correctly by the agents.

We now focus on the other environment shown in Figure \ref{fig:case_study_discussion}.b, where the same spot must be visited twice by any agent: one \smash{\circled{1}{white}{darkcyan}{darkcyan}} when searching for the button that opens the passage, and another one \smash{\circled{2}{white}{darkblue}{darkblue}} to go through the passage itself. In this case, the aforementioned problem is also present, but the complexity increases as the same observation should receive different values depending whether the corridor has been unlocked, even if no information for this purpose is available in the observation at the location.

The novelty of a given observation can be interpreted in two ways:
\begin{itemize}
    \item As the intrinsic reward for a given experience tuple, aiming to quantify \textit{how novel the experience is on its own}.
    \item As the discounted expected return inside a given trajectory taking into account the calculated intrinsic bonus or the experiences that compose that given trajectory, answering \textit{which degree of novelty this experience injects into future steps of the episode}.
\end{itemize}

The first only depends on itself to measure the novelty, whereas the second relies on other experiences. 

From a practical perspective, once the return is calculated, temporal dependencies among the experiences can be modeled with recurrent neural networks with memory and attention mechanisms that can be implemented at the policy (actor), the critic or both \cite{hausknecht2015deep,oh2016control,vaswani2017attention}. As shown in Figure \ref{fig:critic_architecture}, the critic module of our proposed framework employs a LSTM network. However, there are no guarantees that this type of architecture retains the gathered knowledge at long-time horizons, nor is the novelty score used to compute the return stationary (it decreases over time). This makes the expectation term unstable over time, and ultimately jeopardizes the long-term modeling capabilities of the LSTM network. 

A solution would be to generate intrinsic rewards based not only on the current time step, but also on past experiences, i.e. a sequence of experiences. This is, designing a reward function that handles the temporal dependencies and provides a different reward value, so that a experience is determined to be novel taking into account a full episode or path with its inherent consequences. This problem has also been recently depicted at \cite{colas2020intrinsically} related to goals. A debate is held around how to address this issue in an online fashion with no previous knowledge about the environment. This open discussion finds in the action heterogeneity of agents studied in this manuscript another twist of its screw.


\section{Conclusions and Outlook}\label{sec:conclusions}


In this work we have analyzed different ways in which heterogeneous RL agents can share information with each other over the same environment, taking into account that the achievement of their results is unique and exclusive to each agent. The main goal for both agents is to learn faster than they would independently on their own, without any kind of knowledge sharing strategy. To address this scenario and arrive at informed conclusions, we have proposed a collaborative learning framework that embeds a centralized critic module that allows accelerating the learning process, together with the use of an exploration scheme governed by intrinsic motivation driven not only by the state the agent, but also by the action performed when entering that state.

An extensive experimentation has been performed over a modified 3D ViZDooM environment to evaluate the results of the collaborative framework configured differently in terms of \emph{what} is shared between agents (critic, curiosity) and \emph{when} it is done. In our experiments agents are only fed with first-view observations/images collected from the environment, only receiving a reward if they reach the given destination. Likewise, changes made to the environment have been done so that one of the agents has a different optimal solution space than the other due to the heterogeneity between their action domains. Several conclusions have been drawn from these experiments:
\begin{itemize}
    \item First, results have clearly elucidated that a centralized critic yields better stability and quicker convergence in learning compared to the case when critics are not shared. Another advantage of centralized the critic is the fact that only a single neural network is trained over diverse experiences fed from the agents. 
    
    \item On the other hand, exploration between agents can be affected depending on whether the novelty is shared: centralizing it entails some minor advantages, but the differential impact was noted when the novelty is set dependent not only on the state, but also on the action. This interesting result had gone unnoticed in past literature dealing with single-agent environments \cite{tang2017exploration}, where it was reported that a dependence on the action had no effect in the convergence of the learning process. However, in environments with heterogeneous agents, our results show the opposite: an action-dependent novelty computation can have a significant impact when discovering the optimal solution, achieving a reduction of up to 30\% in terms of the number of episodes.
    
    \item Finally, it has been observed that, although the intrinsic bonus favors exploration, after a certain moment it induces noise into the training, degrading the effectiveness of the exploitation phase. Therefore, the obtained policy is not as optimal as desired and remains too stochastic. This aligns with other works where, once a certain degree of knowledge has been obtained and the exploration is already considered sufficient, the fact of continuing to use it results to be counterproductive for the learning process \cite{rosser2021curiosity,taiga2019benchmarking}.
\end{itemize}

As a result of this finding and others found during the course of the investigation, several future research directions have been outlined and argued in Section \ref{sec:discussion}. Ideas posed therein do not focus only on heterogeneous agents, but also on scenarios with sparse rewards and intrinsic motivation that have room for improvement. Among them, we envision that a great challenge resides in the definition of the best exploration strategy to follow, as well as the on-line decision when to explore in all types of scenarios \cite{pislar2021should}. To this end, recent advances in the combination between intra-episodic and experiment-long bonuses seem to be very promising \cite{badia2020agent57}. On the other side, clustering-based state representations strategies have protruded as a good alternative to address the detachment-derailment issue. In accordance to the insights from our study with heterogeneous agents, it will be interesting to verify whether it can be extrapolated to (state, action) spaces. Finally, problems derived from the temporal dependencies in the computation of the intrinsic bonus are particularly challenging in our tackled scenario, as two very similar trajectories can diverge in the future because of changes happening in the present. This requires different definitions of novel rewards, this is, the design of a reward module that relies on the information obtained from a sequence of experiences rather than by a single (state, action) tuple. 

\begin{acknowledgements}
The authors would like to thank the Basque Government for its support through the ELKARTEK program (3KIA project, KK-2020/00049). A. Andres receives funding support from the Basque Government through its BIKAINTEK PhD support program. J. Del Ser also acknowledges support received from the same institution through the consolidated research group MATHMODE (ref. T1294-19).
\end{acknowledgements}

\section*{Author contributions}

Conceptualization: A. Andres, E. Villar-Rodriguez, J. Del Ser; Me\-thodology: A. Andres, E. Villar-Rodriguez; Formal analysis and investigation: A. Andres, E. Villar-Rodriguez; Writing - original draft preparation: A. Andres; Writing - review and editing: E. Villar-Rodriguez, J. Del Ser; Funding acquisition: E. Villar-Rodriguez, J. Del Ser; Supervision: E. Villar-Rodriguez, J. Del Ser.

\section*{Conflict of Interest}

The authors declare that they have no conflicts of interest regarding this work.

\bibliographystyle{unsrt}
\bibliography{references}

\end{document}